\documentclass[ms, nonblindrev]{informs3}

\OneAndAHalfSpacedXI %

\usepackage{amsmath}
\usepackage{amssymb} 
 \usepackage[ruled,vlined]{algorithm2e} 
\usepackage{bm}
\usepackage{booktabs}
\usepackage{diagbox}
\usepackage{footnote}
\usepackage{mathrsfs}
\usepackage{mathtools}
\usepackage{verbatim}
\usepackage{multirow}
\usepackage{natbib}
\usepackage{subfigure}  
\definecolor{DarkBlue}{rgb}{0.0,0.0,0.55}
\usepackage[backref = false, bookmarks, colorlinks = true, plainpages = false, citecolor = DarkBlue , urlcolor = DarkBlue, filecolor = DarkBlue, linkcolor = DarkBlue]{hyperref}
\usepackage[hyphenbreaks]{breakurl}
\usepackage{fancyhdr,graphicx} 
\usepackage{threeparttable}

\bibpunct[, ]{(}{)}{,}{a}{}{,}%
\def\bibfont{\small}%
\def\bibsep{\smallskipamount}%
\def\bibhang{20pt}%
\newcommand{\Tr}{\mathrm{Tr}}

\newcommand{\prob}{{\mathbb{P}}}

\TheoremsNumberedThrough     %
\ECRepeatTheorems

\EquationsNumberedThrough    %

\begin{document}

	 \RUNAUTHOR{Bu, Simchi-Levi, and Xu}
	
 	\RUNTITLE{Online Pricing with Offline Data}

	\TITLE{Online Pricing with Offline Data:\\ Phase Transition and Inverse Square Law}

\ARTICLEAUTHORS{%
        \AUTHOR{Jinzhi Bu}
        \AFF{Department of Logistics and Maritime Studies, Faculty of Business, The Hong Kong Polytechnic University, Hung Hom, Kowloon, Hong Kong, \EMAIL{jinzhi.bu@polyu.edu.hk}}
		\AUTHOR{David Simchi-Levi}
		\AFF{Institute for Data, Systems, and Society, Massachusetts Institute of Technology, Cambridge, MA 02139, \EMAIL{dslevi@mit.edu}}
		\AUTHOR{Yunzong Xu}
		\AFF{Institute for Data, Systems, and Society, Massachusetts Institute of Technology, Cambridge, MA 02139, \EMAIL{yxu@mit.edu}}
} %

	\ABSTRACT{%
		This paper investigates the impact of pre-existing  offline data on online learning, in the context of dynamic pricing. We study a single-product dynamic pricing problem over a selling horizon of $T$  periods. The  demand in each period is determined by the price of the product  according to a linear demand model with unknown  parameters.  We assume that before the start of the selling horizon, the seller already has some pre-existing offline data. The offline data set contains $n$ samples, each of which is an input-output pair consisting of a \textit{historical price} and an associated demand observation.  
		The seller wants to utilize both the pre-existing offline data and the sequentially-revealed online data to minimize the regret of the online learning process.
		
		We  characterize the joint effect of the \textit{size}, \textit{location} and \textit{dispersion} of the offline data on the optimal regret of the online learning process. Specifically, the \textit{size}, \textit{location} and \textit{dispersion} of the offline data are measured by the number of historical samples $n$,    the distance between the average historical price and the optimal price $\delta$, and  the standard deviation of the historical prices $\sigma$, respectively. 
		For the single-historical-price setting where the $n$ historical prices are identical, we prove that the best achievable regret is  $\widetilde{\Theta}\left(\sqrt{T}\wedge \frac{T \log T}{(n\wedge T)\delta^2}\right)$.  
		For the (more general) multiple-historical-price setting where the historical prices can be different, we  show that  the best achievable regret is $\widetilde \Theta\left(\sqrt{T}\wedge \frac{T\log T}{n\sigma^2+(n\wedge T)\delta^2}\right)$.
		For both settings, we design a learning algorithm based on  the ``Optimism in the Face of Uncertainty'' principle, which strikes a balance between exploration and exploitation, and achieves the optimal regret up to a logarithmic factor. Our results reveal surprising transformations of  the optimal regret rate with respect to the size  of the  offline data, which we refer to as \textit{phase transitions}. In addition, our results demonstrate that the location and dispersion of the offline data also have an intrinsic effect on the optimal regret, and we quantify this effect via the \textit{inverse-square law}. 
	}%
	\KEYWORDS{dynamic pricing, online learning, offline data, phase transition, inverse-square law}

	\maketitle

	\parskip=5.3pt

\section{Introduction} 
Classical statistical  learning theory distinguishes between offline learning and online learning. 
Offline learning deals with the 
problem of finding a predictive function based on the entire training 
data set. 
The  performance of an offline learning algorithm is typically measured by    its \text{generalization error} (also known as the out-of-sample error) or \text{sample complexity}  (see, e.g., \citealt{hastie2005elements}). 
In contrast to the 
offline learning setting where the entire training data set is directly 
available before the offline learning algorithm is applied, online learning deals with a setting where 
data become available in a sequential manner that may depend on the   actions taken by the online learning algorithm. The performance of  online learning algorithms is typically measured by the regret\footnote{In this paper, when we discuss online learning, we focus more on the literature of stochastic online learning, where the online sequential data arrive in a stochastic manner. There is a vast literature of online learning focusing on the non-stochastic setting where the online sequential data arrive in an adversarial manner (see \citealt{cesa2006prediction}), which is not the emphasis of this paper.}.  While offline learning assumes access to offline data (but not online data) and online learning assumes access to online data (but not offline data), in reality, a broad class of real-world problems incorporate both aspects:   there is an offline historical 
data set (based on historical actions) at the time that the learner starts an online learning process.  

Currently, there is no standard framework for the above type of learning 
problems, as classical offline learning theory and online learning theory 
have different settings and goals. While establishing a framework that  bridges all aspects of offline learning  and online learning is generally a very complicated task, in this paper, we propose a   framework that  bridges the gap between
offline learning and online learning in a specific problem setting,  which, 
however, already captures the essence of many dynamic pricing problems 
that sellers face in practice.

\subsection{The Model: Online Pricing with Offline Data}
In this paper, we study the  \textit{Online Pricing with Offline Data} (\texttt{OPOD}) problem. Consider a firm selling a single product with an infinite amount of inventory over a selling horizon of $T$ periods. In each period $t =1, 2, \ldots, T$, the seller chooses a price $p_t $ from a given interval $[l,u] \subset [0,\infty)$ to offer to its customers, and then observes  random  demand $D_t$.   
	We assume that the demand in each period is a linear function of the price plus some random noise. Specifically,  for each $t \ge 1$, 
\begin{align}\label{eq-linear demand function}
D_t=\alpha^*+\beta^* p_t+\varepsilon_t,
\end{align}
where $\alpha^*$ and $\beta^*$ are two unknown demand parameters in the known interval  $  [\alpha_{\min},\alpha_{\max}] \subseteq (0,\infty)$ and $[\beta_{\min}, \beta_{\max}]  \subseteq (-\infty,0)$ respectively, and  $\{\varepsilon_t\}_{t=1}^T$ are \textit{i.i.d.}  random variables  with  zero mean and   unknown   distribution.   We    assume that   $\varepsilon_t$ is an $R^2$-sub-Gaussian random variable, i.e., there exists a constant $R>0$ such that $\mathbb{E}[e^{x\varepsilon_t}] \le e^{\frac{x^2R^2}{2}}$ for any $x \in \mathbb{R}$. 
 For notational convenience, let $\theta^*:=(\alpha^*,\beta^*)$ and  
	$\Theta^{\dag}:=[\alpha_{\min}, \alpha_{\max}]\times [ \beta_{\min}, \beta_{\max}]$, and  
we use $\theta:=(\alpha,\beta)$ to denote any possible vector in parameter space   $\Theta^{\dag}$.  

The seller's single-period expected revenue is the price $p$ offered to the customer multiplied by the associated expected demand. To emphasize the dependence on the parameter values, for any $\theta=(\alpha,\beta) \in \Theta^{\dag}$, we  define the expected revenue  function  $r(p; \theta)$  as  $
r(p; \theta)=p(\alpha+\beta p), \,\, \forall p \in [l,u]$. 
Let $\psi(\theta)$ be the price that maximizes  $r(p; \theta)$   over the interval $[l,u]$,   i.e., 
$
\psi(\theta)=\arg\max \big\{r(p; \theta) : p \in [l,u] \big\}$, and use $p^*$ to denote the   true optimal price, i.e., $p^*=\psi(\theta^*)$.  Let $r^*(\theta)$ be the optimal expected revenue under demand parameter $\theta$, i.e., $r^*(\theta)=\psi(\theta)(\alpha+\beta \psi(\theta))$.  
 Without loss of generality,\footnote{This is because for any $\theta \in \Theta^{\dag}$, $\frac{\alpha}{-2\beta} \in [\frac{\alpha_{\min}}{-2\beta_{\min}} , \frac{\alpha_{\max}}{-2\beta_{\max}} ]$, and we can  choose $l$ and $u$ such that $[\frac{\alpha_{\min}}{-2\beta_{\min}} , \frac{\alpha_{\max}}{-2\beta_{\max}} ] \subset [l,u]$, which guarantees that $\frac{\alpha}{-2\beta}$ is an interior point of   interval $[l,u]$.} 
we assume that for any $\theta \in \Theta^{\dag}$,  the optimal price is an interior point of  price range $[l,u]$, and therefore  $\psi(\theta) = \frac{\alpha}{-2\beta}$. 

\textbf{Historical prices and offline data.} In reality, the  seller does not know the true demand model, but has to learn such information from  the  historical data.  
In this paper, we   assume that the seller has some pre-existing offline data before the  start of the online learning process. 
The offline data set contains $n$ independent samples: $\{(\hat p_1, \hat D_1), (\hat p_2, \hat D_2), \ldots, (\hat p_n, \hat D_n)\}$, where $\hat p_1, \hat p_2, \ldots, \hat p_n$ are $n$ fixed prices, and   each $\hat D_i$ is a demand observation under  historical price $\hat p_i$, drawn  independently  according to the underlying  linear demand model \eqref{eq-linear demand function}.  Therefore,   for each $1 \le i \le n$, $ \hat D_i=\alpha^*+\beta^* \hat p_i+\hat \varepsilon_i$  for some   \textit{i.i.d.} random variables $\{\hat \varepsilon_i\}_{i=1}^n$  with the same distribution as that of  $\{\varepsilon_t\}_{t=1}^T$.

\textbf{Pricing policies and performance metrics.}   For each $t \ge 0$, let $H_t$ be the vector of information available  at the beginning of period $t+1$, i.e., $H_t=(\hat p_1, \hat D_1, \ldots, \hat p_n, \hat D_n, p_1, D_1,\ldots, p_t,D_t)$.  
A pricing policy is defined as a sequence of functions $\pi=(\pi_1, \pi_2, \ldots)$, where  
each $\pi_{t}$ is a measurable function which maps the realization of $H_t$ (and possibly some external  randomness) to a feasible price in $[l,u]$.  
Let $\Pi$ be the set of all   pricing policies.  
For any  policy $\pi \in \Pi$,   
the \textit{regret} of $\pi$, denoted by $R^{\pi}_{\theta^*}(T)$,  is  defined as the difference between the optimal expected revenue  generated by  the clairvoyant policy that  knows the exact value of $\theta^*$  and the expected revenue generated by   pricing policy $\pi$, i.e., 
 \begin{align*}
	R^{\pi}_{\theta^*}(T) =T\cdot r^*(\theta^*)- \mathbb{E}^{\pi}_{\theta^*}\Big[\sum_{t=1}^{T} r(p_t; \theta^*)\Big], 
	\end{align*}

\subsection{Research Question, Observations and  Challenges}\label{sec-challenges}
This  paper  is inspired by the objective  of bridging the gap between offline  learning  and online learning. The following question naturally arises  
whenever the  offline data are incorporated into the online decision making: how do the offline data affect the \textit{statistical complexity} of  online learning? 
 To address this question, the first challenge is to identify the key characteristics of the offline data  that intrinsically affect the complexity of the online learning task.

Intuitively, the {\textit{size}} of the offline data, measured by the number of historical samples $n$, and  the {\textit{dispersion}} of  the offline data, measured by the standard deviation of historical prices $\sigma$, i.e., $\sigma=\sqrt{\sum_{i=1}^n (\hat{p}_i-\frac{1}{n}\sum_{j=1}^n\hat{p}_j)^2}$, provide two important metrics that enable quantifying   the amount of information collected before the online learning process starts. 
As   $n$ becomes larger, or   $\sigma$ increases, the seller can form a better estimation for the unknown demand parameters using   offline regression, and the regret may decrease accordingly.

Another crucial, and more intriguing metric of  the offline data, is the \textit{location} of   the offline data,  which is measured by $\delta= \vert \frac{1}{n}\sum_{i=1}^n\hat{p}_i-p^*\vert$, 
i.e.,  the distance between the average historical price and the optimal price $p^*$. 
We refer to $\delta$ as the \textit{generalized distance}, as it intuitively quantifies how far the offline data set is ``away'' from the (unknown) optimal decision. 
 This is a crucial metric that {uniquely} appears when offline data are incorporated into the online  learning process. Indeed, if there are no offline data   available before the start of the online learning process, then there is no  $\delta$ at all. Also, if the offline data are only used for estimation or prediction, with no need of online decision making, i.e., the seller is purely interested in estimating the model parameters from the offline data and does not need to make any online pricing decisions, then $\delta$ does not affect her estimation accuracy.
 Surprisingly, as we prove in this paper, when the offline data are incorporated into   online learning, this metric will play a fundamental role.

Besides identifying the above three  characteristics of the offline data, 
a key challenge is to precisely quantify the effects of these  offline data characteristics on the online learning task. 
Specifically, we seek to understand to what extent these three metrics of the offline data  influence the behavior  and growth rate of the best-achievable regret bound.   
 On the algorithmic side, we also seek to design a simple, intuitive and easy-to-implement pricing policy that exploits the values of  both the pre-existing  offline data and sequentially-revealed online data, and achieves a tight regret bound with respect to  the selling horizon $T$, as well as the  three metrics of the offline data, i.e., $n$, $\sigma$, $\delta$.  
Moreover, since  
   the generalized distance $\delta$  is   completely  {unknown}  to the seller,  the algorithm itself  cannot use any information about $\delta$, which implies  a more challenging task of designing a learning algorithm whose performance is as good as if   $\delta$ were known.

 \subsection{Main Results and Technical Highlights}\label{subsec-main results}  
  In this paper, we  address the above challenges in two settings: (i)   \textit{single-historical-price} setting where all the historical prices are identical,  i.e., $\sigma=0$, and (ii)   \textit{multiple-historical-price} setting where the historical prices can be different,  i.e., $\sigma>0$.  
 We next summarize our main results and technical highlights. 
 Throughout this paper, we use   $\widetilde{\mathcal{O}}(\cdot)$ and $\widetilde{\Omega}(\cdot)$   to present upper and lower bounds on the growth rate up to logarithmic factors, and $\widetilde{\Theta}(\cdot)$ to characterize the rate when the upper and lower bounds match (up to logarithmic factors). In addition,  
  we use $A \lesssim B$ and $A \gtrsim B$ to denote $A=  \widetilde{\mathcal{O}}(B)$ and $A=  \widetilde{\Omega}(B)$  respectively. More formal definitions of these notations are provided in \S \ref{subsec-structure}. For any $a, b \in \mathbb{R}$,   $a\wedge b=\min\{a, b\}$ and  $a \vee b = \max\{a, b\}$. 
  
 \textbf{Single-historical-price setting.}  
 For the single-historical-price setting, we develop a learning algorithm called   \textit{Online and Offline--OFU} (O3FU) algorithm, where OFU refers to   the principle of \textit{Optimism in the Face of Uncertainty},  which arises from  multi-armed bandits and is widely used in 
the literature on bandits (see, e.g., \citealt{dani2008stochastic}, \citealt{abbasi2011improved}). In general,  this principle  suggests taking actions based on an optimistic guess of the reward associated with each action in each period. 
    {We show that  the regret of O3FU algorithm  has an upper bound  $\widetilde{\mathcal{O}}\big(\sqrt{T}\wedge  \frac{T\log T}{(n\wedge T)\delta^2}\big)$.} 
    	Although this upper bound  depends on the unknown quantity  $\delta$, the   algorithm itself does not require any information about $\delta$.
     In addition, we prove an information-theoretic lower bound which matches the upper bound, showing that the regret bound cannot be further improved by other algorithms (in a certain sense);  
   we define such an unimprovable regret bound  as the \textit{optimal  (instance-dependent) regret} for the \texttt{OPOD} problem in the single-historical-price setting. We summarize its rate  in Table  \ref{table-general case}. In particular, when $n=0$, or  $n=\infty$ and $\delta$ is a constant independent of $T$, the results in
    the leftmost and rightmost
cases with $\delta\gtrsim T^{-1/4}$ in Table \ref{table-general case} recover those in \cite{KeskinZeevi2014}.

 {\textbf{Multiple-historical-price setting.}  
  For the general setting that the historical prices may be different,  we
  modify   O3FU algorithm by  adding a preliminary step that detects whether a \textit{corner case} $\delta^2 \lesssim \frac{1}{n\sigma^2} \lesssim \frac{1}{\sqrt T}$ happens or not, and propose the \textit{Modified  O3FU} (M-O3FU) algorithm.  
   {We prove that  
   	M-O3FU algorithm achieves the  
   	regret upper bound   $\widetilde{\mathcal{O}}(\sqrt T \wedge \frac{T\log T}{n\sigma^2+(n\wedge T)\delta^2})$,  except for a corner case 
   	where the upper bound becomes $\widetilde{\mathcal{O}}(T \delta^2)$.}\footnote{This corner case rarely happens because it requires the generalized distance $\delta$ to be very small and the price variance $n\sigma^2$ to be very large, such that  there is no need of online learning. See the discussion in \S \ref{subsec-extension-lower bound}. 
   	} In addition, we prove an information-theoretic lower bound that matches the upper bound for both cases, showing that 
   	our regret bound cannot be further improved  (in a certain sense); we define such an unimprovable  regret bound   as the optimal (instance-dependent) regret for the \texttt{OPOD} problem in the  multiple-historical-price setting. 
   We  summarize its rate  in Table~\ref{table-multiple historical prices}.}

{\textbf{Sufficient condition for self-exploration.}  
As a byproduct,  
 we   provide a sufficient condition  for the \textit{myopic} (i.e., greedy) policy to self-explore   in the online learning  process. Specifically, if  the variance of historical prices is sufficiently large, and the average historical price is found to be bounded away from the confidence interval  for the optimal price constructed from   offline regression, then the myopic policy, the one that always charges the optimal price associated with the least-square estimate obtained in each round, achieves the optimal regret under mild conditions.  
 This result generates additional insights for the performance guarantee of the myopic policy with the help of offline data, and also provides analytical support for the wide use of such policies in practice.
}

 \textbf{Methodology contributions.}
 	From a technical perspective, the tight upper and lower bounds that we obtain in this paper are both  \textit{instance-dependent} regret bounds, which are much stronger and more challenging to prove than the traditional \textit{worst-case} regret bounds.   
 	  To prove the instance-dependent   upper bound, we conduct  a period-by-period trajectory analysis, and 
 	develop novel inductive arguments, integrated with the specific property guaranteed by OFU principle,  
   to obtain a sharp characterization on the distance between the algorithm's price and the average historical price.   
 	To prove the instance-dependent  lower bound, we reduce  the \texttt{OPOD} problem to a hybrid of estimation and hypothesis testing problems,  
 	which  requires constructing an instance-dependent prior distribution and an instance-dependent  hypothesis set, respectively. 
 	To the best of our knowledge, these are the first tight and general instance-dependent  regret bounds obtained in (i) the linear-demand online pricing problem, and (ii) a continuous-armed bandit problem where the optimal action may not be an extremal point (in contrast to the extremal-point requirement in \citealt{dani2008stochastic} and \citealt{abbasi2011improved}).

\subsection{Key Insights: Phase Transitions and Inverse-Square Law} The characterization  of the optimal  instance-dependent regret also leads to two important implications on the value of offline data. First,  when the offline sample size $n$ changes,  the optimal regret  rate  exhibits significantly  different decaying patterns,  and we refer to such significant   transitions between the regret-decaying patterns  as \textit{phase transitions}\footnote{We borrow this terminology from statistical physics, see \cite{domb2000phase}. See also the discussion of phase transitions in the optimal stopping problem studied by \cite{correa2018prophet}, and in the multi-armed bandit problem studied by   \cite{simchi2019phase}.}. 
		For example,  
		when  $\sigma=0$ and  $\delta \gtrsim T^{-\frac{1}{4}}$ (see Table 1), the  optimal regret rate remains at the level of $\widetilde \Theta(\sqrt{T})$ whenever $n \lesssim \frac{\sqrt T}{\delta^2}$, and then gradually decays   according to $  \widetilde \Theta(\frac{T}{n\delta^2})$ when $\frac{\sqrt T}{\delta^2} \lesssim n \lesssim T$, and finally stays at the level of $\widetilde \Theta(\frac{\log T}{\delta^2})$  when $n \gtrsim T$.  
		Second, in the regular case,  the optimal regret is inversely proportional to the square of the standard deviation $\sigma$ and generalized distance $\delta$, which is referred to as  the \textit{inverse-square law}. 
		The optimal regret's dependence on $\sigma$ is consistent with our intuition, as more dispersive historical prices indicate more information gained before the online learning  process starts, and therefore a smaller regret.  
		The optimal regret's dependence on $\delta$ is more intriguing, as it  suggests that the closer the historical prices are to the optimal price, the worse the optimal regret will be.   In fact,  this is a consequence of the tradeoff between exploration (i.e., experimenting to  improve  estimates of the unknown demand model) and exploitation (i.e., leveraging current estimates to maximize revenue). 
		Specifically, whenever an algorithm tries to learn the true demand model, it has to make substantial efforts in  charging various prices ``far away''  from the average historical price. Therefore, when $\delta$ is   small, such a deviation will also lead to a significant gap with the optimal price, leading to greater revenue loss.   These two findings contribute new insights to the fundamental problem of dynamic pricing with demand learning.

 \linespread{1.3}
\begin{table}[h] \caption{Optimal regret for the  single-historical-price setting}
	\label{table-general case}\vspace{.1in}
	\begin{center}
		\setlength{\tabcolsep}{2.5mm}{
			\begin{tabular}{l||c|c|c}
				\toprule 	\hline 
				\multicolumn{4}{l}{ $\delta \gtrsim T^{-\frac{1}{4}}$}                                                                                 \\ \hline %
				offline sample size        & $0\le n \lesssim \frac{\sqrt T}{\delta^{2}}$ & $\frac{\sqrt T}{\delta^{2}} \lesssim n \lesssim T$  & $n \gtrsim T$                        \\ \hline
				optimal regret  & $\widetilde \Theta(\sqrt T)$               & $\widetilde  \Theta(\frac{T}{n\delta^2})$                        & $\widetilde  \Theta(\frac{\log T}{\delta^2})$                       \\ \hline \hline 
				\multicolumn{4}{l}{ $\delta \lesssim T^{-\frac{1}{4}}$}                                                                                \\ \hline %
				offline sample size                                          & \multicolumn{3}{c}{$n \ge 0$} 
				 \\ \hline
				optimal regret & \multicolumn{3}{c}{$\widetilde\Theta(\sqrt T)$}                                                                                                               \\ \hline \bottomrule
		\end{tabular}}
	\end{center}
\end{table}

	\begin{table}[h] \caption{Optimal regret for the multiple-historical-price setting }
		\label{table-multiple historical prices}\vspace{.1in}
		\begin{center}
			\setlength{\tabcolsep}{4.85mm}{
				\begin{tabular}{l||c|c|c|c}
					\toprule 	\hline 
					
					\multicolumn{5}{l}{  $\delta \gtrsim T^{-\frac{1}{4}}$ and  $\sigma \lesssim  \delta $}                                                                                \\ \hline %
					offline sample size                                 & $0\le n \lesssim \frac{\sqrt{T}}{\delta^2}$   & $  \frac{\sqrt{T}}{\delta^2} \lesssim n \lesssim T$ & $T \lesssim  n  \lesssim \frac{T \delta^2}{\sigma^2}$ & $n \gtrsim \frac{T \delta^2}{\sigma^2}$ \\ \hline 
					optimal regret & $\widetilde\Theta(\sqrt T)$ & $\widetilde{\Theta}(\frac{T}{n\delta^2})$   &$\widetilde \Theta(\frac{1}{\delta^2})$    & $\widetilde \Theta(\frac{T}{n\sigma^2})$         \\ \hline
					\hline
					\multicolumn{5}{l}{$\delta \gtrsim T^{-\frac{1}{4}}$ and $\sigma \gtrsim \delta$}                                                                                \\ \hline %
					offline sample size       
					& \multicolumn{2}{c|}{$0 \le n \lesssim \frac{\sqrt T}{\sigma^2}$ } &  \multicolumn{2}{c}{$n \gtrsim \frac{\sqrt T}{\sigma^2}$}  \\  
					\hline 
					optimal regret 
					&\multicolumn{2}{c|}{$\widetilde\Theta(\sqrt T)$}  & \multicolumn{2}{c}{$\widetilde\Theta( \frac{T}{n\sigma^2})$}   \\
					\hline 
					\hline 
 					\multicolumn{5}{l}{ $\delta \lesssim T^{-\frac{1}{4}}$}                                                                                \\ \hline %
 					offline sample size       & \multicolumn{2}{c|}{$0 \le n \lesssim \frac{\sqrt T}{\sigma^2}$ } &  \multicolumn{1}{c|}{$ \frac{\sqrt T}{\sigma^2} \lesssim  n \lesssim \frac{1}{\delta^2\sigma^2}$}   &  \multicolumn{1}{c}{$  n \gtrsim \frac{1}{\delta^2\sigma^2}$}  \\  
 					\hline 
 					optimal regret  &\multicolumn{2}{c|}{$\widetilde\Theta(\sqrt T)$}  & \multicolumn{1}{c|}{$\widetilde\Theta(T\delta^2)$} &  \multicolumn{1}{c}{$\widetilde\Theta(  \frac{T}{n\sigma^2})$}    
					\\ \hline \bottomrule 
			\end{tabular}}
		\end{center}
\end{table}

\subsection{Structure and Notations}\label{subsec-structure}
This paper is organized as follows. In \S \ref{sec-lit}, we review the relevant literature.    
 In \S \ref{sec-UCB} and \S   \ref{sec-extension-historical price},  we study the \texttt{OPOD} problem in the single-historical-price setting  and multiple-historical-price setting respectively.  We conduct a numerical study in \S \ref{sec:num}, and discuss the self-exploration of the myopic policy in \S \ref{subsec-self exploration}. 
In \S \ref{sec-conclusion}, we summarize our paper with extensions and future research directions. Most of the technical proofs are deferred to  the appendix. 

Throughout the paper, all the vectors are  column vectors unless otherwise specified. For any $m \in \mathbb N$, we use $[m]$ to denote the set $\{1,2,\ldots, m\}$. For any column vector $x \in \mathbb{R}^n$ and positive semi-definite matrix $A \in \mathbb{R}^{n\times n}$,  $||x||:=(\sum_{i=1}^{n}x_i^2)^{\frac{1}{2}}$, and $||x||_A:=\sqrt{x^{\top}A x}$.  
	The  notations $\mathcal{O}(\cdot)$, $\Omega(\cdot)$ and $\Theta(\cdot)$ are applied to hide constant factors,  and $\widetilde{\mathcal{O}}(\cdot)$, $\widetilde \Omega(\cdot)$ and $\widetilde \Theta(\cdot)$ are applied to hide both constant   and  logarithmic factors. That is,  $f(T)=\mathcal{O}(g(T))$ means that  there exists a constant $C>0$ such that $f(T) \le Cg(T)$  for any $T$, and  $f(T)=\widetilde{\mathcal{O}}(g(T))$ means that there exist   constants $C$ and $\lambda>0$, such that  $f(T) \le Cg(T)(\log T)^{\lambda}$  for any $T$.   In addition, $f(T)= \Omega(g(T))$ (resp. $f(T)= \widetilde \Omega(g(T))$) means $g(T)= {\mathcal{O}}(f(T))$ (resp. $g(T)= \widetilde{\mathcal{O}}(f(T))$),  and $f(T)=\Theta(g(T))$ (resp. $f(T)=\widetilde{\Theta}(g(T))$)  means $f(T)=\mathcal{O}(g(T))$ and $f(T)=\Omega(g(T))$ (resp. $f(T)=\widetilde{\mathcal{O}}(g(T))$ and $f(T)=\widetilde{\Omega}(g(T))$).

\section{Related Literature}\label{sec-lit} 
\subsection{Dynamic Pricing with Online Learning}\label{subsec-online pricing} 
  When there are no offline data,  the \texttt{OPOD} problem becomes a pure online learning problem,  
  i.e. dynamic pricing with an unknown linear demand model, and  belongs to a broad category referred to as the  \textit{online pricing} problems.
Online pricing problems have generated great interest in  recent years in the operations research and management science (OR/MS)  community, see \cite{den2015dynamic} for a comprehensive survey. 
In particular, there is a vast literature (e.g.,  
\citealt{den2013simultaneously}, \citealt{den2014dynamic}, \citealt{KeskinZeevi2014},
\citealt{wang2014close}, \citealt{keskin2016chasing}, \citealt{qiang2016dynamic}, \citealt{den2017dynamic},  \citealt{nambiar2019dynamic}, \citealt{ban2020personalized}) studying  dynamic pricing problems with an unknown linear (or generalized linear) demand model, which is arguably one of the most fundamental demand models for pricing.    
All  of the existing papers purely focus on  online learning. 
In this paper, we take the fundamental problem of dynamic pricing with a linear demand model as our baseline,  but significantly extend it by incorporating   offline data into   online decision making.

  \cite{KeskinZeevi2014} is the  most relevant paper to this work. The authors consider dynamic pricing with an unknown linear demand model, 
studying an important question of how   knowing an \textit{exact} point at the demand curve (i.e., the exact expected demand under a single price) in advance helps  reduce the optimal regret. 
Depending on whether the seller knows this exact point or not, they prove that the best  achievable regret is $\Theta(\log T)$ and $\widetilde \Theta(\sqrt{T})$ respectively. Compared with their work, the \texttt{OPOD} problem studied in this paper seems more relevant to practice, and is more general in theory. Practically, while firms will never know the true expected demand under a given price exactly (which requires infinitely many demand observations), they usually have some pre-existing offline  data (which are finitely many) prior to the online learning process. Theoretically, the results in \cite{KeskinZeevi2014} (for the single-product setting) can be viewed as two special cases of our results when (i)  $n=0$; and (ii) $\sigma=0$, $n=\infty$, and $\delta=\Theta(1)$, with an additional   assumption that $\delta$ is lower bounded by a \textit{known} constant 
(as their algorithms for case (ii)  rely on this knowledge).  
Since $\delta$ is completely unknown and can be small  in our setting (and in reality), their algorithms and analysis do not apply here. In fact, the principle of our algorithm  design and the approach of our regret analysis  are very different from theirs.

There is also a stream of literature in Bayesian learning, 
	where 
	the decision maker is assumed to have a   known prior  distribution for  the  unknown   parameter, and  can update her belief on the prior distribution from  online observations. For recent works on  dynamic pricing with Bayesian learning,  we  refer the interested readers to   \cite{harrison2012bayesian} and  \cite{agrawal2017thompson} that focus on the worst-case regret,  and to  \cite{ferreira2018online} and  \cite{miao2020dynamic} that focus on the Bayesian regret.  
 While the	prior distribution in  Bayesian learning can be estimated using offline data,  the modeling approach and results of these  papers are very different from this work.  
First, in Bayesian learning, it is usually assumed
that the decision maker knows the exact prior distribution, which typically belongs to some  specific parametric family.  
By
contrast, in this work, we do not assume any prior distribution or impose any parametric assumption on the distribution of   demand parameter, but directly incorporate offline data into online learning.  
Second, as a main contribution of this
paper, we characterize the effects of the size, dispersion and location of the offline data on the
statistical complexity of online learning, which are not discussed in and not the focus of the current literature on Bayesian learning.

\subsection{Multi-Armed Bandits}\label{subsec-literature}
 Our paper is also related to the literature of  multi-armed bandits (MAB). In the classical $K$-armed bandit problem, the decision maker chooses one of the $K$ arms in each round and observes a random reward generated from some unknown distribution associated with the arm being played, with the  goal of minimizing the regret, see \cite{lattimore2018bandit} for more references on this topic. 
In most of the literature on bandit problems (see, e.g.,  \citealt{auer2002finite},  \citealt{dani2008stochastic}, \citealt{rusmevichientong2010linearly}, \citealt{abbasi2011improved},  \citealt{filippi2010parametric}), the decision maker has to start from scratch (i.e., with no historical information).  By contrast,  a few papers study bandit problems in  settings where the algorithms may utilize different types of historical information, see, e.g., \cite{shivaswamy2012multi},  \cite{bouneffouf2019optimal}, \cite{bastani2019meta}, \cite{hsu2019empirical}, \cite{gur2019adaptive},   \cite{ye2020combining}, of which \cite{shivaswamy2012multi}  and \cite{gur2019adaptive} are the most relevant to this paper. 

\cite{shivaswamy2012multi}  study the MAB problem with offline observations of rewards  collected before the online learning algorithm starts.  While  our idea of  incorporating offline data into an online learning problem is similar to theirs, 
 there are significant differences between the two papers in terms of model settings, main results and analytical techniques. First,  \cite{shivaswamy2012multi} study the MAB problem with discrete and finitely many arms, while our model builds on  the literature of online pricing problems (see \S \ref{subsec-online pricing}  for references),  
where the prices are  continuous and infinitely many, and the rewards are nonlinear with respect to prices. The properties and results for these two classes of problems are very different.  
Second, under the 
\textit{well-separated condition}, 
\cite{shivaswamy2012multi} prove  some regret upper bounds that change from $\mathcal{O}(\log T)$ to $\mathcal{O}(1)$ when the amount of offline observations of rewards for \textit{each} arm exceeds $\Omega(\log T)$, with no regret lower bound proven and  hence no discussion of phase transitions. In comparison, we characterize the optimal regret via  \text{matching} upper and lower bounds, and figure out surprising phase transitions of the optimal regret rate as the offline sample size changes. Moreover, we also discover the inverse square law, which does not appear in the previous literature. Third,  while  \cite{shivaswamy2012multi}  use a conventional approach in bandit literature to upper-bound the regret via the so-called \textit{sub-optimality gap}, since we are bounding the regret via $\sigma$ and $\delta$, we present   different regret analysis that may be of independent interest. 

In a recent paper by \cite{gur2019adaptive}, a generalized MAB formulation is studied, where some additional information may become available before each online decision is made.
Under   the {well-separated} condition, the authors characterize the optimal regret as a function of the information arrival process, and study the effect of the characteristics of this process on the algorithm design and the best achievable regret bound.   In particular, their results include  the MAB with offline data as a special case. 
 Although our paper shares similar spirits with \cite{gur2019adaptive} in the focus of  identifying key characteristics of some ``additional'' information that affect the optimal regret, and quantifying the magnitude of such effects,  the model settings, results and insights in these two papers are very different.

Interestingly, although neither  \cite{shivaswamy2012multi} nor \cite{gur2019adaptive} makes an attempt to characterize the optimal regret for the MAB with offline data under general case (i.e.,  when the well-separated condition does not necessarily hold), or discuss the   phase transitions, combining the regret upper bound in \cite{shivaswamy2012multi} with the regret lower bound in \cite{gur2019adaptive}  gives a  characterization on the optimal regret for the MAB with offline data  under some mild conditions, which also leads to phase transitions not discussed before.  We provide more   discussions on this finding in  Appendix G.

 		\section{Single Historical Price}\label{sec-UCB}
 		In this section, we study the single-historical-price setting: 
 		 where all the $n$ historical prices are identical  to $\hat p$. As pointed out in \cite{harrison2012bayesian} and  \cite{KeskinZeevi2014}, in  finance industry,   for many consumer lending products,  banks often keep a fixed interest rate over some  periods of time before they conduct price experimentation. 
 Similarly,  in the retail industry, 
	there are many scenarios where the seller charges a fixed price based on  the manufacturer's suggestion, branding or competitors' price before using a dynamic pricing strategy.  
 		 	Thus, we start with  this simple but important single-historical-price setting in this section.  
 		 We first  design a  learning algorithm with a per-instance regret upper bound in \S \ref{subsec-O3FU}, and then  characterize the regret lower bound in \S \ref{sec-lower bound}.   Some important  implications are discussed in \S  \ref{sec-single-implication}.

 	\subsection{ O3FU Algorithm and Regret Upper Bound}\label{subsec-O3FU}

 	Our proposed algorithm  \textit{Online and Offline--Optimism in the Face of Uncertainty} (O3FU) is  constructed  based on the celebrated  \textit{Optimism
 		in the Face of Uncertainty} (OFU) principle, which effectively addresses the exploration-exploitation dilemma inherent in many online learning problems (see, e.g., \S7.1 of \citealt{lattimore2018bandit} for a reference). 
 	 For any $t \ge 0$, we define a confidence radius $w_t$   that will be used to construct a confidence ellipsoid for the  demand parameter at the end of period $t$, and the expression of $w_t$ is as follows: 
 	\begin{align}\label{OFU-radius}
 	w_t=R\sqrt{
 			2\log \Big(\frac{1}{\epsilon}\big(1+(1+u^2)(t+n)/\lambda\big)\Big)}+\sqrt{\lambda (\alpha_{\max}^2+\beta_{\min}^2)},
 	\end{align}  
 	where   $\epsilon$ and $\lambda$ will be specified in the description of the algorithm. 
 The choice of $w_t$ is based on the high-probability confidence bound developed in  Theorem 2 of \cite{abbasi2011improved}, which will also be used throughout our regret analysis. 	The pseudo-code of O3FU algorithm is provided  in Algorithm~\ref{Alg-UCB}.

 	\begin{algorithm}[htbp] 	\label{Alg-UCB}
 		\caption{O3FU Algorithm}
 		\textbf{Input}: historical price $\hat p$, offline demand data $\hat D_1, \hat D_2, \ldots, \hat D_n$, support of unknown parameters $\Theta^{\dag}$,  price range $[l,u]$,   regularization parameter $\lambda = 1+u^2$, $\{w_t\}_{t\ge 1}$ defined in \eqref{OFU-radius} with $\epsilon=\frac{1}{T^2}$\; %
 		\textbf{Initialization}: $V_{0,n}=\lambda I+n[1 \, \, \hat p]^{\mathsf{T}}[1 \, \, \hat p]$, $Y_{0,n}=(\sum_{i=1}^{n}\hat D_i )[1 \,\, \hat p]^{\mathsf{T}}$\;
 		\For{$t \in [T]$}{
 			\eIf{$t=1$}{
 				Charge   price $p_1=l\cdot\mathbb{I}\{\hat{p}>\frac{u+l}{2}\}+u\cdot\mathbb{I}\{\hat{p}\le\frac{u+l}{2}\}$, and observe demand realization $D_1$\; 
 				Compute $V_{1,n}=V_{0,n}+[1 \, \, p_1]^{\mathsf{T}}[1 \, \,  p_1]$, $Y_{1,n}=Y_{0,n}+D_1[1 \,\, p_1]^{\mathsf{T}}$, $\hat \theta_1=V_{1,n}^{-1}Y_{1,n}$\;
 				Compute   confidence ellipsoid $
 				\mathcal C_1=\big\{\theta \in \mathbb{R}^2: ||\theta-\hat \theta_1||_{V_{1,n}} \le w_1\big\}$\;
 				
 			}{
 				If $\mathcal C_{t-1} \cap \Theta^{\dag} \ne \emptyset$, let  $
 				(p_t,\tilde \theta_t) \in \argmax_{p \in [l,u], \theta \in \mathcal C_{t-1} \cap \Theta^{\dag}} p(\alpha+\beta p) $; otherwise, let $p_t=p_{1}$\; 
 				Charge   price $p_t$, and observe demand realization $D_t$\;
 				Update $V_{t,n}=V_{t-1,n}+[1 \,\, p_t]^{\mathsf{T}}[1 \,\, p_t]$,  $Y_{t,n}=Y_{t-1,n}+D_t[1 \,\, p_t]^{\mathsf{T}}$, $\hat \theta_t=V_{t,n}^{-1}Y_{t,n}$\;
 				Update    confidence ellipsoid $\mathcal C_t=\big\{\theta  \in \mathbb{R}^2: ||\theta -\hat \theta_t||_{V_{t,n}} \le w_t\big\}$. 
 			} 
 		}
 	\end{algorithm}

\begin{figure}[htbp!]
 		\centering
 		\includegraphics[width=0.5\textwidth]{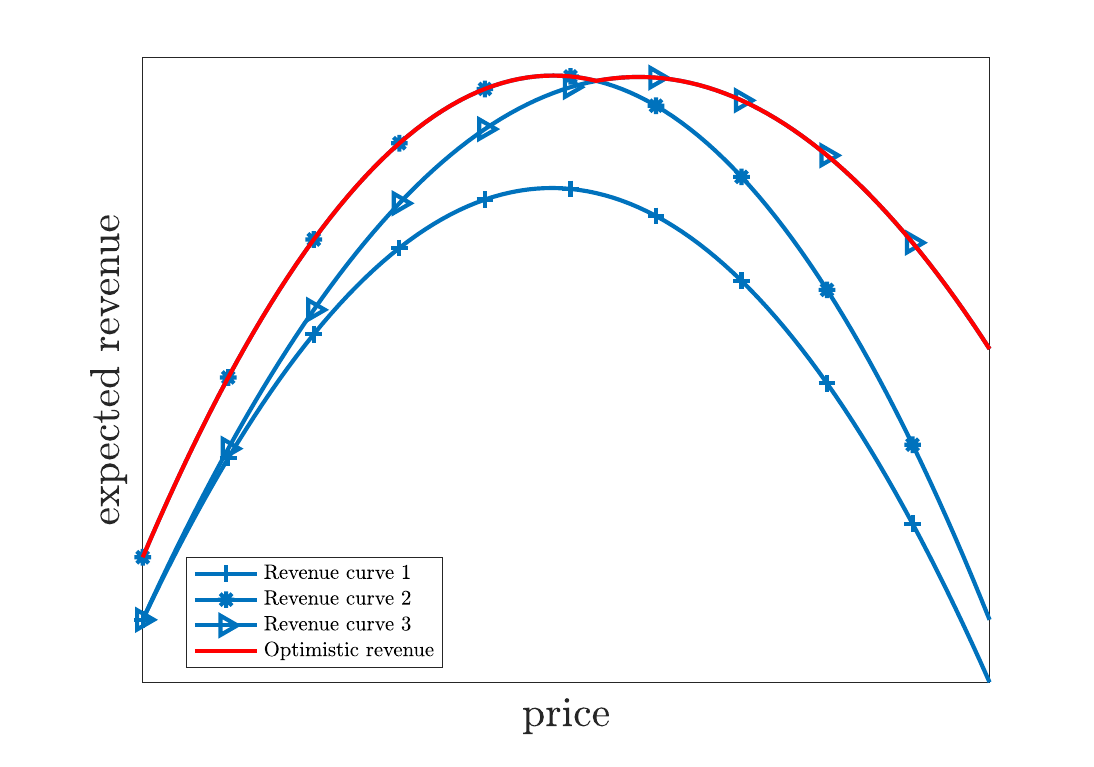}
 		\caption{Revenue curves under three  different parameters (blue), and the optimistic revenue (red)}
 		\label{O3FU-envelope}
 	\end{figure}

 	In   O3FU algorithm,
when $t=1$,  the   price is chosen from   boundary points $\{l,u\}$, depending on which one has a  larger  distance from   historical price $\hat p$.   The choice of such an initial price   is  not unique, and any price that is bounded away from $\hat p$ by a constant distance will also work.  
 For  each   $t\ge 2$, we first maintain a confidence ellipsoid $\mathcal C_{t-1}$ for the unknown parameter $\theta^*$, and  then  O3FU algorithm   selects  an optimistic estimator $\tilde \theta_t \in  \argmax_{\theta \in \mathcal{C}_{t-1} \cap \Theta^{\dag}}\max_{p \in [l,u]} p(\alpha+\beta p)$, and charges   price $p_t = \argmax_{p \in [l,u]}p(\tilde\alpha_t+\tilde \beta_t p)$, which is optimal with respect to    estimator $\tilde \theta_t$. 
 Note that when $\max_{p \in [l,u]} p(\alpha+\beta p)$, as a function of  $ \theta \in \mathcal  C_{t-1} \cap \Theta^{\dag}$,  has multiple maximizers, $\tilde \theta_t$ can be set as any maximizer.  Figure \ref{O3FU-envelope} shows how O3FU algorithm works, where the three blue curves depict the expected revenues with three different  parameters belonging to  set $\mathcal C_{t-1} \cap \Theta^{\dag}$ (we only draw three curves for illustration), and the red curve is the upper envelope of all the possible candidate revenue curves, which is also the revenue function associated with the demand parameter $\tilde \theta_t$, i.e., $r(p;\tilde \theta_t)$.

 Intuitively, if we knew that  generalized distance $\delta$ would be large, then trying prices far away from $\hat p$ is beneficial for both exploration and exploitation. By contrast, if we knew that $\delta$ would be small, then striking a  balance between  exploration and exploitation  would be very important, because choosing prices close to $\hat p$ is only effective for exploitation  but not for exploration.  Of course, the seller does not know the true value of $\delta$, which makes designing a learning algorithm that achieves the right balance between exploration and exploitation a  more challenging task. 
 O3FU algorithm achieves this objective  
  by maximizing the  \textit{optimistic revenue}, which is defined as  $\max_{ \theta \in \mathcal  C_{t-1} \cap \Theta^{\dag}} p(\alpha+\beta p)$, and 
can be treated as the  estimated revenue plus a ``bonus'' of exploration.  
In fact, as implied from equation (19.8) in \cite{lattimore2018bandit},  when $\mathcal{C}_{t-1} \subseteq \Theta^{\dag}$, we have $\max_{ \theta \in \mathcal  C_{t-1} \cap \Theta^{\dag}} p(\alpha+\beta p) = p(\hat  \alpha_{t-1}+\hat \beta_{t-1}p)+w_{t-1}\sqrt{[1 \, \, p]V_{t-1,n}^{-1}[1 \, \, p]^{\top}}$. Therefore, 
exploitation and exploration are both incorporated into the objective function  through the first term and the second term, respectively.

It's worth noting that our O3FU algorithm is \textit{parameter-free} in the sense that it does not need to use any information about  $\delta$. In addition,   while O3FU algorithm  takes $T$ as input, one  can easily extend the current algorithm to work with unknown $T$ using the standard \emph{doubling trick} (see, e.g.,  \citealt{lattimore2018bandit}) and construct an  \textit{anytime} algorithm that does not need to know~$T$.

 We now provide an upper bound on the regret of O3FU algorithm.  
 	\begin{theorem}\label{thm-UCB}  
 		Let $\pi$  be O3FU algorithm. Then there exists a finite   constant $K_1>0$ such that  for any   $T \ge 1$,  $n\ge 0$ and $\hat p \in [l,u]$, and for any possible value of  $\theta^* \in \Theta^{\dag}$, 
 		\begin{align*}
 		R^{\pi}_{\theta^*}(T) \le    K_1 \cdot \Big( \sqrt T \wedge \frac{T\log T}{(n \wedge T)\delta^2}\Big)\cdot  \log T.  
 		\end{align*} 
 	\end{theorem}

 	 	Theorem \ref{thm-UCB} provides a regret upper bound $\widetilde{\mathcal{O}}(\sqrt T \wedge \frac{T\log T}{(n \wedge T)\delta^2})$  that depends on the  problem instance through the value of  $\delta$, which is therefore called 
the	\textit{instance-dependent} upper bound.  
 	If
 	$\delta$ is a constant, when $n=0$ or  $n=\infty$, i.e., there are no offline data or infinitely many offline data under price $\hat p$, the upper bound reduces to $\widetilde{\mathcal{O}}(\sqrt{T})$ and $\widetilde{\mathcal{O}}(\log T)$  respectively. If $\delta$ is not a constant,  with an order shrinking to zero as $T$ grows, the regret upper bound is then inversely proportional to $\delta^2$.  
 We   summarize   the regret upper bound under different $(n,\delta)$ combinations in  Table \ref{table-general case-upper bound-single} of  Appendix~H.

 We next outline the key ideas to prove Theorem \ref{thm-UCB} and leave the detailed analysis to  Appendix~A.1. From the statement of Theorem \ref{thm-UCB}, it suffices to show an \textit{instance-independent} upper bound $\widetilde{\mathcal{O}}(\sqrt T)$ and an \textit{instance-dependent} upper bound $\widetilde{\mathcal{O}}(\frac{T}{(n \wedge T)\delta^2})$.  
	The  instance-independent   bound can be proved using  similar arguments from stochastic linear bandits, e.g., \cite{abbasi2011improved}, by noting that  the expected revenue is the  inner product of the unknown parameter  $[\alpha \, \, \beta]$  and the action vector $ [p\, \, p^2]$.  
	Showing the  instance-dependent   bound is the novel   part in our proof, which relies on  the   following crucial lemma.  
	
		\begin{lemma} \label{lemma-proof-UCB-step2}
	Suppose $T \ge T_0$, $\delta \ge \frac{\sqrt{2(\alpha_{\max}^2+\beta_{\max}^2)}w_T}{\beta_{\max}^2n^{1/4}}$, and $\theta^* \in \mathcal C_t$ for each $t \in [T]$, then two sequences of events $\{U_{t,1}\}_{t=1}^T$ and $\{U_{t,2}\}_{t=2}^T$   also hold, where 
		\begin{align*}
	&U_{t,1}=\Big\{|p_t -\hat p| \ge \min\Big\{1-\frac{\sqrt 2}{2},\, \frac{C_0}{2}\Big\} \cdot \delta\Big\},\\
		&U_{t,2}=\Big\{||\tilde \theta_t-\theta^*||^2 \le  C_2\cdot \frac{w_{t-1}^2}{(n \wedge (t-1))\delta^2}\Big\},  
		\end{align*}  
	and  $C_0= \frac{l|\beta_{\max}|}{u |\beta_{\min}|}$,  $C_1 = 4(C_0+1)^2C_0^{-2} \big(1+(4u+1)^2\big)$, $C_2=\max\Big\{4(u-l)^2, \,2C_1,\, 4((4u+1)^2+1)(\min\{\frac{C_0^2}{4}, (1-\frac{\sqrt 2}{2})^2\})^{-1}\Big\}$, $T_0=\min\Big\{t \in \mathbb{N}: w_{t} \ge  \sqrt C_1 \beta_{\max}^2(2(\alpha_{\max}^2+\beta_{\max}^2))^{-1/2} \Big\}$.
	\end{lemma}

Lemma \ref{lemma-proof-UCB-step2} is interpreted as follows.  When   the   optimal price has a certain distance from historical price $\hat p$, i.e., $\delta \ge \frac{\sqrt{2(\alpha_{\max}^2+\beta_{\max}^2)}w_T}{\beta_{\max}^2n^{1/4}}$,   given that the    demand parameter $\theta^*$ belongs to the   confidence ellipsoid $\mathcal{C}_t$ in each period $t$, 
  the algorithm's pricing sequence  $\{p_t\}_{t=1}^T$ is also  uniformly bounded away from  $\hat p$  proportional to the unknown quantity $\delta$ (as implied by   events $\{U_{t,1}\}_{t=1}^T$), and will gradually approach the true  optimal price   in a  rate of  $\mathcal{O}(\frac{w_t^2}{(n\wedge t)\delta^2})$ (as implied by    events $\{U_{t,2}\}_{t=2}^T$). 
This implies that the algorithm can ``adaptively'' explore to a suitable degree, to create an efficient ``collaboration'' between the online prices and the historical price, while concurrently approaching the unknown optimal price. 
This property is nontrivial and cannot be implied from the existing analysis of the OFU-type algorithms. To prove this   lemma, we conduct a period-by-period trajectory analysis of the random pricing sequence generated by our algorithm. Specifically, we find that the occurrence of $U_{t,2}$ relies on the joint occurrence of $U_{1,1}, \ldots, U_{t-1,1}$, while the occurrence of $U_{t,2}$ (combined with the specific structure of the optimistic revenue curve) in turn leads to the occurrence  of $U_{t,1}$. We thus introduce novel induction-based arguments to prove Lemma \ref{lemma-proof-UCB-step2}, see details in Appendix A.2.   The induction-based arguments also explain why we set the initial price in the algorithm  to be a boundary point (or any price that has a constant distance from $\hat p$), since this enables  $U_{1,1}$ to occur.

	We remark that for the stochastic linear bandit problem with  a polytope action set, \cite{abbasi2011improved} prove  an instance-dependent upper bound of $\mathcal{O}(\frac{\log T}{\Delta})$, where  $\Delta$ is defined as  the sub-optimality gap between the rewards of the best and second best extremal points of the action set. We emphasize that their result and analysis  cannot be applied to prove our instance-dependent upper bound due to the following reasons. First,  the instance-dependent upper bound in our problem is developed to capture the effect of the generalized distance $\delta$ on the regret bound, which does not exist in the stochastic linear bandit problem. Second,  the instance-dependent upper bound  in \citet{abbasi2011improved} relies on two strong conditions: (i) their algorithm only selects actions among the extremal points of the action set, and (ii) every sub-optimal action taken by their algorithm is bounded away from the optimal action by a reward gap $\Delta$. Such conditions only hold under their setting and assumptions.  
Our problem, however,  has a quadratic objective function, with the optimal price  being an interior point of the interval $[l,u]$, which requires the algorithm's actions to converge to the optimal action. As a result, the sub-optimality gap $\Delta$ becomes zero, and standard arguments based on $\Delta$ do  not work.

   \subsection{Lower Bound on   Regret} \label{sec-lower bound} 
   In this subsection, we establish a lower bound on the performance of any algorithm  for  the \texttt{OPOD} problem with a single historical price.  
   We first introduce the following set of \textit{admissible  policies}  denoted by $ \Pi^{\circ}$, which includes   all the  policies
   	whose regret is guaranteed to be $\widetilde{\mathcal{O}}(\sqrt T)$ for  any possible value of demand parameter $\theta^*$, i.e., 
   	\begin{align} \label{def-minimax optimal policy}
   	\Pi^{\circ} = \Big\{\pi \in \Pi: \,  \sup_{ \theta^* \in \Theta^{\dag}}  R^{\pi}_{\theta^*}(T) \le K_0\sqrt T (\log T)^{\lambda_0}\Big\},
   	\end{align}
   where $K_0>0$ and $\lambda_0  \ge 0$ are arbitrary constants.  Intuitively,  $\Pi^{\circ}$ excludes those ``bad'' policies that are  not robust and suffer from large worst-case regret, e.g., a policy that never explores and always chooses $\hat p$, incurring zero regret when $\delta=0$ but linear  regret when $\delta=\Theta(1)$. Restricting our attention to  $\Pi^{\circ}$ (which O3FU and many existing algorithms obviously belong to) ensures that the considered policies are   reasonable enough. Note that  $\Pi^{\circ}$ is specified by a pair of $(K_0, \lambda_0)$, but for  simplicity, when there is no ambiguity, we drop the dependence on $(K_0, \lambda_0)$ in the notation.  To facilitate our discussion,    let $R_{\theta}^{\pi}(T, n, \delta)$ be defined as the regret for admissible policy $\pi \in \Pi^{\circ}$ when the   demand parameter is $\theta=(\alpha, \beta)$, i.e., $R^{\pi}_{\theta}(T, n, \delta) =T\cdot r^*(\theta)- \mathbb{E}^{\pi}_{\theta}[\sum_{t=1}^{T} r(p_t; \theta)]$. We also denote $\mathcal{D}$ as the generic distribution of $\{\hat \varepsilon_i\}_{i=1}^n$ and $\{\varepsilon_t\}_{t=1}^T$, and $\mathcal{E}(R)$ as the class of sub-Gaussian distributions with parameter $R$.

 The following theorem provides a regret lower bound for any admissible policy in terms of the generalized distance $\delta$. For any generalized distance $\delta$, we define an \emph{instance-dependent} environment class $\{\theta\in \Theta^{\dag}: |\hat p-\psi(\theta)| \in [(1-\xi)\delta, (1+\xi)\delta]\}$, which is the set of all possible values of  the demand parameter  whose associated optimal prices are $\Theta(\delta)$-distance away from $\hat{p}$  (here $\xi$ can be any fixed constant in $(0,1)$). This environment class highlights the role of $\delta$ as a key instance-dependent quantity, and enables us to establish an instance-dependent regret lower bound that holds for all possible values of $\delta$; see Theorem \ref{thm-lower bound} (note that the environment class appears under the $\sup$ operator in the LHS of (\ref{ineq-lower bound})).

   \begin{theorem} \label{thm-lower bound}   
   	There exists a  positive constant $K_2$ such that 	  
   	for any  admissible policy $\pi\in\Pi^{\circ}$, for any $\xi\in (0, 1)$, $T\ge 2$ and $n\ge 0$, and for any $\delta \in [0, u-l]$,  
   	\begin{align} \label{ineq-lower bound} 
  \sup_{\mathcal{D} \in \mathcal{E}(R);  \atop \theta\in \Theta^{\dag}: |\hat p-\psi(\theta)| \in [(1-\xi)\delta, (1+\xi)\delta]} 	R^{\pi}_{\theta}(T, n, \delta )\ge  
   	\left\{
   	\begin{array}{ll}
   	K_2 \cdot  \left( (\sqrt T\wedge \frac{T}{(n \wedge T) \delta^2})\vee \log T\right), & \text{if } \delta \gtrsim T^{-\frac{1}{4}};\\
   	K_2 \cdot \left((T\delta^2) \vee \frac{\sqrt T}{(\log T)^{\lambda_0}} \right), & \text{if } \delta \lesssim T^{-\frac{1}{4}}.
   	\end{array}  
   	\right.
   	\end{align}   
   \end{theorem}

 \begin{remark}\label{rmk-en vironment class}
We emphasize that finding a ``right'' definition of the instance-dependent environment class is important for capturing the true role of $\delta$ in determining the instance-dependent regret. While there may be other ways to specify the environment class, they may fail to accurately reflect the instance-dependent complexity of the \texttt{OPOD} problem. For example, if one sets the environment class to be the entire parameter space $\Theta^\dagger$, then one can obtain a single lower bound for the worst-case regret (independent of $\delta$); however, such a definition is too conservative and does not fully capture the value of offline data. Another seemingly natural way to specify the environment class is to consider  $\{\theta \in \Theta^{\dag}: |\psi(\theta)-\hat p |  =\delta\}$, which is the set of all possible values of  the demand parameter whose associated  optimal price has a distance from $\hat p$ exactly equal to $\delta$.  However, this definition cannot preclude certain speculative behavior of algorithms, and would result in an unrealistic regret bound that cannot be attained by any single algorithm. We refer to Appendix~D for more details regarding the limitations of the above two definitions of the environment class.  
\end{remark}

 We explain Theorem \ref{thm-lower bound}  as follows. 
   	First,  
   	when $\delta \gtrsim T^{-\frac{1}{4}}$, the regret lower bound is $\Omega((\sqrt T \wedge \frac{T}{(n \wedge T)\delta^2})\vee \log T)$,    
   	and in particular, if $ \delta$ is   a  constant and $n=\infty$,  the regret lower bound reduces to  
   	$\Omega(\log T)$,  which recovers Theorem~3  in   \cite{KeskinZeevi2014} for their incumbent-price setting.   
   	Second, when  
   	$\delta \lesssim T^{-\frac{1}{4}}$,  the  regret lower bound is  always $\widetilde \Omega(\sqrt T)$, regardless of   offline sample size $n$. The intuition is as follows. 
   	When restricting attention to  $\Pi^{\circ}$, we exclude those ``unreasonable''  policies that seldom explore but make   pricing decisions in a naive way, e.g., the one that always chooses price  $\hat{p}+\delta$,  because the regret of such policies cannot always  be upper bounded by $\widetilde{\mathcal{O}}(\sqrt T)$ for any  possible value of  $\theta^*$. In this case, any admissible policy $\pi \in \Pi^{\circ}$ should be able to make sufficient exploration to  distinguish between     different demand curves. However, to achieve this, the policy must deviate from   $\hat p$, which is \textit{less informative}  since the seller already has collected  some data under  this  price, to gain more information about the true demand curve. When $\delta$ is very small, charging prices away from  $\hat p$  
   	leads to a significant gap relative to the optimal price, and therefore   a large regret bound.  
   We  summarize  the  regret lower bound under different $(n,\delta)$ combinations in Table \ref{table-general case-lower bound-single} of Appendix~H.

   We next highlight the key steps  in proving Theorem  \ref{thm-lower bound} and leave the detailed analysis to Appendix~A.3.   
   The proof idea is to reduce the \texttt{OPOD} problem to a hybrid of an estimation problem (see Step 1) and a hypothesis testing problem  
   (see Step 2). 
   
   \underline{\textbf{Step 1.}}
   In this step, we prove that when $\varepsilon$ follows a  normal distribution, for any pricing policy $\pi \in \Pi$ (not necessarily in $\Pi^{\circ}$),   
   \begin{align} \label{ineq2-proof-thm1}
   \sup_{\theta \in \Theta^{\dag}: |\hat p-\psi(\theta)| \in  [(1-\xi)\delta, (1+\xi)\delta]} R^{\pi}_{\theta}(T, n, \delta) =\Omega\Big( \big(\sqrt T \wedge  \big(\frac{T}{\delta^{-2}+(n \wedge T)\delta^2}\big)\big)\vee \log (1+T\delta^{2})\Big). 
   \end{align}   
   {To prove \eqref{ineq2-proof-thm1}, we consider an ``auxiliary'' estimation problem for the optimal price $\psi(\theta)$, and appeal to the multivariate van Trees inequality (cf. \citealt{gill2001applications}) to construct a lower bound for the Bayesian regret.  
   	 In particular, when applying the   van Trees inequality, we need to carefully choose a  suitable instance-dependent prior distribution $q(\cdot)$  whose  Fisher information grows at an appropriate rate with respect to $\delta$,  and upper-bound the resulting Fisher information of the sequential estimators $\{p_t\}_{t=1}^T$ in different cases. Then we can rightly control the growth rate of the Bayesian regret.} 
   
   \underline{\textbf{Step 2.}}
   In the second step, we  show that when  $\varepsilon$ follows a   normal distribution and $\delta \lesssim T^{-\frac{1}{4}}(\log T)^{-\frac{1}{2}\lambda_0}$,  for any admissible policy $\pi \in \Pi^{\circ}$, there exists    $\theta \in \Theta^{\dag}$ satisfying $|\psi(\theta)-\hat p| \in [(1-\xi)\delta, (1+\xi)\delta]$ such that 
   \begin{align} \label{ineq1-thm1}
   R^{\pi}_{\theta}(T, n, \delta) = \Omega\Big(\frac{\sqrt T}{(\log T)^{\lambda_0}}\Big). 
   \end{align}   
The proof of \eqref{ineq1-thm1} is based on arguments using Kullback-Leibler divergence and Bretagnolle-Huber inequality  (Theorem 2.2 in \citealt{Tsybakov2009}), whose key idea is as follows.  We  construct two  problem instances with parameters $\theta_1$ and $\theta_2$   such that  (i) the two demand curves under $\theta_1$ and $\theta_2$ intersect at price $\hat p$; (ii)  the   optimal prices under $\theta_1$ and $\theta_2$ are $\hat p+\delta$ and $\hat p+\delta+\Delta$ respectively, with  $\Delta = \Theta(T^{-\frac{1}{4}} (\log T)^{\frac{1}{2}\lambda_0})$. For  any  pricing policy $\pi \in \Pi^{\circ}$, it has to perform well under both constructed problem instances, i.e.,   the regret upper bound is $\mathcal{\widetilde{O}}(\sqrt T)$ under either instance, and therefore should be able to    distinguish between the demand environments under $\theta_1$ and $\theta_2$. Moreover, any   policy with the goal of separating  $\theta_1$ and $\theta_2$ should charge prices significantly different from the intersected price $\hat p$, i.e., the KL-divergence between the two probability measures under $\theta_1$ and $\theta_2$ induced by policy $\pi$   is large. Nevertheless, since the optimal price associated with  $\theta_1$ is $\hat p+\delta$, which is extremely close to $\hat p$, 
		 the policy will incur large regret when the underlying
		parameter is in fact $\theta_1$. Therefore, the regret under $\theta_1$ is always lower bounded by  $\Omega(\frac{\sqrt T}{(\log T)^{\lambda_0}})$ no matter how large $n$ is.

 	\subsection{Phase Transitions  and   Inverse-Square Law} \label{sec-single-implication}
 	In this subsection, we discuss two important implications. 
 	By comparing Theorems~\ref{thm-UCB} and   \ref{thm-lower bound}, one can easily verify that  
 		the regret upper bound $\widetilde{\mathcal{O}}(\sqrt T \wedge \frac{T\log T}{(n \wedge T)\delta^2})$ achieved by O3FU algorithm, after ignoring the logarithm factor,  is unimprovable   within the class of all admissible policies
 		under the instance-dependent   environment class considered in Theorem~\ref{thm-lower bound}.  
 		 Motivated by this result, for $\Pi^{\circ}$ with $\lambda_0\ge 1$, 
 		 we define the \textit{optimal (instance-dependent)  regret} $R^*(T, n, \delta)$ as   
 		\begin{align}\label{def1-optimal regret}
 		R^*(T, n,\delta) = \inf_{\pi \in \Pi^{\circ}} \sup_{\mathcal{D} \in \mathcal{E}(R);  \atop \theta \in \Theta^{\dag}: |\psi(\theta)-\hat p| \in [(1-\xi)\delta, (1+\xi)\delta]} R^{\pi}_{\theta}(T).
 		\end{align}  
 		Thus, 
 		$R^*(T, n,\delta)$ characterizes the   statistical complexity of the \texttt{OPOD} problem in the sense that no algorithm in the admissible policy class can perform better than this rate when the true optimal price is allowed to center around  $\hat p$ within $\Theta(\delta)$. We state this result   in the following corollary.

 	\begin{corollary} \label{coro-opt regret-single}
 		The optimal regret defined in \eqref{def1-optimal regret}  for the single-historical-price setting is
 		\begin{align*}
 		R^*(T,n, \delta) =\widetilde{\Theta}\left(\sqrt{T}\wedge \left(\frac{T}{n\delta^2} \vee \frac{\log T}{\delta^2}\right)\right). 
 		\end{align*}
 	\end{corollary}

\begin{figure}[htbp!]
 		\centering
 		\includegraphics[width=0.4\textwidth]{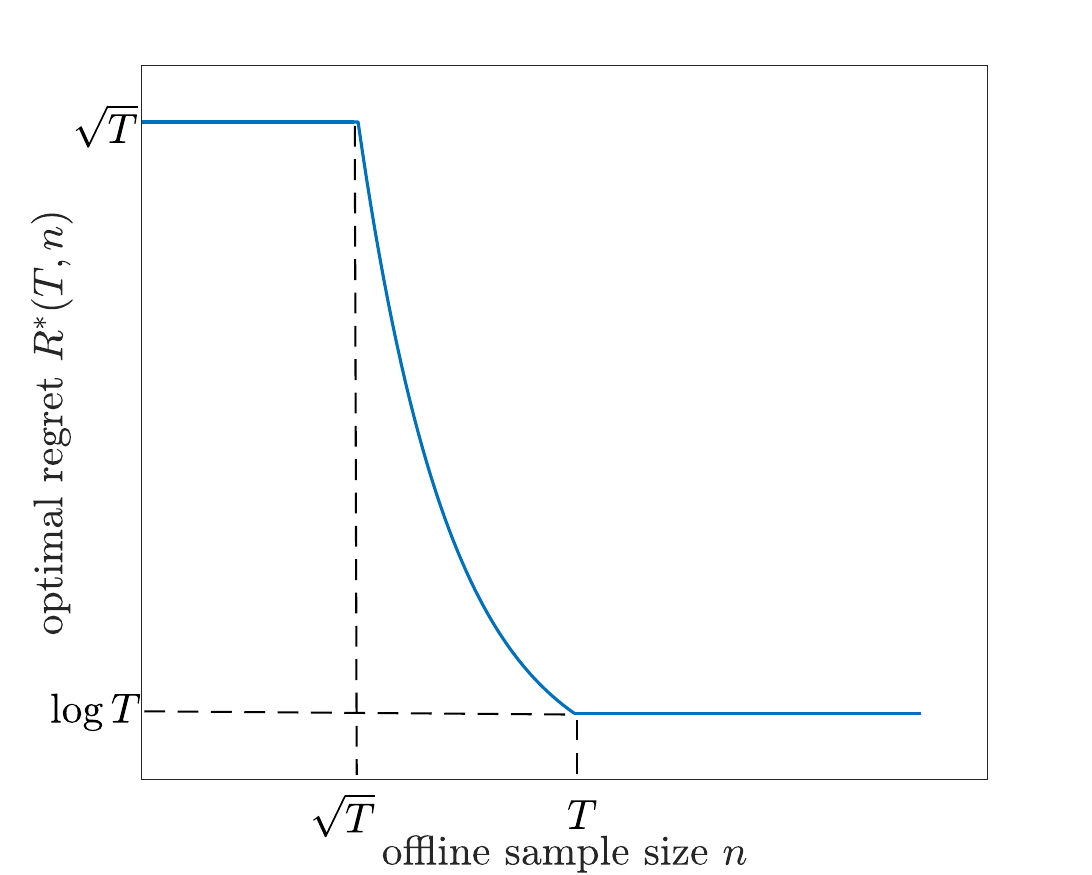}
 		\caption{Phase transitions for the single-historical-price setting with constant $\delta$}
 		\label{figure1}
 	\end{figure}

 The characterization of the optimal regret leads to two important implications.  
 	First,	the decaying patterns of the optimal regret  rate  are different when   offline sample size $n$ belongs to different ranges.    	
 	 To better illustrate this phenomenon, we first consider the \textit{well-separated} case where $\delta$ is a constant independent of $T$. This case frequently happens in reality as it suggests that the seller's historical price is suboptimal and quite different  from the true optimal price. In this case, as  $n$ increases,  the optimal regret rate first remains at the level of $\widetilde \Theta(\sqrt T)$ when $n \lesssim  \sqrt T$, then gradually decays according to $\widetilde \Theta(\frac{T}{n})$ when  $\sqrt T \lesssim n \lesssim   T$, and finally reaches   $\widetilde \Theta(\log T)$ when $n \gtrsim T$. This is  depicted in Figure \ref{figure1},   from which we can clearly see that there are three ranges of   $n$, i.e., $0<n \lesssim \sqrt{T}$, $\sqrt{T}\lesssim n \lesssim T$  and $n \gtrsim T$, referred to as the first, second and third \textit{phase} respectively, and the optimal regret shows  different  properties  in different  phases. 
 	 We  refer to the significant transitions between the regret-decaying patterns of different phases  as \textit{phase transitions}. 
 	 
\begin{figure}[htbp!]
 		\centering
 		\includegraphics[width=0.8\textwidth]{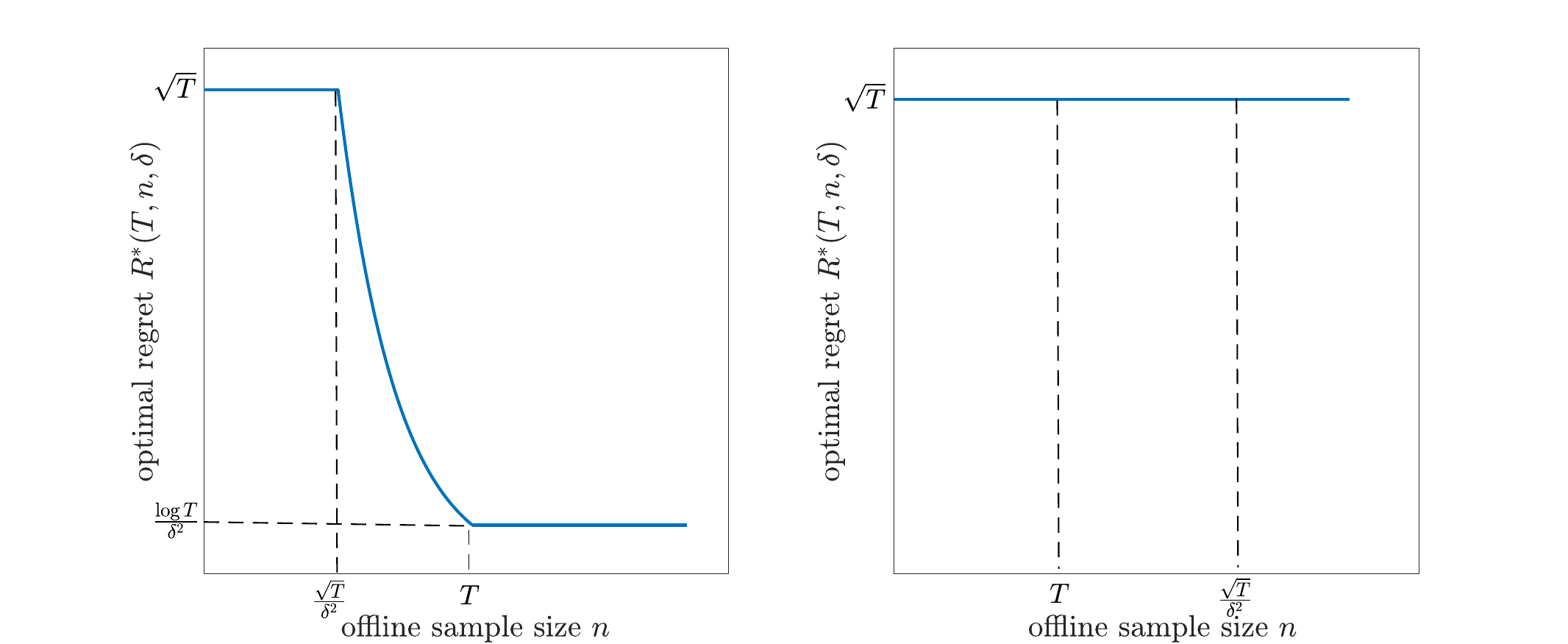}
 		\caption{Phase transitions for the single-historical-price setting with general $\delta$. Left figure: $\delta \gtrsim T^{-\frac{1}{4}}$; right figure: $\delta \lesssim T^{-\frac{1}{4}}$}
 		\label{figure2}
 	\end{figure}

 	  	In contrast to the well-separated case where the phase transitions do  not depend on the value of $\delta$, in the general case where $\delta$ may be very small, we cannot simply ignore the effect of $\delta$ in  the optimal regret, and as a result, the number of phases and the thresholds of the offline sample size that define different phases are closely related to the magnitude of $\delta$.   
 	  	As illustrated in Figure~\ref{figure2}, when  $\delta=  \widetilde \Omega(T^{-\frac{1}{4}})$, similar to the well-separated case, there are three phases defined by two thresholds of $n$: the optimal regret remains at the level of $\widetilde \Theta(\sqrt T)$ in the first phase, and gradually decays according to $\widetilde \Theta(\frac{T}{n\delta^2})$ in the second phase, and stays at the level of  $\widetilde \Theta(\frac{\log T}{\delta^2})$  in the third phase.   
 	  	When $\delta= \widetilde{\mathcal{O}}(T^{-\frac{1}{4}})$,  there is no phase transition, and the optimal regret rate is always $\widetilde \Theta(\sqrt T)$.

Second, Corollary \ref{coro-opt regret-single}  also  characterizes  the impact  of the location of  offline data relative to the optimal price on the optimal regret, which can be stated in the following  \textit{inverse-square law}:  whenever   offline data take effect, i.e., $\delta=\widetilde \Omega(T^{-\frac{1}{4}})$, and $n$ is in the second phase or the third phase, the optimal regret is inversely proportional to the square of generalized distance $\delta$.  
 	Therefore, the factor $\delta^{-2}$ is intrinsic in the regret bound.  
 	Seemingly counter-intuitive, the inverse-square law indicates that the closer  the historical price is to the optimal price, the more difficult it is to learn the demand parameter, and the larger the optimal regret will be.  In fact, this is a consequence of the exploration-exploitation trade-off. In the presence of   offline data,   a ``good'' learning algorithm needs to deviate from   historical price $\hat p$ to conduct price experimentation. However, when $\delta$ is extremely small, such a deviation
 	will also lead to a significant gap with the optimal price, and therefore incurs greater revenue loss to the seller.  
 	 In an extreme case when the historical price happens to be the optimal price, i.e., $\delta=0$,  even if   $n=\infty$, the optimal regret is always $\widetilde \Theta(\sqrt T)$.

 \section{Multiple Historical Prices}\label{sec-extension-historical price}  
  In this section,  
  we consider the multiple-historical-price setting, where  the $n$ historical prices can be    different. In this case, $\sigma$ can be strictly positive and will play an important role to further reducing the complexity of the online learning task.

 \subsection{M-O3FU Algorithm and Regret Upper Bound}\label{subsec-multiple-alg-upper bound}
 In this subsection,  we develop a  learning algorithm for the multiple-historical-price setting. 
We  first make the following observations.

	\begin{enumerate}
		\item[(i)] If 
		$n\sigma^2\gtrsim\sqrt{T}$ and 
		$\delta \lesssim T^{-\frac{1}{4}}$,   
		then the offline data provide so much information   that there is no need for online learning.  In fact,  by simply running  linear regression on the offline data, we can  obtain the estimate  $\hat \theta_0$ for the true demand parameter with  the squared  estimation error  of  $\mathcal{O}(\frac{1}{n\sigma^2})$, i.e., $\mathbb{E}[||\hat \theta_0 - \theta^* ||^2]=\mathcal{O}(\frac{1}{n\sigma^2})$, which means that by simply charging price $\psi(\hat \theta_0)$  in each online period, we achieve the regret  of $\mathcal{O}(\frac{T}{n\sigma^2})$. Note that this $\mathcal{O}({\frac{1}{n\sigma^2}})$-type  estimation error  cannot be further  improved in the online process by policies within  $\Pi^{\circ}$, since when  $T \delta^2\lesssim \sqrt T \lesssim n\sigma^2$, we have 
		\begin{align}\label{ineq-corner case}
		\mathbb{E}[J(\hat p_1,\ldots, \hat p_n , p_1, \ldots, p_T)]
		& \le 2 \Big(  \mathbb{E}[J(\hat p_1, \ldots, \hat p_n)]+\sum_{t=1}^{T} \mathbb{E}[(p_t-p^*)^2]+ T(\bar p_{1:n}-p^*)^2\Big)  \lesssim n\sigma^2, 
		\end{align} 
		where $J(x_1, x_2, \ldots, x_k) :=\sum_{i=1}^{k}(x_i -\frac{1}{k}\sum_{j=1}^{k}x_j)^2$ for any sequence $\{x_i\}_{i=1}^k$ and $k \ge 1$.  This suggests that  in the online process, exploration is ``useless'' in the sense that it cannot bring any theoretical improvement (in terms of reducing the \emph{order} of estimation error) beyond   offline regression.
		Therefore, if the algorithm knew   that conditions $n\sigma^2 \gtrsim \sqrt T$ and $\delta \lesssim T^{-\frac{1}{4}}$ hold, then there is  no exploration-exploitation trade-off at all.
		\item[(ii)] If  in addition to the conditions in   (i), a further extreme condition  $\delta^2\lesssim \frac{1}{n\sigma^2}$ occurs, then even the above offline-regression-based approach may  still be  conservative: if an algorithm knew that   $\delta^2\lesssim \frac{1}{n\sigma^2} \lesssim \frac{1}{\sqrt T}$,
		then by simply charging $\bar{p}_{1:n}$ in every online period, it achieves  the regret of ${\mathcal{O}}(T\delta^2)$,
		which is even better than 
		${\mathcal{O}}(\frac{T}{n\sigma^2})$. 
		We refer to     $\delta^2 \lesssim \frac{1}{n\sigma^2}\lesssim \frac{1}{\sqrt{T}}$ as the  {\textit{corner case}}, and its complement as the  {\textit{regular case}}. 
		\item[(iii)] However, since the algorithm does not know the value of $\delta$   in advance, it does not know  whether it is in the corner case (i.e., whether $\delta^2 \lesssim \frac{1}{n\sigma^2}\lesssim \frac{1}{\sqrt{T}}$ is true) in advance. If the conditions  in (i)  do not hold, then the algorithm still needs online exploration; if the condition in (ii) does not hold, then the algorithm still needs offline regression.
\end{enumerate}

Motivated by the above  observations, we design the following \textit{Modified O3FU} (M-O3FU) algorithm. With an abuse of terminology,   
	we refer to    O3FU algorithm in this section  
	as the one proposed in \S \ref{subsec-O3FU} after natural  modification to the multiple-historical-price setting by letting $V_{0,n}=\lambda I+\sum_{i=1}^{n}[1\, \,  \hat p_i]^{\top} [1 \, \,   \hat p_i]$, $Y_{0,n}=\sum_{i=1}^{n}\hat D_i [1\,\, \hat p_i]^{\top}$, and $p_1=l\cdot\mathbb{I}\{\bar p_{1:n}>\frac{u+l}{2}\}+u\cdot\mathbb{I}\{\bar p_{1:n}\le\frac{u+l}{2}\}$.  
 \begin{algorithm}[thbp] 	\label{Alg-M-O3FU}
 	\caption{M-O3FU Algorithm}
 	\textbf{Input}: offline data $\{(\hat p_i,\hat D_i)\}_{i=1}^n$, support of demand parameters $\Theta^{\dag}$, price range $[l,u]$,  regularization parameter $\lambda = 1+u^2$, $\{w_t\}_{t\ge 0}$ defined in \eqref{OFU-radius} with {$\epsilon=\frac{1}{T^2}\wedge \frac{1}{n\sigma^2}$},   parameter $K>1$\;%
 	\textbf{Initialization}: $V_{0,n}=\lambda I+\sum_{i=1}^{n}[1 \, \, \hat p_i]^{\mathsf{T}}[1 \, \, \hat p_i]$, $Y_{0,n}=\sum_{i=1}^{n}\hat D_i [1 \,\, \hat p_i]^{\mathsf{T}}$,  $\hat \theta_{0} = V_{0,n}^{-1}Y_{0,n}$,  $\mathcal C_0=\{\theta \in \Theta^{\dag}: ||\theta-\hat \theta_0||_{V_{0,n} }\le w_0 \}$\; 
 	\eIf{$\frac{\min_{\theta \in \mathcal C_0} |\bar{p}_{1:n}-\psi(\theta)|}{\max_{\theta_1,\theta_2\in \mathcal C_0} |\psi(\theta_1)-\psi(\theta_2)|}\le K$, and  $n\sigma^2 \ge \sqrt T$}{Charge price $p_t=\bar p_{1:n}$ for each $t \in [T]$\;}
 	{Run O3FU Algorithm.} 
 \end{algorithm}

We next make several highlights about   M-O3FU algorithm.  
	First,  in comparison with   O3FU algorithm, before the start of the   online learning process, M-O3FU  algorithm  takes a  preliminary step  that tests whether the distance between    $\bar p_{1:n}$ and   interval $  \{\psi(\theta):\theta \in \mathcal C_0\}$ is smaller than a constant times the length of   interval $\{\psi(\theta):\theta \in \mathcal C_0\}$. 
	 The goal of this step is to test whether   condition $\delta^2\lesssim \frac{1}{n\sigma^2}$ holds or not.   
	 If this condition is inferred to hold based  on the empirical observation, and in addition, $n\sigma^2\ge \sqrt T$, the algorithm keeps using  $\bar p_{1:n}$ for each online period. Otherwise, the algorithm simply runs O3FU algorithm.  
	  Second,   parameter $\epsilon$ defined  in $w_t$   is modified from $\frac{1}{T^2}$ (used  in O3FU)  to $\frac{1}{T^2}\wedge \frac{1}{n\sigma^2}$, which guarantees that  $\theta^*$ belongs to each confidence ellipsoid $\mathcal{C}_t$   with sufficiently high probability, and the revenue loss incurred when $\theta^*$ does not belong to some confidence ellipsoid can be bounded by  $\mathcal{O}(\frac{T}{n\sigma^2}\wedge\frac{1}{T})$.

 The following theorem provides an upper bound on the regret of    M-O3FU algorithm. 
 \begin{theorem}\label{thm-upper bound-multiple prices-T}
 	Let $\pi$  be   M-O3FU algorithm. Then there exists a finite  constant $K_3>0$  such that for any  $T \ge 1$, $n\ge 0$,  $\sigma \ge 0$ and $\bar p_{1:n} \in [l,u]$, and for any possible value of $\theta^* \in \Theta^{\dag}$, we have 
 	\begin{align*}%
 	R^{\pi}_{\theta^*}(T)  \le  
 	\left\{  
 	\begin{array}{ll}
 	K_3 \cdot \big(T\delta^2+1\big), &\text{if } \delta^2 \lesssim \frac{1}{n\sigma^2} \lesssim\frac{1}{\sqrt T}; \\ 
 	K_3\cdot \Big(\big(\sqrt T\log T\big) \wedge \frac{T(\log T)^2}{n\sigma^2+(n\wedge T) \delta^2}+1\Big),  \quad & 
 	\text{otherwise}.\\
 	\end{array} 
 	\right. 
 	\end{align*} 
 \end{theorem}

{Theorem \ref{thm-upper bound-multiple prices-T} shows that the regret upper bound has different forms in two different cases.  
	 When   $\delta^2 \lesssim \frac{1}{n\sigma^2} \lesssim\frac{1}{\sqrt T}$,   M-O3FU algorithm achieves the regret upper bound $\mathcal{O}(T\delta^2+1)$, which matches the ideal regret bound  in the above item (ii) discussed at the beginning of this subsection. Otherwise,  the regret upper bound becomes $\widetilde{\mathcal{O}}(\sqrt{T}\wedge \frac{T}{n\sigma^2+(n \wedge T)\delta^2}+1)$.  
	  Compared with the upper bound in Theorem~\ref{thm-UCB}, there is  an additional term $n\sigma^2$ in the denominator capturing the effect of the dispersion of offline data. 
	 We summarize the regret upper bound under different  $(n, \sigma, \delta)$ combinations  in Table \ref{table-multiple historical prices-upper bound} of Appendix~H.}

The proof of Theorem \ref{thm-upper bound-multiple prices-T}  can be found in Appendix B.1. Similar to the proof of  Theorem \ref{thm-UCB}, we also need an important technical lemma stated as follows.
\begin{lemma} \label{lemma-proof-multiple historical prices-step2}
	Suppose we run    O3FU algorithm   from $t=1$ with given input offline data $\{(\hat p_i, \hat D_i)\}_{i=1}^n$, 
	$\sigma \le \delta$, $\delta \ge \max\{ \frac{\sqrt{2(\alpha_{\max}^2+\beta_{\max}^2)}}{\beta_{\max}^2}\frac{T^{1/4}w_T}{n^{1/2}}, \sqrt C_1 T^{-1/4}\}$, and  $\theta^* \in \mathcal C_t$ for each $t \in [T]$, then  two  sequences of events $\{U_{t,3}\}_{t=1}^T$ and $\{U_{t,4}\}_{t=2}^T$  also hold, where 
	\begin{align*} 
	&U_{t,3}=\Big\{|p_t -\bar p_{1:n}| \ge \min\big\{1-\frac{\sqrt 2}{2}, \frac{C_0}{2}\big\}\cdot \delta\Big\},\\
	&U_{t,4}=\Big\{ ||\tilde \theta_{t}-\theta ^*||^2 \le C_3 \frac{w_{t-1}^2}{n\sigma^2+(n\wedge (t-1))\delta^2}\Big\},
	\end{align*}
	and $C_0$ and $C_1$ are  defined in Lemma \ref{lemma-proof-UCB-step2}, and
	$ 
	C_3=\max\Big\{8(u-l)^2, 4C_1, 2\max\{2(\sqrt 2+1)^2, \frac{4}{C_0^2}\}\cdot ((4u+1)^2+1) \Big\}.$
\end{lemma}

{Similar to  Lemma \ref{lemma-proof-UCB-step2},  Lemma \ref{lemma-proof-multiple historical prices-step2} is also proved based on induction  arguments. Besides,    
we   need to use the following lower bound on  the sum of squared price deviations: 
	\begin{align}\label{ineq-variance p lower bound}
	J(\hat p_1, \ldots, \hat p_n, p_1, \ldots, p_t) 
	&\ge J(\hat p_1, \ldots, \hat p_n)+ \frac{n}{n+t}\sum_{s=1}^{t}(p_s-\bar p_{1:n})^2,
	\end{align}
	where $J(x_1, x_2, \ldots, x_k) :=\sum_{i=1}^{k}(x_i -\frac{1}{k}\sum_{j=1}^{k}x_j)^2$ for any sequence $\{x_i\}_{i=1}^k$ and $k \ge 1$. 
	We can interpret $J(x_1, x_2, \ldots, x_k)$ as the information metric capturing the   variation for a sequence $\{x_i\}_{i=1}^k$. Then inequality \eqref{ineq-variance p lower bound} bounds the information accumulated up to period $t$ from below, through  the pre-existing offline information, plus the information due to the deviation of the algorithm's prices  from the average historical price. The  proof  of  Lemma \ref{lemma-proof-multiple historical prices-step2} is  provided  in Appendix B.2.   
}

\subsection{Lower Bound on Regret}\label{subsec-extension-lower bound}
In this subsection, we establish a lower bound on the best-achievable regret for the \texttt{OPOD} problem among the class of admissible policies $\Pi^{\circ}$   defined in a similar way to \eqref{def-minimax optimal policy}.  Again, we denote $R_{\theta}^{\pi}(T, n, \delta, \sigma)$ as the regret incurred by policy $\pi \in \Pi^{\circ}$ under the   demand parameter   $\theta$.  
\begin{theorem} \label{thm-lower bound-multiple historical prices}  
	There exists a  positive constant $K_4$ such that for any admissible policy $\pi \in \Pi^{\circ}$,
	for any $\xi \in (0, 1)$,  
	$T \ge 2$,   $n \ge 0$ and $\sigma \ge 0$, and for any $\delta \in [0, u-l]$,
	\begin{align*}
\sup_{\mathcal{D} \in \mathcal{E}(R);  \atop \theta \in \Theta^{\dag}: |\bar p_{1:n} -\psi(\theta)| \in [(1-\xi)\delta, (1+\xi)\delta]}  	R^{\pi}_{\theta}(T, n, \delta,  \sigma) \ge  
	\left\{  
	\begin{array}{ll} 
	K_4\cdot   T\delta^2,  \quad & \text{if } \delta^2 \lesssim \frac{1}{n\sigma^2} \lesssim  \frac{1}{\sqrt T}; \\
	K_4\cdot  \Big(\frac{\sqrt{T}}{(\log T)^{\lambda_0}} \wedge \frac{T}{n\sigma^2+(n\wedge T)\delta^2}\Big),  \quad & \text{otherwise}.  
	\end{array} 
	\right.
	\end{align*}
\end{theorem} 

 Similar to Theorem \ref{thm-lower bound}, the  instance-dependent environment class is defined as the set of instances whose associated optimal prices are away from  $\bar p_{1:n}$ by a distance  $\Theta(\delta)$. 
	Since   M-O3FU algorithm achieves the regret upper bound $\mathcal{O}(\sqrt T \log T)$ for  any value of $\theta^* \in \Theta^{\dag}$ (thus belongs to  the admissible policy class $\Pi^{\circ}$ with $\lambda_0\ge 1$), 
	Theorem \ref{thm-lower bound-multiple historical prices}  demonstrates  that for both the corner and regular cases, the regret  rate achieved by   M-O3FU algorithm in Theorem \ref{thm-upper bound-multiple prices-T}  cannot be further improved by any policy in  $\Pi^{\circ}$. 	%
	The proof of Theorem \ref{thm-lower bound-multiple historical prices} is provided in Appendix B.4, which is a generalization to that of Theorem \ref{thm-lower bound}.  We also summarize the regret lower bound under different $(n, \sigma, \delta)$ combinations in Table \ref{table-multiple historical prices-lower bound} of Appendix~H.

\subsection{Phase Transitions and Generalized Inverse-Square Law}\label{subsec-implication-multiple}
Motivated from the matching upper and lower bounds (after ignoring logarithm factors) in Theorems \ref{thm-upper bound-multiple prices-T} and  \ref{thm-lower bound-multiple historical prices} respectively,  
 we define the optimal instance-dependent regret for the \texttt{OPOD} problem in the  multiple-historical-price setting as follows: 
\begin{align}\label{def2-optimal regret}
R^*(T, n,\delta, \sigma) = \inf_{\pi \in \Pi^{\circ}} \sup_{\mathcal{D} \in \mathcal{E}(R);  \atop \theta \in \Theta^{\dag}: |\psi(\theta)-\bar p_{1:n} | \in [(1-\xi)\delta, (1+\xi)\delta]} R^{\pi}_{\theta}(T,n,\delta,\sigma),
\end{align}  
where a slight difference compared with \eqref{def1-optimal regret}  is the modification from the single historical price $\hat p$ to the average historical price $\bar p_{1:n}$.

Combining  Theorem \ref{thm-upper bound-multiple prices-T} and Theorem \ref{thm-lower bound-multiple historical prices}, we are able to characterize the optimal regret of the \texttt{OPOD} problem for the multiple-historical-price setting.

\begin{corollary}\label{corollary-multiple-optimal regret}
	The optimal regret defined in \eqref{def2-optimal regret} for the multiple-historical-price setting is
	\begin{align*}
	R^*(T,n, \delta, \sigma)=
	\left\{  
	\begin{array}{ll}
	\widetilde \Theta \Big( \sqrt T \wedge \frac{T }{n\sigma^2+(n\wedge T)  \delta^2}\Big),  \quad & \text{for the regular case}; %
	\\ 
	\widetilde \Theta \big(T \delta^2\big),  \quad & \text{for the corner case}. 	\end{array} 
	\right. 
	\end{align*}
\end{corollary}

Recall that in the single-historical-price setting,   the threshold  $ \widetilde{\Theta}(T^{-\frac{1}{4}})$ of $\delta$ plays an important role in characterizing the behavior of the optimal regret rate. 
This threshold $ \widetilde{\Theta}(T^{-\frac{1}{4}})$  also plays a  role in the optimal regret rate of the multiple-historical-price setting. 

\begin{figure}[htbp!]
	\centering
	\includegraphics[width=0.7\textwidth]{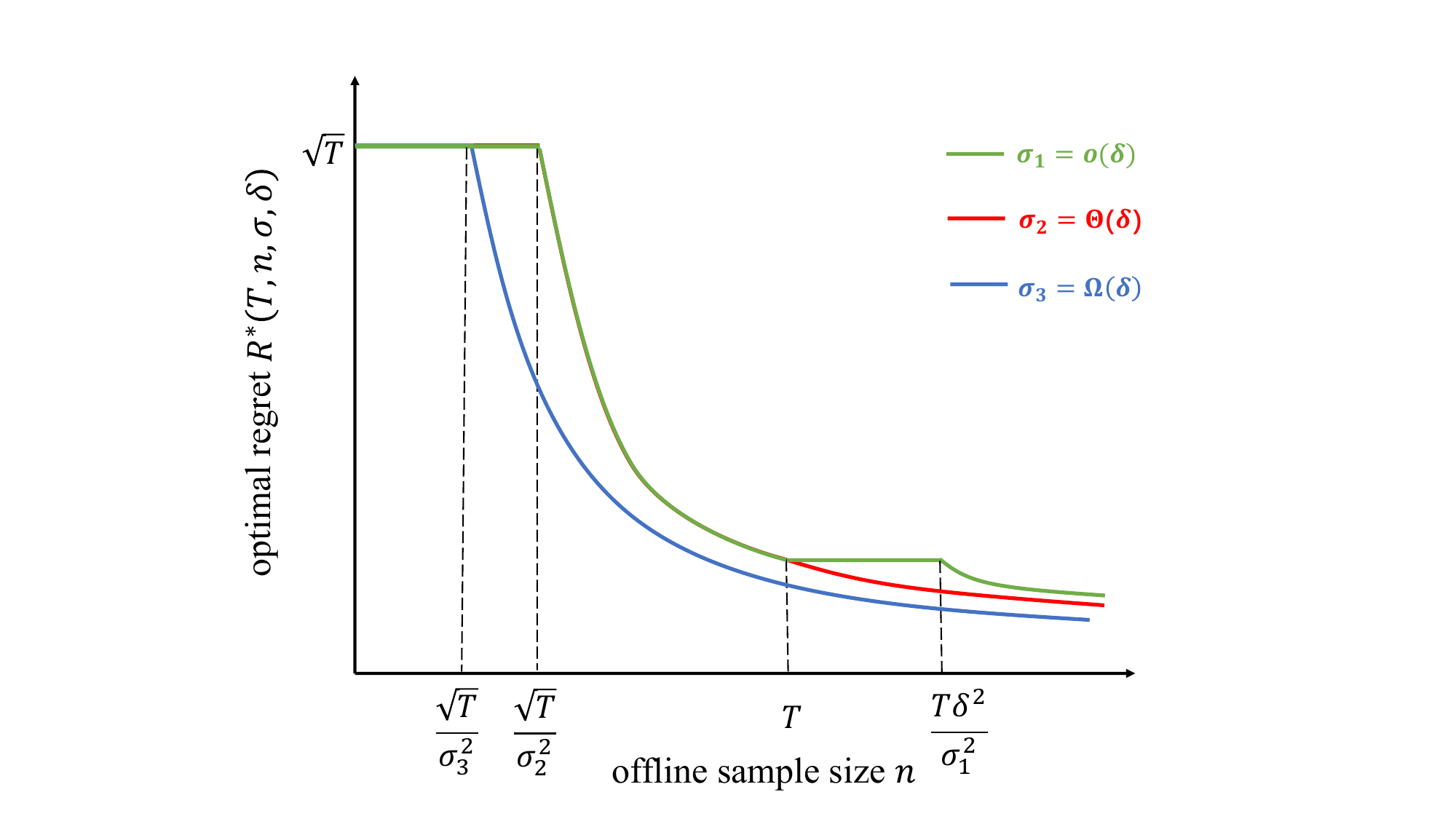}
	\caption{ Multiple-historical-price setting with $\delta \gtrsim T^{-\frac{1}{4}}$ and different  $\sigma$}
	\label{figure-changing sigma}
\end{figure}

When  $\delta \gtrsim T^{-\frac{1}{4}}$, there are significant differences for the behaviors of the optimal regret rate, depending on whether    $\sigma$ is less than, equal to or greater than  $\delta$. This is illustrated in Figure \ref{figure-changing sigma}, where the  green, red  and  blue curves depict   the above three cases respectively. 
If $\sigma=o(\delta)$, as  shown in the green curve,  
the optimal regret rate exhibits four decaying patterns as $n$ changes between different ranges. Specifically, the optimal regret rate  first remains at  $\widetilde \Theta(\sqrt T)$ when $n\lesssim \frac{\sqrt{T}}{\delta^2}$, and then decreases according to $\widetilde \Theta(\frac{T}{n\delta^2})$ when  $ \frac{\sqrt T}{\delta^2} \lesssim n \lesssim  T$. After that, the optimal regret rate  stays at $\widetilde \Theta(\frac{\log T}{\delta^2})$ when    $ T \lesssim n \lesssim  \frac{T\delta^2}{\sigma^2}$, and finally, it decreases according to $\widetilde \Theta(\frac{T}{n\sigma^2})$ when $n\gtrsim \frac{T\delta^2}{\sigma^2}$. 
If $\sigma=\Theta(\delta)$ or $\sigma=\Omega(\delta)$  as shown in the red or blue curve, the optimal regret rate exhibits two phases: it remains at the level of $\widetilde \Theta(\sqrt T)$ when $n\lesssim \frac{\sqrt T}{\sigma^2}$, and decays according to $\widetilde \Theta(\frac{T}{n\sigma^2})$ when $n\gtrsim \frac{\sqrt T}{\sigma^2}$.  
Therefore, when $\sigma$ gradually increases, depending on its magnitude compared with $\delta$, the number of phases of the optimal regret rate also experiences the change from four phases to two phases, and the entire patterns of the phase transitions of the optimal regret rate also change accordingly.  

\begin{figure}[tbp!]
	\centering
	\includegraphics[width=0.8\textwidth]{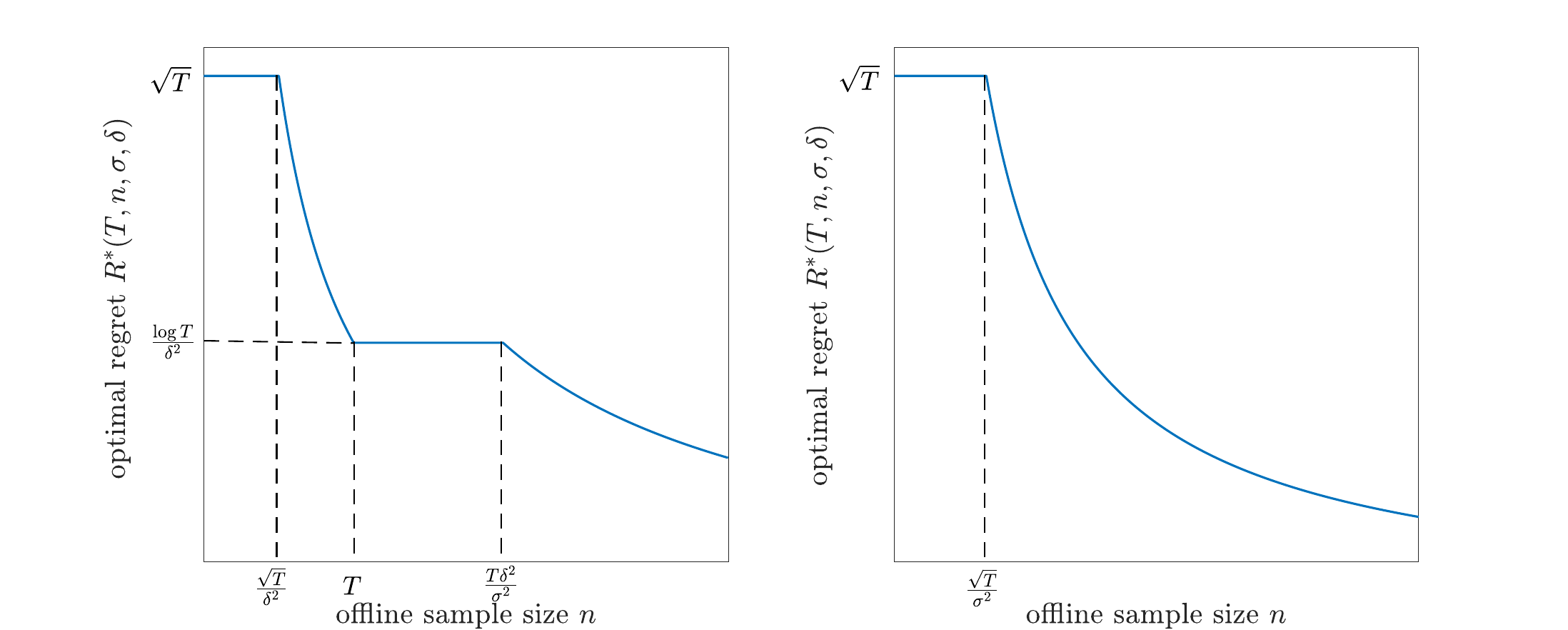}
	\caption{Phase transitions for the   multiple-historical-price setting with $\delta\gtrsim T^{-\frac{1}{4}}$. Left figure: $\sigma=o(\delta)$; right figure: $\sigma=\Omega(\delta)$}
	\label{figure3}
\end{figure}

Corollary \ref{corollary-multiple-optimal regret} also reveals a generalized inverse-square law.  Specifically,   the optimal regret is inversely proportional to the square of both $\delta$ and $\sigma$, which quantifies the effect of the location and dispersion of   the offline data on the optimal regret. The intuition for the dependence of the optimal regret on $\delta$ is similar to the  single-historical-price setting. For the dependence of the optimal regret on $\sigma$,  as the historical prices become more dispersive, i.e., $\sigma$ increases,   the seller can obtain a more accurate estimate for the unknown demand parameter from   offline regression,  which helps to further reduce the optimal regret of the online learning process.

It's also worth noting that the thresholds of  the offline sample size  that define different phases of the optimal regret depend on  both $\delta$ and $\sigma$. When $\sigma=\mathcal{O}(\delta)$ and  $\delta= \widetilde{\Omega}(T^{-\frac{1}{4}})$, the first threshold of $n$ that defines the first and   second phases, i.e., $\widetilde{\Theta}(\frac{\sqrt T}{\delta^2})$, decreases in   $\delta$. When $\sigma=\Omega(\delta)$ and  $\delta=\widetilde{\Omega}(T^{-\frac{1}{4}})$, the threshold of $n$ that defines the first and   second phases, i.e., $\widetilde{\Theta}(\frac{\sqrt T}{\sigma^2})$, decreases in the standard deviation $\sigma$.  This implies that  more offline data will be required to overcome the challenges caused by a shorter generalized distance $\delta$ or a smaller standard deviation $\sigma$.

\begin{figure}[htbp!]
	\centering
	\includegraphics[width=0.4\textwidth]{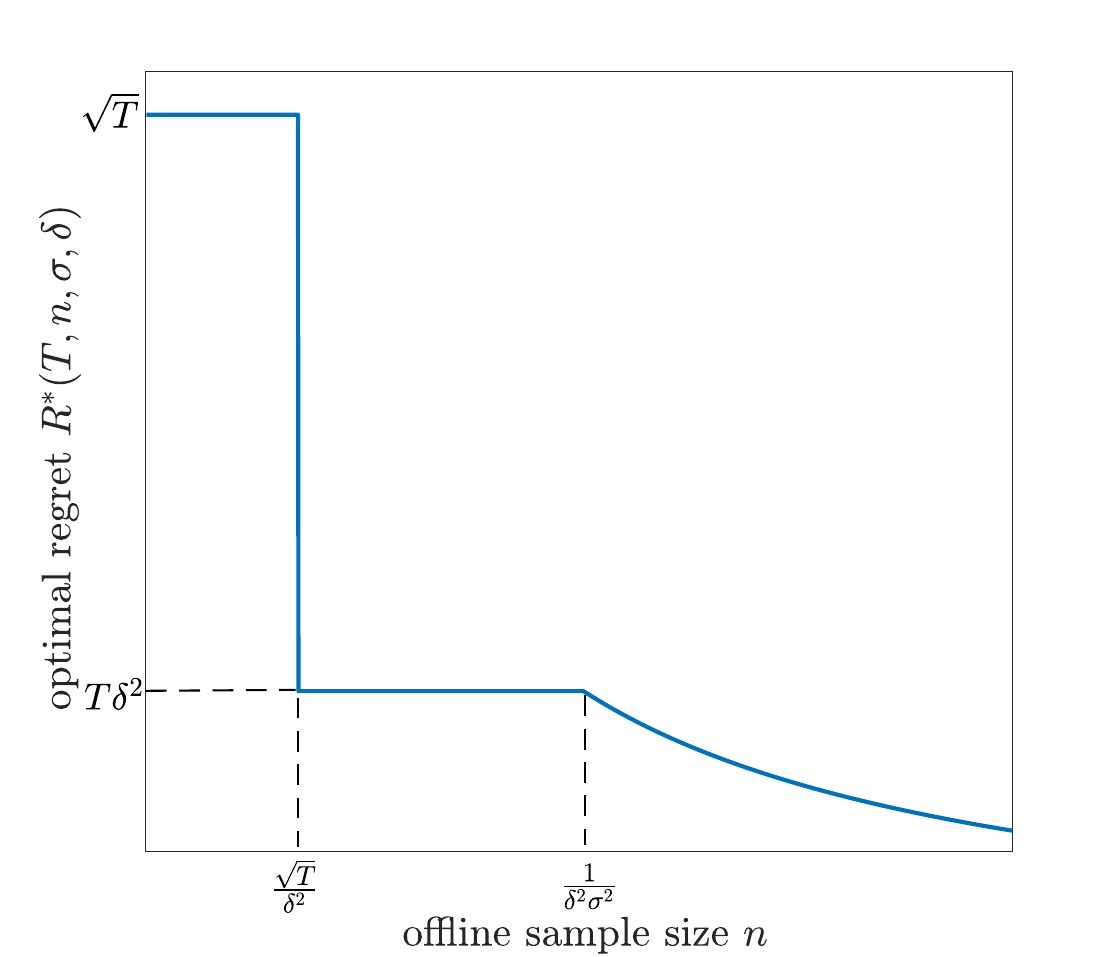}
	\caption{Phase transitions for the   multiple-historical-price setting with $\delta \lesssim T^{-\frac{1}{4}}$}
	\label{figure3_2} 
	
\end{figure}

When   $\delta\lesssim  T^{-\frac{1}{4}}$, Corollary \ref{corollary-multiple-optimal regret} indicates that there are three phases of the optimal regret rate as $n$ changes. When $n \lesssim \frac{\sqrt T}{\delta^2}$, the optimal regret remains at $\widetilde{\Theta}(\sqrt T)$. When   $\frac{\sqrt T}{\delta^2} \lesssim  n \lesssim \frac{1}{\delta^2\sigma^2}$, the optimal regret experiences a \textit{sudden} drop from $\widetilde{\Theta}(\sqrt T)$ to $\widetilde \Theta(T\delta^2)$. When  $n \gtrsim \frac{1}{\delta^2\sigma^2}$, the optimal regret decays according to $\widetilde 
\Theta(\frac{T}{n\sigma^2})$. Such transitions of the optimal regret with different $n$ are illustrated in Figure \ref{figure3_2}. In particular, the second phase corresponds to the corner case defined in \S \ref{subsec-multiple-alg-upper bound}. 
In this case,  smaller $\delta$ leads to lower optimal regret, which is in contrast to the inverse-square law in the regular case.  
This is 
because in the corner case, as discussed in \S \ref{subsec-multiple-alg-upper bound}, there is no need for online learning and therefore no exploration-exploitation trade-off, and   
 the policy that always charges $\bar p_{1:n}$
incurs very small regret. In this case,  the closer the average historical price is to the optimal price,   the smaller the optimal regret will be.  
 By contrast, the inverse-square law in the regular case  is a consequence of the exploration-exploitation trade-off.

\section{Numerical Study}\label{sec:num}
In this section, we test the performance of our algorithm on a synthetic data set. We define the relative regret  for a given learning algorithm $\pi$ as 
$ \frac{ Tp^*\cdot (\alpha^*+\beta^* p^*)-\sum_{t=1}^{T}\mathbb{E}_{\theta^*}^{\pi}[p_t(\alpha^*+\beta^* p_t)] }{Tp^*\cdot (\alpha^*+\beta^* p^*)}\times 100\%$, and    the following three  problem instances are tested:  
\begin{itemize}
	\item[(1)] $\theta^*=[2.6, -1.8]$,  $[\alpha_{\min},\alpha_{\max}]=[2.5, 3.5]$,   $[\beta_{\min},\beta_{\max}]=[-2, -1.3]$,  $[l,u]=[0.1, 2]$, $R=2.2$; 
	\item[(2)]  $\theta^*=[3.7, -3.15]$,  $[\alpha_{\min},\alpha_{\max}]=[3.5, 5]$, $[\beta_{\min}, \beta_{\max}]=[-3.2, -2.5]$, $[l, u]=[0.5, 1.3]$,  $R=2.5$; 
	\item[(3)] $\theta^*=[2.9, -2.6]$,  $[\alpha_{\min},\alpha_{\max}]=[2.8, 3.5]$, $[\beta_{\min}, \beta_{\max}]=[-2.8, -1]$, $[l,u]=[0.2, 2]$,  $R=1.8$. 
\end{itemize}
and  $\varepsilon$ follows a normal distribution with standard deviation $R$. For each of the above instance, we repeat the experiments for 500 times, and the results are computed  after averaging over the 500 experiments. Under the  multiple-historical-price setting, we   test a simplified version of  M-O3FU algorithm by directly running   O3FU, without checking the preliminary condition. Thus, throughout this section,  we  simply call our algorithm    ``O3FU algorithm.''  

\begin{figure}[tbp!]
	\centering
	\includegraphics[width= 0.7\textwidth]{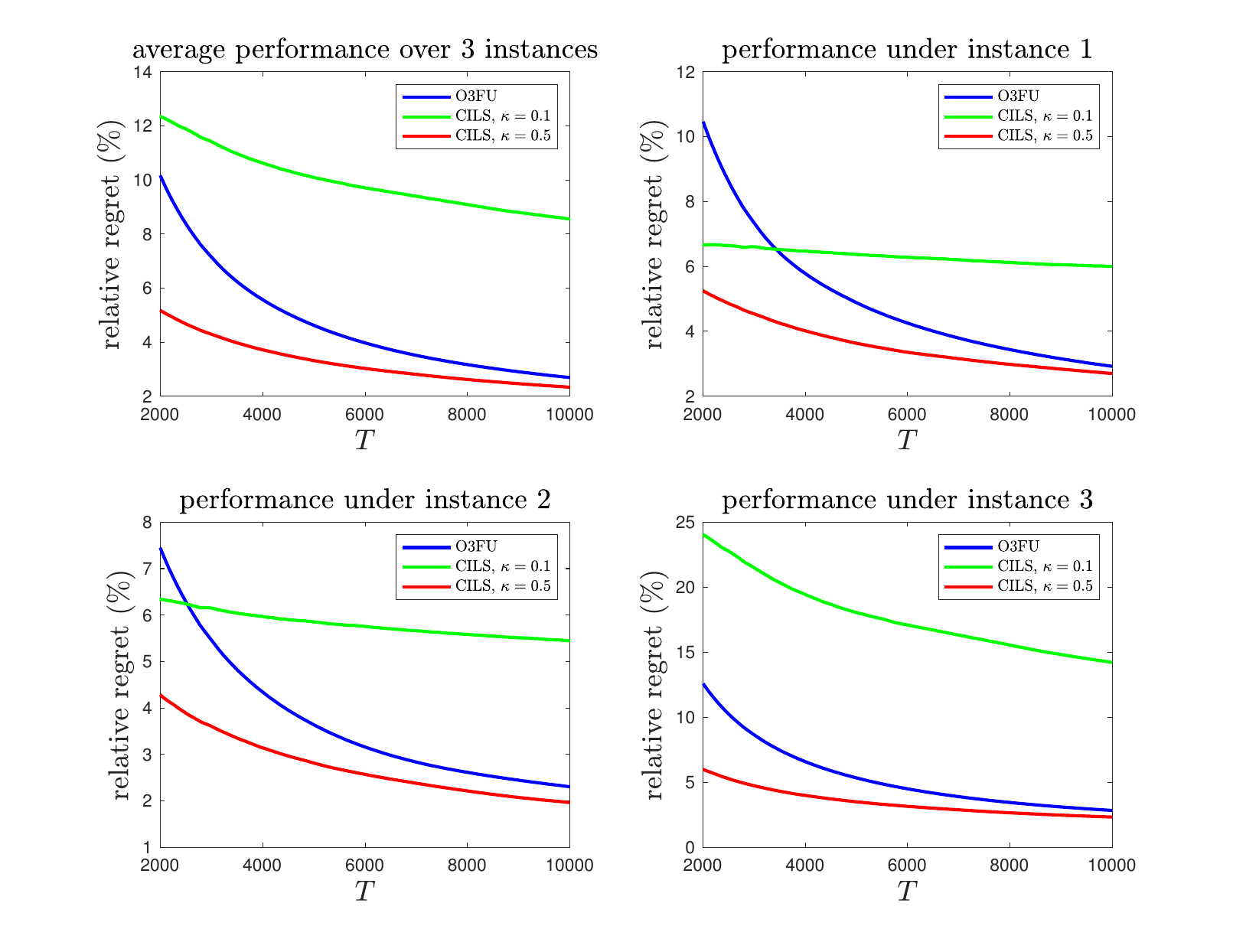}
	\caption{Comparison between  O3FU and CILS when there are no offline data} \label{figure=compare-CILS-no offline data}
	\includegraphics[width=0.7 \textwidth]{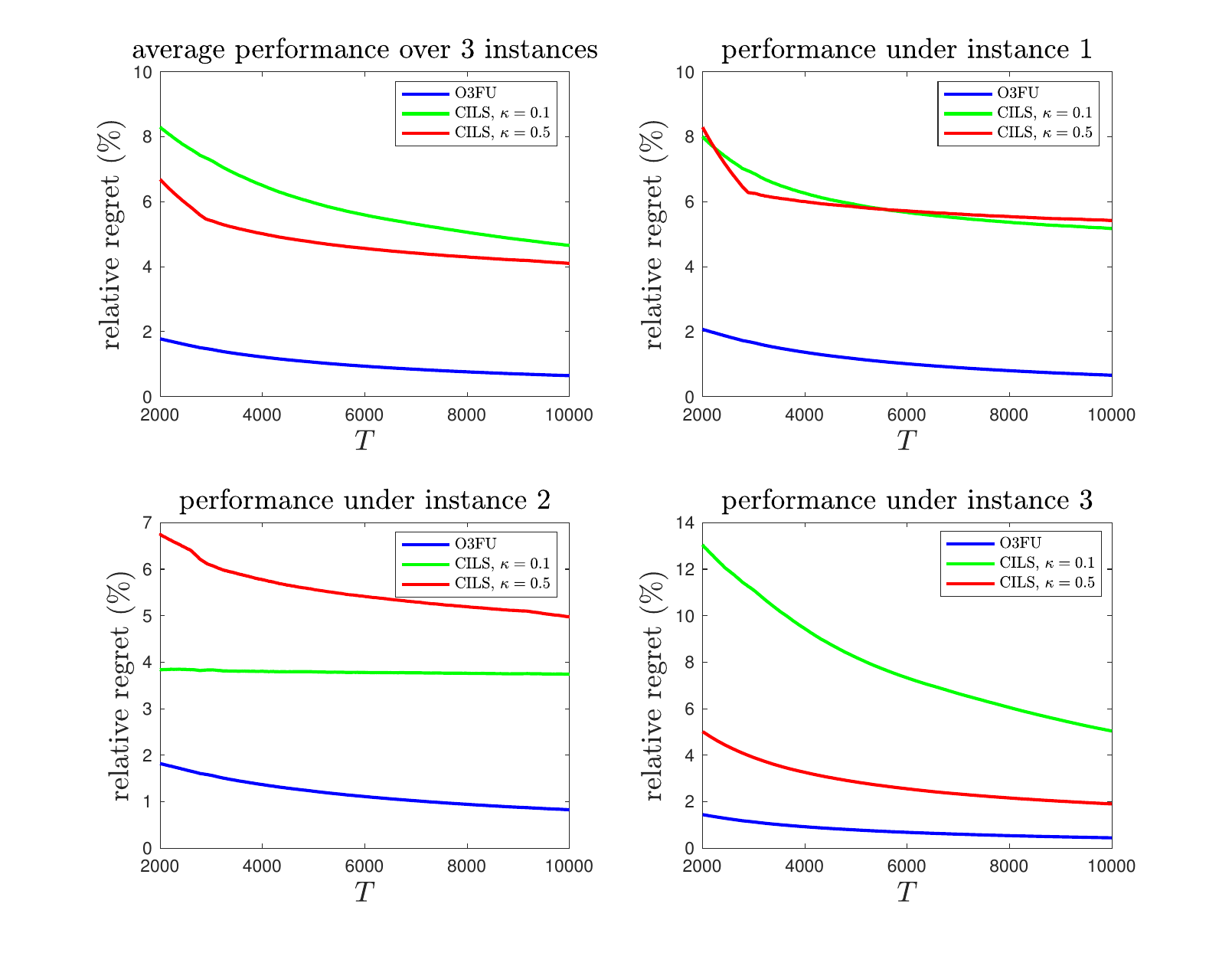}
	\caption{Comparison between  O3FU and CILS when there are $n=1000$ offline demand data} 
	\label{figure-compare-CILS-offline data} 
\end{figure}

First, we compare our O3FU algorithm with the modified Constrained Iterated Least Squares (CILS) algorithm. When there are no offline data, we adopt   CILS algorithm directly from \cite{KeskinZeevi2014}. When there are offline data, no existing learning algorithm in prior literature is directly suitable for this setting, so we  make a natural modification to the original CILS  by incorporating offline data into the least-square estimation. In both cases, we set the tuning parameter $\kappa$ in CILS  to be $0.1$ following  \cite{KeskinZeevi2014}, and also $0.5$ which seems to lead to the best performance of CILS.  
Figures \ref{figure=compare-CILS-no offline data} and  \ref{figure-compare-CILS-offline data} show the performances of O3FU and CILS algorithms for the settings when there are no offline data, and when there are  
$n=1000$ offline demand data under a single historical price (specifically, we  set $\hat p=1.8, 0.9, 1$ for instances (1)-(3) respectively).
As seen from Figure \ref{figure=compare-CILS-no offline data}, without offline data, O3FU performs   better than CILS with $\kappa=0.1$ and  comparably to CILS with $\kappa=0.5$ as $T$ becomes larger. 
 Figure  \ref{figure-compare-CILS-offline data} reveals that with the help of offline data, the regret of O3FU algorithm is  {significantly reduced} for all $T$ under all   instances. 
 By contrast, for CILS algorithms, the impact of offline data on the empirical regret is not obvious and heavily relies on the  tuning parameter and specific problem instance. For CILS with $\kappa=0.1$, the improvement of the relative regret is clear under  instance (3), but rather minimal under instances (1) and (2). For CILS  with $\kappa=0.5$, the regret only decreases a little under instance (3), and  even becomes larger under  instances (1) and (2). Therefore, compared with CILS algorithms, O3FU algorithm  better   exploits the value of offline data and is more robust to different problem instances.

\begin{figure}[tbp]	
	\centering
	\begin{minipage}[t]{0.48\textwidth}
		\centering
		\includegraphics[width=7cm,height=5cm]{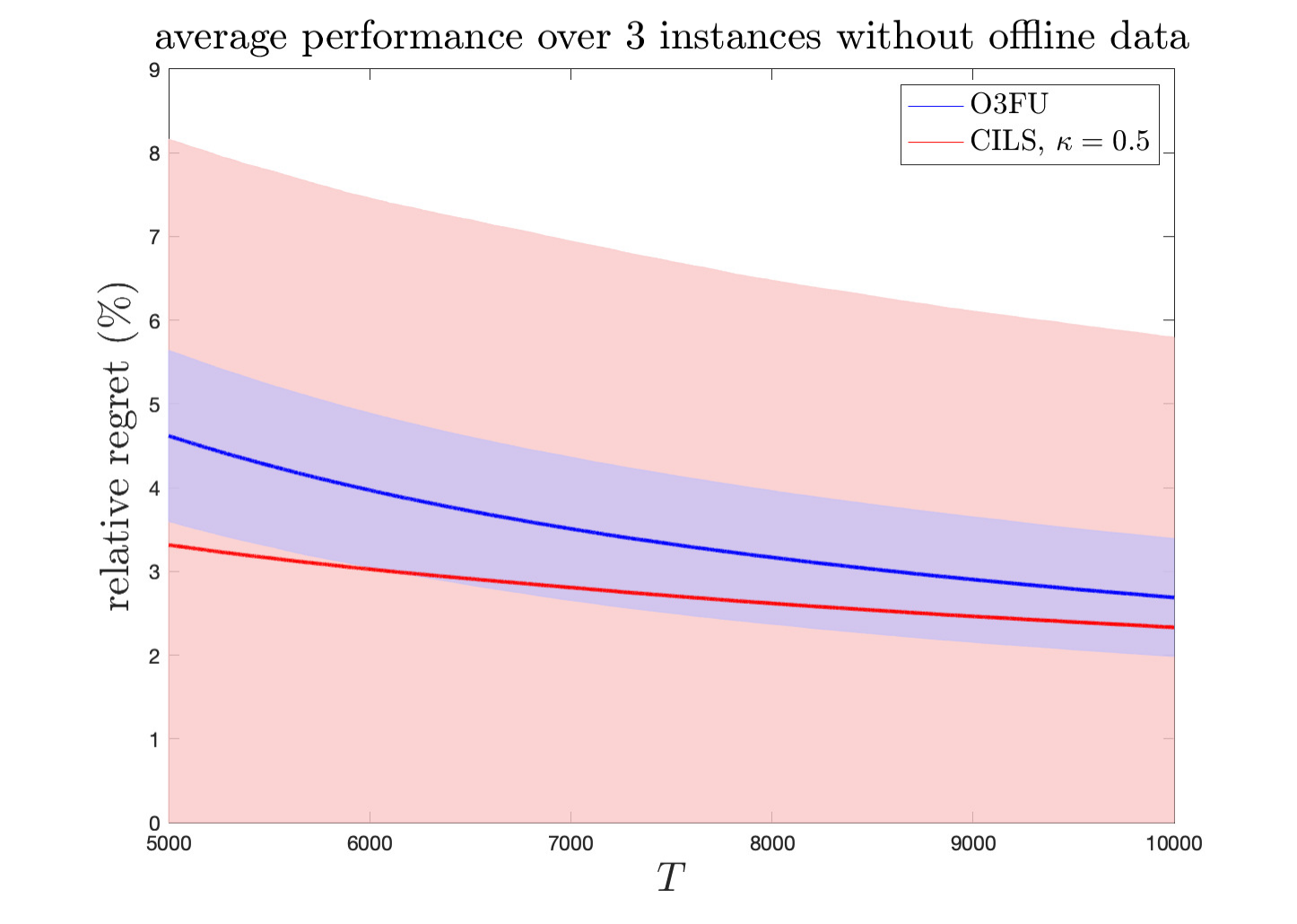}%
	\end{minipage}
	\begin{minipage}[t]{0.48\textwidth}
		\centering
		\includegraphics[width=7cm,height=5cm]{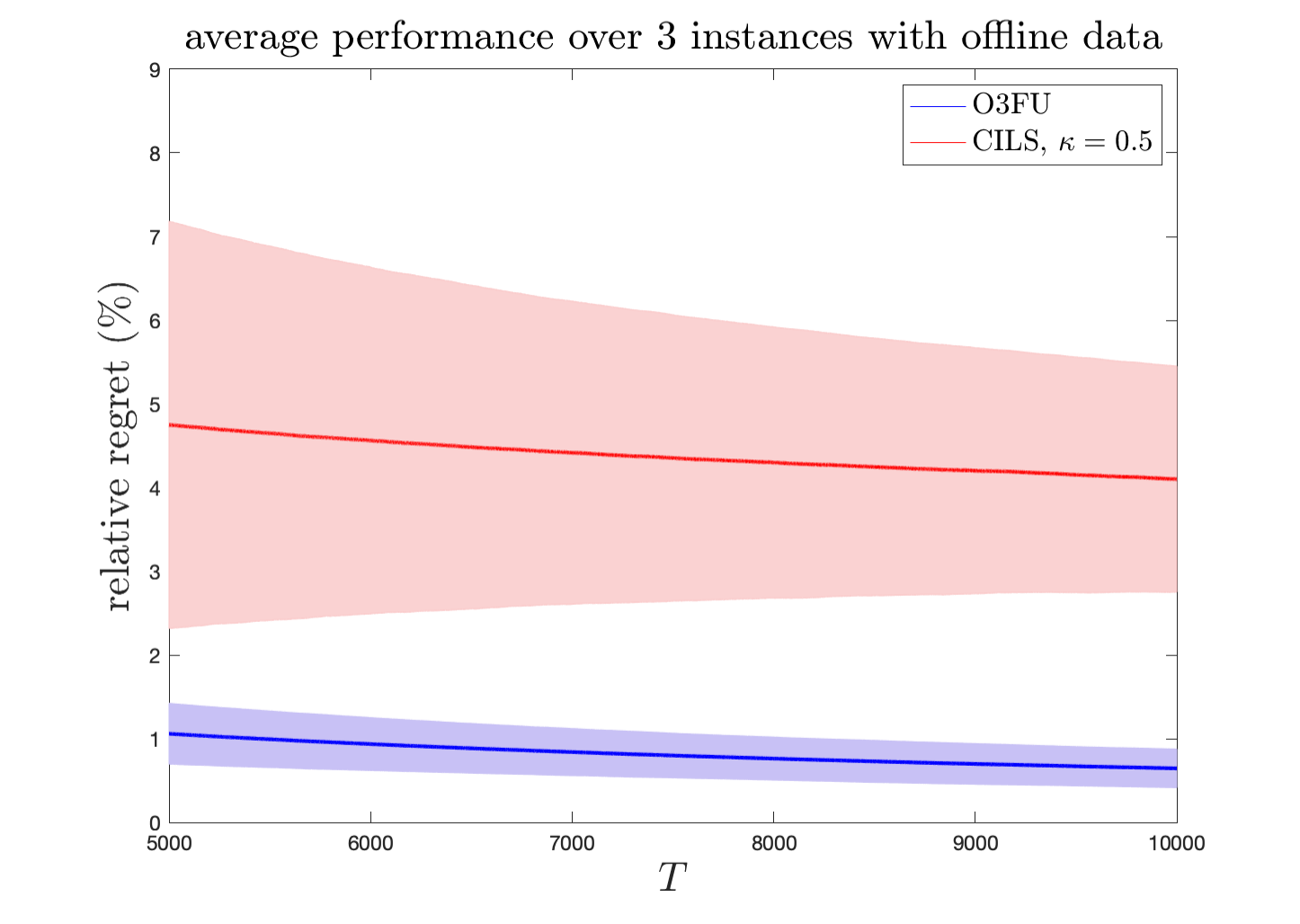}%
	\end{minipage}
	\caption{95\% confidence-region comparison between O3FU and CILS with $\kappa=0.5$}	\label{figure-95 confidence}  
\end{figure} 

Second,  Figure \ref{figure-95 confidence}   plots the
   95\% confidence region of O3FU algorithm and CILS algorithm with $\kappa=0.5$, for both cases when there are no offline data and when there are $n=1000$ offline data. The left figure shows that while CILS with properly tuned parameter  performs slightly better than O3FU on average when there are no offline data, the standard deviation of CILS among the 500 simulations is much larger than O3FU. 
   This implies that O3FU is more stable than CILS. The right figure shows that with offline data, O3FU always outperforms CILS,  in terms of both the average regret and standard deviation.   
Since O3FU algorithm has highly stable performance, we believe that it should be preferable in many real-life business settings.  

\begin{figure}[htbp!]
	\centering
	\includegraphics[width= 0.9\textwidth, trim=0 0 0 0,clip]{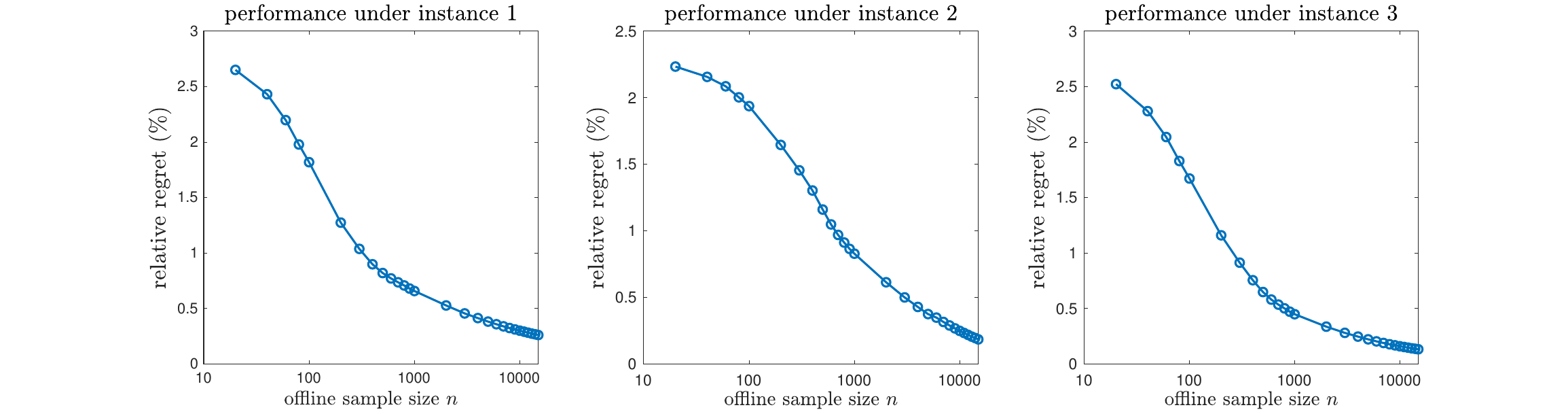}%
	\caption{$T=10^4$-period relative regret for the single-historical-price setting with different $n$} 
	\label{figure-effect_n-offline data} 
	
	\includegraphics[width= 0.9 \textwidth, trim=0 0 0 0,clip]{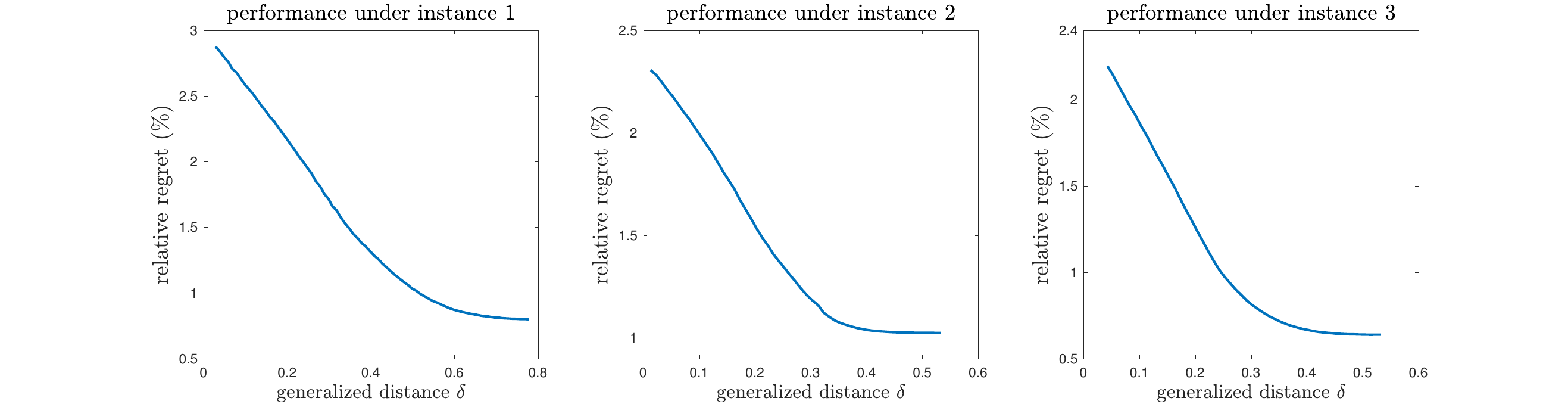}%
	\caption{$T=10^4$-period relative regret for the single-historical-price setting  with different $\delta$} 
	\label{figure-effect_delta}  
	\includegraphics[width=  0.9\textwidth, trim=0 0 0 0,clip]{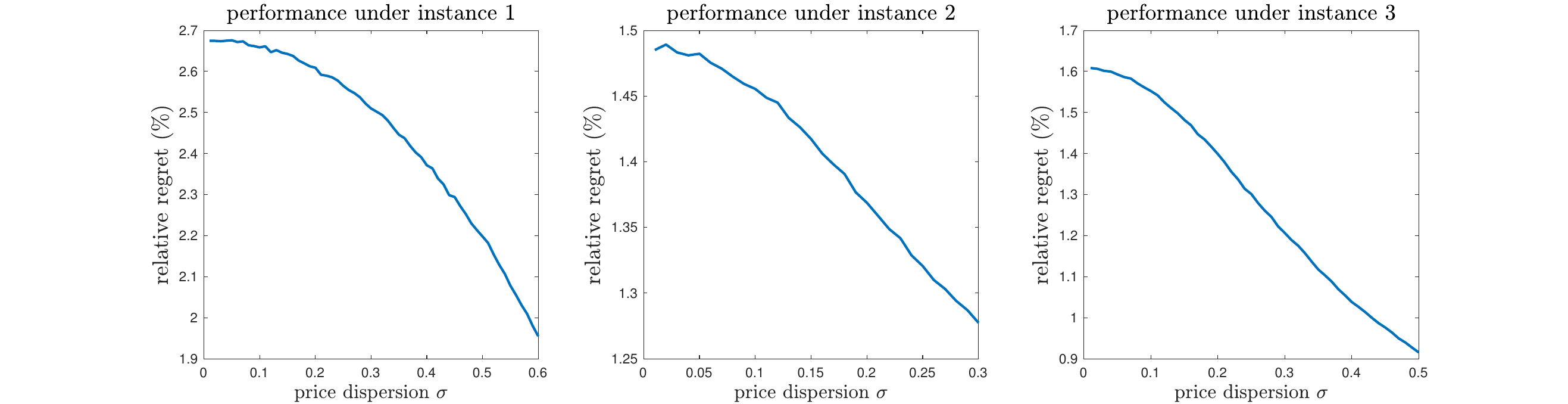}%
	\caption{$T=10^4$-period relative regret for the multiple-historical-price setting with different $\sigma$} 
	\label{figure-effect_sigma}
\end{figure}

Third, we investigate the effect of offline sample size $n$ on the empirical regret of O3FU algorithm. 
In Figure \ref{figure-effect_n-offline data}, we plot the relative regret of O3FU algorithm given different amount of offline data (with $n$ ranging from 20 to 12000), under the single-historical-price setting (with $\hat  p =1.8, 0.9, 1$ for instances (1)-(3) respectively).  The x-axis is depicted on a log scale.  We can see clearly that for each problem instance, as the offline sample size increases, the relative regret   decreases, 
which is consistent with the phase transitions implied from our theoretical results.

Finally, we investigate the effects of    generalized distance $\delta$ and price dispersion $\sigma$ on the empirical regret of our algorithm. Figure  \ref{figure-effect_delta}  shows the relative regret  of O3FU algorithm given $n=500$ offline demand observations under historical price $p^*+\delta$ with different $\delta$,  and Figure \ref{figure-effect_sigma} shows the relative regret of O3FU algorithm  
given 250 offline demand observations under historical price
$\bar{p}_{1:n}-\sigma$, and 250 offline demand observations under historical price $\bar{p}_{1:n}+\sigma$ with different $\sigma$, where $\bar{p}_{1:n}= 0.8, 0.8, 0.7$ for instances (1)-(3) respectively. As seen from Figure \ref{figure-effect_delta}   and \ref{figure-effect_sigma}, when $\delta$ or $\sigma$ increases, the empirical regret of our algorithm decays, which also matches the  \text{inverse-square law}. 

\begin{remark}
We  remark that the empirical evidence  for the phase transitions and inverse-square law is not always observed under every problem instance. This is because according to its definition through the supremum  over some instance-dependent environment class, the optimal regret 
should be attained at some ``hard"   instances, and so do its implications of the
phase transitions and inverse-square law.  Besides, when discussing the optimal regret rate and its implications,  we require $T$ to be
sufficiently large, and ignore all the constant factors. Our choices of instances (1)-(3) capture the aforementioned hard instances, and
also avoid that the problem falls into the regimes where constant factors significantly  affect the overall  regret rate. 
\end{remark}

	\section{Further Discussion: Offline Data and Self-Exploration}\label{subsec-self exploration} 
In  M-O3FU algorithm proposed in \S \ref{subsec-multiple-alg-upper bound}, there is a preliminary step  testing whether $\frac{\min_{\theta\in \mathcal C_0} |\bar{p}_{1:n}-\psi(\theta)|}{\max_{\theta_1,\theta_2\in \mathcal C_0} |\psi(\theta_1)-\psi(\theta_2)|}\le K$ holds or not. 
We  find that this step also  has an important implication  in practice:  $\frac{\min_{\theta \in \mathcal C_0} |\bar{p}_{1:n}-\psi(\theta)|}{\max_{\theta_1,\theta_2\in \mathcal C_0} |\psi(\theta_1)-\psi(\theta_2)|}> K$ 
 is actually a sufficient condition for \textit{self-exploration} in our \texttt{OPOD} problem. That is, with high probability, when this condition holds, the myopic (i.e.,  greedy) policy can  achieve the optimal regret  without any active exploration.

The myopic  policy is defined as follows. Let $\mathcal{C}_0=\{\theta \in \Theta^{\dag}: || \theta -   \theta_0^{\text{LS}}||^2_{V_{0,n}} \le w_0^2\}$, where $V_{0,n} = \lambda I+\sum_{i=1}^{n}[1\,\, \hat p_i]^{\top}[1\,\, \hat p_i]$, $\theta_0^{\text{LS}}= \arg\min_{\theta \in \Theta^{\dag}} \sum_{i=1}^{n}((\hat D_i-\alpha -\beta \hat p_i)^2+\lambda(\alpha^2+\beta^2))$, and $w_0=R\sqrt{2\log((T^2\vee n\sigma^2)(1+(1+u^2)n/\lambda))}+\sqrt{\lambda(\alpha_{\max}^2+\beta_{\min}^2)}$.  
	Let $\{p_t^{\text{myopic}}\}_{ t \ge 1}$ be the   sequence of prices charged by the myopic policy. For $t=1$,   $p_t^{\text{myopic}}=\psi(\theta_0^{\text{LS}})$, and  for each   $t\ge 2$, we first compute  the least-square estimator $\theta_{t-1}^{\text{LS}}$ based on offline data and all the available online data within   confidence ellipsoid $\mathcal{C}_0$: 
\begin{align*} 
\theta_{t-1}^{\text{LS}} = \arg\min_{\theta \in \mathcal{C}_0  
} \Big(\sum_{i=1}^{n}(\hat D_i-\alpha-\beta \hat p_i)^2+ \sum_{s=1}^{t-1}( D_s-\alpha-\beta p_s)^2+\lambda(\alpha^2+\beta^2)\Big), 
\end{align*}
and then let  
$p_t^{\text{myopic}}=\psi(\theta_{t-1}^{\text{LS}})$.  
  The next proposition shows that  the myopic policy is guaranteed to be optimal if  certain condition  
 holds.  
\begin{proposition}\label{prop-myopic alg}   
  Suppose {$n\sigma^2 \ge \sqrt T$}. Then with probability at least  $1-\frac{1}{T^2}\wedge\frac{1}{n\sigma^2}$, the following event holds: if  $\frac{\min_{\theta \in \mathcal C_0} |\bar{p}_{1:n}-\psi(\theta)|}{\max_{\theta_1,\theta_2\in \mathcal C_0} |\psi(\theta_1)-\psi(\theta_2)|}> K$ for some $K>1$,  then the myopic policy ensures that  the   regret is $\widetilde{\mathcal{O}}( \frac{T}{n\sigma^2+(n\wedge T)\delta^2})$. 
	\end{proposition}

{The intuition of Proposition \ref{prop-myopic alg} is as follows. Note that the key step to  prove the instance-dependent upper bound $\widetilde{\mathcal{O}}( \frac{T}{n\sigma^2+(n\wedge T)\delta^2})$ in Theorem~\ref{thm-upper bound-multiple prices-T}  is to show  events $\{U_{t,3}\}_{t=1}^T$ and $\{U_{t,4}\}_{t=2}^T$ in Lemma \ref{lemma-proof-multiple historical prices-step2} hold. 
	Since the myopic policy charges prices based on estimator $\theta_{t-1}^{\text{LS}} \in \mathcal{C}_0$ in each period $t$, and $\mathcal{C}_0$ contains  $\theta$ with high probability, under the condition in Proposition  \ref{prop-myopic alg},  we can easily verify that the myopic price $p_t^{\text{myopic}}$ is bounded away from  $\bar p_{1:n}$ by a distance proportional to $\delta$.  
 In other words, event $U_{t,3}$  in Lemma~\ref{lemma-proof-multiple historical prices-step2} is automatically satisfied  for each period $t$, and in this case,   when $\theta^* \in \mathcal{C}_t=\{\theta \in \Theta^{\dag}: ||\theta-\theta_{t}^{\text{LS}}||^2_{V_{t,n}} \le w_t^2\}$, we can further show that  event $U_{t,4}$ also holds. Therefore, the myopic policy ensures that the regret is   $\widetilde{\mathcal{O}}(\frac{T}{n\sigma^2+(n\wedge T)\delta^2})$. 
}

{We also make several remarks about Proposition~\ref{prop-myopic alg}. First, the interpretation of probability ``$1-\frac{1}{T^2}\wedge\frac{1}{n\sigma^2}$'' is similar to the interpretation of ``95\%'' in a 95\% confidence interval. Such a probabilistic statement is common in frequentist statistics, when one wants to make some inference (e.g., myopic policy is optimal or not) based on some empirical observations (e.g.,    $\frac{\min_{\theta \in \mathcal C_0} |\psi(\theta)-\bar{p}_{1:n}|}{\max_{\theta_1,\theta_2\in \mathcal C_0} |\psi(\theta_1)-\psi(\theta_2)|}> K$). Second,  if the regular case happens, i.e., $n\sigma^2 \gtrsim \sqrt T$ and $\delta^2 \gtrsim \frac{1}{n\sigma^2}$, one can easily verify that the empirical condition described in Proposition~\ref{prop-myopic alg} always holds. In this case, the myopic algorithm always ensures $\widetilde{\mathcal{O}}( \frac{T}{n\sigma^2+(n\wedge T)\delta^2})$ regret. Nevertheless, verifying the condition $\delta^2 \gtrsim \frac{1}{n\sigma^2}$ requires knowing the true parameter $\theta^*$ in advance, which is not practical in reality. Thus, we make a probabilistic statement in Proposition~\ref{prop-myopic alg} about the regret bound under an empirical condition that can be directly verified by the algorithm. 
	Third, the choice of $1- \frac{1}{T^2}\wedge\frac{1}{n\sigma^2}$  is not essential  in Proposition~\ref{prop-myopic alg}. In fact, one can achieve any higher   probability bound that is arbitrarily close to 1 by defining a larger confidence ellipsoid $\mathcal{C}_0$, although in that case, the condition $\frac{\min_{\theta \in \mathcal C_0} |\psi(\theta)-\bar{p}_{1:n}|}{\max_{\theta_1,\theta_2\in \mathcal C_0} |\psi(\theta_1)-\psi(\theta_2)|}> K$ will be more difficult to be satisfied.}

In reality, myopic policies are commonly adopted in many industries, since they are quite easy to explain to managers,  and relatively simple to implement in practice. See the discussion of   myopic policies in,  e.g. \cite{harrison2012bayesian,KeskinZeevi2014,qiang2016dynamic}. 
However, 
 due to the lack of active exploration,  myopic policies typically suffer  from \textit{incomplete learning}, thus usually have poor theoretical performance in dynamic pricing. Proposition~\ref{prop-myopic alg} shows how offline data may help myopic policies to achieve self-exploration in dynamic pricing:  when there are enough dispersive  offline data, then with high probability, as long as    $\bar{p}_{1:n}$ is bounded away from  offline confidence interval of $p^*$,  the  issue of incomplete learning could be resolved, and the myopic policy could  achieve self-exploration.

 	\section{Conclusion}\label{sec-conclusion}
 	In this paper, we investigate the   impact of  offline data on   online learning in  the context of dynamic pricing. 
 	In contrast to previous literature that involves only offline data   or only online data, we consider a more practical problem involving both offline data and online data, aiming to understand whether and how the pre-existence of offline data would benefit the online learning process. For both single-historical-price and multiple-historical-price settings, we  design a  learning algorithm based on the OFU principle with a provable instance-dependent regret upper bound, and establish a   regret lower bound that matches the upper bound up to  logarithmic factors.  
 	 Two important and nontrivial implications  implied by our results are \textit{phase transitions} and the \textit{inverse-square law},   characterizing the joint effect  of the size, location, and dispersion of the offline data on the  optimal regret. The numerical experiments demonstrate the effectiveness, robustness and stability of our algorithm, and reveal the empirical evidence for phase transitions
and the inverse-square law. Besides, we also develop a sufficient condition for the myopic policy to achieve the optimal regret in the regular case.  
 	
We discuss two extensions of this paper. First, while we focus on the linear demand model in this paper,  the regret upper bounds developed in Theorems~\ref{thm-UCB} and \ref{thm-upper bound-multiple prices-T} can be   extended to the generalized linear model $D_t=g(\alpha^*+\beta^* p_t)+\varepsilon_t$ for some link function $g(\cdot)$,  under certain smoothness conditions. In particular, these conditions guarantee that the regret in each single period $t$ for any given policy $\pi$ is of the same order as the quadratic estimation error $\mathbb{E}^{\pi}_{\theta^*}[(\psi(\theta^*)-p_t^{\pi})^2]$, and that     Lemmas~\ref{lemma-proof-UCB-step2} and \ref{lemma-proof-multiple historical prices-step2}  
continue to hold.  We refer the interested readers to online Appendix E for more details. Second, 	we assume that   historical prices are fixed constants in this paper.   
In reality,   offline pricing decisions can also be made based on the previous prices and sales observations according to some offline pricing  policy, in which case offline data will be generated in an adaptive way. By modifying the performance metric to the  expected regret conditioned on  the observed offline price trajectory, we can extend our results to the setting with adaptive offline data. This extension is discussed in online Appendix F. 
 	
 		This paper also suggests various    directions  for  future research.  
 		First, with the development of information technology,  firms have access to more detailed data that record customer   information and product characteristics. It will be interesting to  incorporate  such contextual information into the model, and study context-based dynamic pricing with online learning and offline data. In this case, it's important to understand how  the definition of the location metric of offline data should be modified accordingly. 
 		Second, we believe that the framework of online learning with offline data is quite general and widely applicable, and it will be also interesting  to explore how to extend  such a framework   to derive 
 		 new  results and insights for other data-driven operational problems, e.g.,  
 		  pricing under substitutable products,  bandit with knapsack constraints,   inventory control with demand learning, etc.  
 	Third, by leveraging the location metric of   offline data, this paper develops the    instance-dependent regret bound, which goes beyond the traditional    worst-case regret  and is new to the literature on dynamic pricing with demand learning. It will be valuable to explore whether   other types of instance-dependent bounds can be developed for dynamic pricing and revenue management problems by utilizing  certain historical information.

 	\bibliographystyle{informs2014}
 	\bibliography{ref}
 	
 	\ECSwitch

 	\begin{center}
 		\large{Online Appendix for ``Online Pricing with Offline Data:  Phase Transition and Inverse Square Law''} 
 	\end{center}
 	
 	\vspace{10pt}

\section*{Appendix A. Proofs of Statements in Section \ref{sec-UCB}}
 {\subsection*{A.1. Proof of Theorem \ref{thm-UCB}}}
 	 As preparations, we  introduce two results from \citealt{abbasi2011improved}, which will be used in  the analysis.  
 	 \begin{lemma}[Lemma 11 in  \citealt{abbasi2011improved}]\label{proof-UCB-lemma1}
 	 	Let $\{X_t: t\ge 1\}$ be a sequence in $\mathbb R^d$, $V$ be a $d\times d$ positive definite matrix and define $V_t=V+\sum_{s=1}^{t}X_sX_s^{\mathsf{T}}$. If $||X_t||_2\le L$ for all $t$ and $\lambda_{\min}(V) \ge \max\{1,L^2\}$, then 
 	 	\begin{align*}
 	 	\sum_{t=1}^{T} ||X_t||_{V_{t-1}^{-1}}^2 \le 2\Big(d\log\frac{\Tr(V)+TL^2}{d}-\log \det V\Big). 
 	 	\end{align*}
 	 \end{lemma} 
  
 	 \begin{lemma}[Theorem 2 in  \citealt{abbasi2011improved}]\label{UCB-concentration} 
 	 	For any $0<\epsilon<1$,    any $t \ge 1$, 
 	 	\begin{align*}
 	 	\prob\Bigg(||\theta^*-\hat \theta_s||_{V_{s,n}} \le R\sqrt{2\log \Big(\frac{1+(1+u^2)(s+n)/\lambda}{\epsilon}\Big)}+\sqrt{\lambda(\alpha_{\max}^2+\beta_{\min}^2)}, \forall 1 \le s \le t\Bigg) \ge 1-\epsilon. 
 	 	\end{align*} 
 	 \end{lemma}

 	 We now divide the proof for Theorem \ref{thm-UCB} into two steps  by proving  the instance-independent upper bound $\mathcal{O}(\sqrt{T} \log T)$ and the instance-dependent upper bound $\mathcal{O}(\frac{T (\log T)^2}{(n \wedge T)\delta^2})$.  
 	 
 	 \underline{\textbf{Step 1.}} In this step, we prove that the regret of   O3FU algorithm is $\mathcal{O}(\sqrt{T}\log  T)$.  
 	 Let $x_t=[1 \,\, p_t]^T$ for each $t \ge 1$. 
 	 For any $t \ge 2$, suppose $\theta^* \in \mathcal C_{t-1}$ (note that in this case, $\mathcal{C}_{t-1}\cap \Theta^{\dag}\neq \emptyset$, and thus, $\tilde \theta_t$ is well-defined), then we have 
 	 \begin{align}\label{proof-UCB-step1-ineq1}
\psi(\theta^*)(\alpha^*+\beta^* \psi(\theta^*))-p_t(\alpha^*+\beta^* p_t) 
 	 &\le 
 	 p_t(\tilde \alpha_t+\tilde \beta_t p_t)-p_t(\alpha^*+\beta^* p_t) \nonumber \\
 	 &\le u||x_t||_{V_{t-1,n}^{-1}} \cdot ||\tilde \theta_t-\theta^*||_{V_{t-1,n}} \nonumber  \\
 	 &\le 2u||x_t||_{V_{t-1,n}^{-1}} \cdot w_{t-1}
 	 \end{align}
 	 where the first inequality follows from the definition of $(p_t, \tilde{\theta}_t)$ in O3FU algorithm,  the second inequality follows from Cauchy-Schwarz  inequality, and the last inequality follows from $\theta^*, \tilde{\theta}_t \in \mathcal C_{t-1}$. Therefore, 
 	 \begin{align}\label{proof-UCB-step1-ineq2}
 	 \sum_{t=2}^{T}\big(\psi(\theta^*)(\alpha^*+\beta^* \psi(\theta^*))-p_t(\alpha^*+\beta^* p_t)\big)
 	& \le \sqrt{(T-1)\sum_{t=2}^{T}\big(\psi(\theta^*)(\alpha^*+\beta^*\psi(\theta^*))-p_t(\alpha^*+\beta^*p_t)\big)^2} \nonumber   \\
 	 & \le 2u\sqrt{(T-1) w_{T-1}^2 \sum_{t=2}^{T}||x_t||_{V_{t-1,n}^{-1}}^2},  
 	 \end{align}
 	 where the first inequality follows from Cauchy-Schwarz inequality, and the second inequality follows from   inequality \eqref{proof-UCB-step1-ineq1} and the fact that $w_t$ increases in $t$. 

 	 Then we use Lemma \ref{proof-UCB-lemma1} 
 	 to bound the term $\sum_{t=1}^{T}||x_t||_{V_{t-1,n}^{-1}}^2$.  To apply Lemma \ref{proof-UCB-lemma1},  
 	let  $d=2$, $L=\sqrt{1+u^2}$, $\lambda = 1+u^2$, 
 	 \begin{align*}
 	 X_t=\Big[
 	 \begin{matrix}
 	 1\\p_t
 	 \end{matrix}
 	 \Big], \quad V=\lambda I+n\Bigg[
 	 \begin{matrix} 
 	 1 & \hat p \\
 	 \hat p & \hat p^2
 	 \end{matrix}
 	 \Bigg],  \quad 
 	 V_t=V+\sum_{s=1}^{t}\Bigg[
 	 \begin{matrix} 
 	 1 & p_s \\
 	 p_s& p_s^2
 	 \end{matrix}
 	 \Bigg].
 	 \end{align*}
 	 Then we have 
 	 \begin{align*}
 	 \sum_{t=1}^{T}||x_t||_{V_{t-1,n}^{-1}}^2 
 	 &\le 2\Big(2\log\frac{(2\lambda+n(1+\hat p^2))+T(1+u^2)}{2}-\log \big(\lambda(\lambda+n(1+\hat p^2))\big)\Big)\\
 	 &\le  2\log\Big(\frac{(1+u^2)(2+n+T)^2}{4(1+l^2)(1+n)}\Big), 	 \end{align*}
 	 which, combined with   inequality \eqref{proof-UCB-step1-ineq2}, the definition of $w_T^2=\mathcal{O}(\log  T)$,
 	  implies that  when $\theta^* \in \mathcal C_{t-1}$ for any $t \ge 2$, 
 	 \begin{align}\label{ineq1-proof-thm2}
 	  \sum_{t=2}^{T}\big(\psi(\theta^*)(\alpha^*+\beta^* \psi(\theta^*))-p_t(\alpha^*+\beta^* p_t)\big)  = \mathcal{O}\big(\sqrt T \log T\big). 
 	 \end{align}   
 	Then  
 	  the regret of   O3FU  algorithm is upper bounded as follows: 
 	\begin{align*}
 	& \sum_{t=2}^{T} \mathbb{E}\big[r^*(\theta^*)-r(p_t;\theta^*)\big]
 	\\&\quad = 	\sum_{t=2}^{T} \mathbb{E}\Big[\big(r^*(\theta^*) -r(p_t;\theta^*)\big)\cdot 1_{\{\forall  2 \le s \le  t, \theta^* \in \mathcal{C}_{s}\}}\Big]+ 	\sum_{t=2}^{T} \mathbb{E}\Big[(-\beta^*)(\psi(\theta^*)-p_t)^2\cdot 1_{\{\exists 2 \le s \le  t,  \theta^* \notin \mathcal{C}_{s}\}}\Big] \\
 	& \quad  =   \mathcal{O}(\sqrt T \log T)+|\beta_{\min}|(u-l)^2\sum_{t=2}^{T}\frac{1}{T^2} \\
 	&\quad =\mathcal{O}(\sqrt T \log T),
 	\end{align*}
 	where the second identity follows from inequality \eqref{ineq1-proof-thm2} and  Lemma   \ref{UCB-concentration}   with $\epsilon=\frac{1}{T^2}$ for any $t \ge 2$.

 	 \underline{\textbf{Step 2.}} In this step, we prove that the regret of  O3FU algorithm is also $\mathcal{O}(\frac{T  (\log T)^2}{(n \wedge T) \delta^2})$. It suffices to show the case when $\delta\ge \frac{2\sqrt{\alpha_{\max}^2+\beta_{\max}^2}}{\sqrt 2\beta_{\max}^2}\cdot \frac{w_T}{n^{1/4}}$, since otherwise, 
 	 $\frac{T (\log T)^2}{(n \wedge T)\delta^2} \gtrsim \frac{T\sqrt{n}  (\log T)^2}{(n \wedge T) \log T} \gtrsim   \sqrt T \log T$, and the upper bound in Theorem~\ref{thm-UCB}  becomes $\mathcal{O}(\sqrt T \log T)$, which is already proven in   Step 1.

 	 Note that   
 it suffices to   bound the term $\sum_{t=2}^{T}\mathbb{E}[||\theta^*-\tilde \theta_t||^2]$.
 	 Since  $T_0$ defined in Lemma \ref{lemma-proof-UCB-step2} is an absolute constant, the result is trivial when $T \le T_0$. We then consider  $T \ge T_0$. 
 	 \begin{align*}
 	 \sum_{t=2}^{T}\mathbb{E}[||\theta^*-\tilde{\theta}_t||^2] &=\sum_{t=2}^{T}\mathbb{E}\Big[||\theta^*-\tilde{\theta}_t||^2\cdot 1_{\{\forall 2\le s \le t, \theta^* \in \mathcal C_s\}}\Big] +\sum_{t=2}^{T}\mathbb{E}\Big[||\theta^*-\tilde{\theta}_t||^2\cdot 1_{\{ \exists 2\le s\le t, \theta^* \notin \mathcal C_s\}}\Big]  \\
 	 &\le \sum_{t=2}^{T}\mathbb{E}\Big[||\theta^*-\tilde{\theta}_t||^2\cdot 1_{\{ U_{t,2}\}}\Big] +\sum_{t=2}^{T}\big((\alpha_{\max}-\alpha_{\min})^2+(\beta_{\max}-\beta_{\min})^2\big)\frac{1}{T^2} \\
 	 &\le  C_2\sum_{t=2}^{T}\frac{w_{t-1}^2}{(n \wedge (t-1))\delta^2} + ((\alpha_{\max}-\alpha_{\min})^2+  (\beta_{\max}-\beta_{\min})^2) \frac{1}{T},
 	 \end{align*}
 	 where the first inequality follows from the proof of Lemma \ref{lemma-proof-UCB-step2} and the concentration inequality in Lemma \ref{UCB-concentration} with $\epsilon=\frac{1}{T^2}$ for any $t \ge 2$.  
 It is easy to verify that when $n <  T$,   
 		\begin{align*}
 		\sum_{t=2}^{T}\frac{w_{t-1}^2}{(n \wedge (t-1))\delta^2}
 & = \sum_{t=1}^{n}\frac{w_t^2}{t \delta^2}+\sum_{t=n+1}^{T-1}\frac{w_t^2}{n \delta^2}   = \mathcal{O}\Big(\frac{(\log T)^2}{\delta^2}\Big)+\mathcal{O}\Big(\frac{T\log T}{n\delta^2}\Big) = \mathcal{O}\Big(\frac{T (\log T)^2}{(n \wedge T)\delta^2}\Big),
 		\end{align*}
 	and when  $n \ge  T$,  
 \begin{align*}
 \sum_{t=2}^{T}\frac{w_{t-1}^2}{(n \wedge (t-1))\delta^2}= \sum_{t=1}^{T-1}\frac{w_{t}^2}{t\delta^2} =\mathcal{O}\Big(\frac{\log T\log T}{\delta^2}\Big) = \mathcal{O}\Big(\frac{T ( \log T)^2}{(n \wedge T)\delta^2}\Big). 
 \end{align*}
 Combining both cases of $n <  T$ and  $n \ge  T$,  we have  $\sum_{t=2}^{T}\frac{w_{t-1}^2}{(n \wedge (t-1))\delta^2} %
 = \mathcal{O}(\frac{T(\log T)^2}{(n\wedge T)\delta^2})$, which completes the proof. 
 	 \qed

  \subsection*{A.2.  Proof of Lemma \ref{lemma-proof-UCB-step2}}
 	  	  When $t=1$, since $p_1=l\cdot\mathbb{I}\{\hat{p}>\frac{u+l}{2}\}+u\cdot\mathbb{I}\{\hat{p}\le\frac{u+l}{2}\}$, then $|p_1-\hat p| \ge \frac{u-l}{2}\ge \frac{1}{2}\delta$. Thus, when $t=1$,  $U_{t,1}$ holds.  
 	 
 	 	We next prove the following result: under the assumptions of Lemma \ref{lemma-proof-UCB-step2}, suppose  for each $1\le s \le t-1$ (for a fixed $2 \le t \le T$), the event $U_{s,1}$ holds, then  $U_{t,1}$ and $U_{t,2}$ also hold. To this end, let $\Delta \alpha_t=\tilde{\alpha}_t-\alpha^*$, $\Delta \beta_t=\tilde{\beta}_t-\beta^*$, and $\gamma_t=\frac{\Delta \alpha_t}{\Delta \beta_t}$ (when $\Delta \beta_t \neq 0$).  Since $\theta^* \in \mathcal C_{t-1}$  and $\tilde \theta_t \in \mathcal C_{t-1}$, we have 
 	 $
 	 ||\tilde{\theta}_t -\theta^*||^2_{V_{t-1,n}} \le 2\big(	||\tilde{\theta}_t -\hat \theta_{t-1}||^2_{V_{t-1,n}}  +	||{\theta}^* -\hat \theta_{t-1}||^2_{V_{t-1,n}}  \big) \le 2w_{t-1}^2$, which is equivalent to  
 	 \begin{align}\label{ineq1-proof-lemma8}
 	 \lambda\big((\Delta \alpha_t)^2+(\Delta\beta_t)^2\big)+n\big( \Delta \alpha_t+\Delta\beta_t\hat p \big)^2+\sum_{s=1}^{t-1}\big(\Delta \alpha_t+\Delta\beta_t p_s\big)^2 \le 2w_{t-1}^2.
 	 \end{align} 
 	 We next  divide the proof into three cases. 
 	 
 	 \underline{\textbf{Case 1}: $\Delta \beta_t=0$.} 
 	 In this case, \eqref{ineq1-proof-lemma8} becomes  $
 	 (\Delta \alpha_t)^2(\lambda+n+t-1) \le 2w_{t-1}^2,$ and
 	 \begin{align}\label{ineq8-proof-lemma8}
 	 ||\theta^*-\tilde{\theta}_t||^2 =(\Delta \alpha_t)^2+(\Delta \beta_t)^2= (\Delta \alpha_t)^2 \le \frac{2w_{t-1}^2}{ n+t-1}. 
 	 \end{align}
 	 Therefore, \eqref{ineq8-proof-lemma8}   implies that 
 	 \begin{align*}
 	  ||\theta^*-\tilde{\theta}_t||^2  \le \frac{2w_{t-1}^2}{n\wedge (t-1)} \le  \frac{2(u-l)^2w_{t-1}^2}{(n\wedge (t-1))\delta^2},
 	 \end{align*} 
 	 and  
 	 \begin{align*}
 	 |\hat p-p_t| 
 	 &\ge |\hat p-\psi(\theta^*)|- |p_t-\psi(\theta^*)| \\
 	 &\ge  |\hat p-\psi(\theta^*)|-\frac{\sqrt{\alpha_{\max}^2+\beta_{\max}^2}}{\sqrt{2}\beta_{\max}^2}\cdot \frac{w_{t-1}}{\sqrt{ n+t-1}}\\
 	 &\ge |\hat p-\psi(\theta^*)|-  \frac{|\psi(\theta^*)-\hat p|}{2n^{\frac{1}{4}}}  \\
 	 &\ge \frac{1}{2}\delta,
 	 \end{align*}
 	 where the second inequality follows from  \eqref{ineq8-proof-lemma8} and Lipschitz continuity of the function $\psi(\cdot)$: $
 	 |\psi(\theta_1)-\psi(\theta_2)| \le \frac{1}{2\beta_{\max}^2}\sqrt{\alpha_{\max}^2+\beta_{\max}^2}\cdot ||\theta_1-\theta_2||$,
 	 and the third inequality holds since  the assumption   $\delta\ge \frac{2\sqrt{\alpha_{\max}^2+\beta_{\max}^2}}{\sqrt 2\beta_{\max}^2}\cdot \frac{w_T}{n^{1/4}}$ implies
 	 \begin{align}\label{ineq9-proof-lemma8}
 	 \frac{w_{t-1}}{\sqrt{n+t-1}} \le \frac{w_{T}}{\sqrt{n}} \le \frac{\sqrt 2 \beta_{\max}^2 n^{\frac{1}{4}}}{2\sqrt{\alpha_{\max}^2+\beta_{\max}^2}\cdot \sqrt n} \delta. 
 	 \end{align}

 	 \underline{\textbf{Case 2}: $\Delta \beta_t\neq 0$, $|\gamma_t| \ge 4u+1$.} 
In this case, we have 
 	 \begin{align}\label{ineq2-proof-lemma8}
 	 ||\theta^*-\tilde{\theta}_t||^2 \le \frac{2w_{t-1}^2(1+\gamma_t^2)}{\lambda(1+\gamma_t^2)+n(\gamma_t+\hat p)^2+\sum_{s=1}^{t-1}(\gamma_t+p_s)^2}   \le \frac{2w_{t-1}^2(1+\gamma_t^2)}{ n(\gamma_t+\hat p)^2}   \le \frac{4w_{t-1}^2}{n},
 	 \end{align}
 	 where the first inequality holds since  $||\theta^*-\tilde \theta_t||^2=(\Delta \beta_t)^2(1+\gamma_t^2)$, and from \eqref{ineq1-proof-lemma8},  we have 
 	 $$
 	 (\Delta \beta_t)^2 \le \frac{2w_{t-1}^2}{\lambda(1+\gamma_t^2)+n(\gamma_t +\hat p)^2+\sum_{s=1}^{t-1}(\gamma_t + p_s)^2},$$  and the last inequality follows from  $ 1+\gamma_t^2\le 2(\gamma_t+\hat p)^2$, which is easily verified by noting $
 	 (\gamma_t+2\hat p)^2\ge( |\gamma_t|-2\hat p)^2 \ge (2\hat p+1)^2 \ge 2\hat p^2+1$. 
Then,  \eqref{ineq2-proof-lemma8} implies  that 
 	 \begin{align*}
 	 ||\theta^*-\tilde{\theta}_t||^2 \le \frac{4(u-l)^2w_{t-1}^2}{(n\wedge (t-1))\delta^2},
 	 \end{align*}
 	 and 
 	 \begin{align*}
 	 |\hat p-p_t| \ge |\hat p-\psi(\theta^*)|- |p_t-\psi(\theta^*)| \ge  |\hat p-\psi(\theta^*)|-\frac{\sqrt{\alpha_{\max}^2+\beta_{\max}^2}}{2\beta_{\max}^2} \frac{2w_{t-1}}{\sqrt{n}}\ge (1-\frac{\sqrt 2}{2})\delta,
 	 \end{align*}
 	 where the second inequality follows from Lipschitz continuity of $\psi(\cdot)$ and \eqref{ineq2-proof-lemma8}, and the third inequality follows from \eqref{ineq9-proof-lemma8}.

 	 \underline{\textbf{Case 3}: $\Delta \beta_t\neq 0$, $|\gamma_t| < 4u+1$.} 
 	Recall the  following definitions of $C_0$, $C_1$ and $T_0$ in Lemma \ref{lemma-proof-UCB-step2}: %
 	 \begin{align*}
 	 C_0=\frac{l|\beta_{\max}|}{u |\beta_{\min}|}, \quad C_1 = \frac{4(C_0+1)^2}{C_0^2}\big(1+(4u+1)^2\big), \quad  T_0=\min\Big\{t \in \mathbb{N}: w_{t} \ge  \frac{\sqrt C_1 \beta_{\max}^2}{\sqrt{2(\alpha_{\max}^2+\beta_{\max}^2)}} \Big\}. 
 	 \end{align*}
 	 
 	 \underline{\textbf{Subcase 3.1}: $1+\gamma_t^2 \le C_1 \frac{(\gamma_t+\hat p)^2}{\delta^2}$.} In this subcase, since 
 	 \begin{align*}
 	 ||\theta^*-\tilde{\theta}_t||^2 \le \frac{2w_{t-1}^2(1+\gamma_t^2)}{ n(\gamma_t+\hat p)^2} \le \frac{2C_1w_{t-1}^2}{ n\delta^2}, 
 	 \end{align*}
then  we have 
\begin{align*}
||\theta^*-\tilde{\theta}_t||^2  \le \frac{2C_1 w_{t-1}^2}{(n \wedge (t-1))\delta^2}. 
\end{align*}
In addition,  since $T \ge T_0$, it follows that 
 	 \begin{align*}
 	 |p_t-\hat p| 
 	 &\ge |\psi(\theta^*)-\hat p|-|p_t-\psi(\theta^*)| \\
 	 &\ge |\psi(\theta^*)-\hat p|-  \frac{\sqrt{\alpha_{\max}^2+\beta_{\max}^2}}{2\beta_{\max}^2}\frac{\sqrt{2C_1}w_{t-1}}{\sqrt{n}\delta} \\
 	 &\ge |\psi(\theta^*)-\hat p|-\frac{\sqrt C_1 \beta_{\max}^2}{2\sqrt{2(\alpha_{\max}^2+\beta_{\max}^2)}w_T}\delta \\
 	 &\ge \frac{1}{2}\delta,
 	 \end{align*}
 	 where in the third inequality, we utilize the fact that $\delta\ge \frac{2\sqrt{\alpha_{\max}^2+\beta_{\max}^2}}{\sqrt 2\beta_{\max}^2}\cdot \frac{w_T}{n^{1/4}}$, and the last inequality follows from  $T \ge T_0$ and the definition of $T_0$. 
 	 
 	 \underline{\textbf{Subcase 3.2}: $1+\gamma_t^2 > C_1 \frac{(\gamma_t+\hat p)^2}{\delta^2}$.}
 	 In this subcase,  we have  
 	 \begin{align*}
 	 ||\theta^*-\tilde{\theta}_t||^2   
 	 &\le  \frac{2w_{t-1}^2  ( \gamma_t^2+1)}{	 n( \gamma_t+ \hat p )^2+\sum_{s=1}^{t-1}(\gamma_t+ p_s)^2} \\
 	 &\le  \frac{4w_{t-1}^2  ( \gamma_t^2+1)}{\sum_{s=1}^{(t-1)\wedge n}( p_s-\hat p)^2} \\
 &\le \frac{4w_{t-1}^2((4u+1)^2+1)}{(n\wedge (t-1))\cdot \min\{(1-\frac{\sqrt 2}{2})^2, \frac{C_0^2}{4}\}\cdot \delta^2},
 	 \end{align*}
 	 where the second inequality holds since $n( \gamma_t+ \hat p )^2+\sum_{s=1}^{t-1}(\gamma_t+ p_s)^2 \ge \sum_{s=1}^{n \wedge (t-1)}\big((\gamma_t+ p_s)^2+( \gamma_t+ \hat p )^2\big)\ge \frac{1}{2}\sum_{s=1}^{n \wedge (t-1)}(p_s-\hat p)^2$, and  the last  inequality follows from $|\gamma_t|\le 4u+1$, and the  inductive assumption: for each $1\le s \le t-1$, $|p_s-\hat p| \ge \min\{1-\frac{\sqrt 2}{2}, \frac{C_0}{2}\}\cdot \delta$. 
 	 Now, it suffices to bound the term $|p_t-\hat p|$. 
 	 If we can prove the following inequality:
 	 \begin{align}\label{ineq5-proof-lemma8}  
 	 |\gamma_t+p_t|\ge C_0|\gamma_t+\psi(\theta^*)|,
 	 \end{align}
 	 then $|p_t-\hat p|$  can be bounded as follows:
 	 \begin{align*}
 	 |p_t-\hat p| 
 	 &\ge |p_t+ \gamma_t|-|\gamma_t+\hat p| \\
 	 &\ge C_0|\gamma_t+\psi(\theta^*)|-|\gamma_t+\hat p| \\
 	 &\ge C_0(|\psi(\theta^*)-\hat p|-|\gamma_t+\hat p|)-|\gamma_t+\hat p| \\
 	 &=C_0|\psi(\theta^*)-\hat p|-(C_0+1)|\gamma_t+\hat p| \\
 	 &\ge \Big(C_0-(C_0+1)\frac{\sqrt{1+(4u+1)^2}}{\sqrt{C_1}}\Big)|\psi(\theta^*)-\hat p|\\
 	 &\ge \frac{C_0}{2} \delta,
 	 \end{align*}
 	 where the second inequality follows from \eqref{ineq5-proof-lemma8}, the fourth inequality follows from the  assumption of Subcase 3.2, i.e., $1+\gamma_t^2 > C_1 \frac{(\gamma_t+\hat p)^2}{\delta^2}$ and $|\gamma_t| \le 4u+1$, and the last inequality follows from the definition of $C_1$.  
 	 
 	 Finally, we prove   inequality \eqref{ineq5-proof-lemma8}. We  define 
 	 \begin{align*}
 	 &A_1=p_t(\tilde{\alpha}_t+\tilde \beta_t p_t), \, \, A_2=p_t(\alpha^*+ \beta^* p_t),\,\, A_3=\psi(\theta^*)(\tilde{\alpha}_t+\tilde \beta_t \psi(\theta^*)), \,\, A_4=\psi(\theta^*)(\alpha^*+\beta ^*\psi(\theta^*)).
 	 \end{align*}
 	 Recall that  $p_t$ and $\psi(\theta^*)$ are the maximizers of the following  maximization problem: 
 	 \begin{align*}
 	 p_t = \arg\max_{p \in [l,u]} p(\tilde \alpha_t+\tilde \beta p), \quad  \psi(\theta^*)=\arg \max_{p \in [l,u]} p(\alpha^*+\beta^* p),
 	 \end{align*} 
 	 then we have the following relationships for $A_i$, $1\le i \le 4$: 
 	 \begin{align}
 	 &A_1 \ge A_3,   \label{ineq3-proof-lemma8} \\
 	 & A_1\ge A_4 \ge A_2.
 	\label{ineq4-proof-lemma8}
 	 \end{align}
 	To show   inequality   \eqref{ineq5-proof-lemma8}, we consider the following two cases when $A_3 \ge A_2$ and $A_3< A_2$.  If $A_3 \ge A_2$, then we have 
 	 \begin{align} \label{ineq6-proof-lemma8}
 	 |\Delta \alpha_t+\Delta \beta_t p_t| = \frac{ A_1-A_2 }{p_t}\ge \frac{|A_4-A_3| }{p_t} = \frac{\psi(\theta^*)}{p_t} |\Delta \alpha_t+\Delta \beta_t \psi(\theta^*)| \ge \frac{l}{u}  |\Delta \alpha_t+\Delta \beta_t \psi(\theta^*)|,
 	 \end{align} 
 	 where the first inequality follows from  $A_3, A_4 \in [A_2, A_1]$. 
 	Without loss of generality, we assume that $\Delta \beta_t > 0$, since otherwise,  we can redefine $\Delta \alpha_t$  and $\Delta \beta_t$ as $\alpha^*-\tilde{\alpha}_t$ and $\beta^*-\tilde{\beta}_t$
 	 respectively,  and the  proof will be similar.  
 	 Therefore, by dividing $\Delta \beta_t$ on both sides of \eqref{ineq6-proof-lemma8}, we get   inequality   \eqref{ineq5-proof-lemma8}.  If $A_3<A_2$, 
 	 \begin{align} \label{ineq7-proof-lemma8}
 	 |\Delta \alpha_t+\Delta \beta_t p_t| 
 	 &= \frac{ A_1-A_2 }{p_t}
 	 \ge \frac{A_4-A_2}{p_t}  =\frac{-\beta^*(\psi(\theta^*)-p_t)^2}{p_t}   =\frac{-\beta^* \psi(\theta^*)}{-\tilde \beta_t p_t}\cdot \frac{-\tilde \beta_t (\psi(\theta^*)-p_t)^2}{\psi(\theta^*)} \nonumber \\
 	 & \ge   \frac{l|\beta_{\max}|}{u |\beta_{\min}|}\cdot  \frac{A_1-A_3}{\psi(\theta^*)} \ge  \frac{l|\beta_{\max}|}{u |\beta_{\min}|} \cdot  \frac{A_4-A_3}{\psi(\theta^*)} =  \frac{l|\beta_{\max}|}{u |\beta_{\min}|}\cdot |\Delta \alpha_t+\Delta \beta_t \psi(\theta^*)|,
 	 \end{align}
 	 where the second identity and the second inequality follow from the  property of quadratic functions.  By dividing $\Delta \beta_t$ ($>0$ by assumption) on both sides of  \eqref{ineq7-proof-lemma8},   inequality \eqref{ineq5-proof-lemma8}   holds. It is also worth noting that from the above arguments, inequality \eqref{ineq5-proof-lemma8} holds universally due to the specific property of OFU principle and quadratic structure of the objective function, and does not depend on any inductive assumption.

 	 Combining Cases 1--3, we conclude that 
 	 \begin{align*}
 	 	& |p_t-\hat p| \ge \min\Big\{1-\frac{\sqrt 2}{2}, \frac{C_0}{2}\Big\}\cdot \delta,\\
 	& ||\theta^*-\tilde{\theta}_t||^2  \le \max\Big\{4(u-l)^2, 2C_1, \frac{4((4u+1)^2+1)}{\min\{\frac{C_0^2}{4}, (1-\frac{\sqrt 2}{2})^2\}}\Big\} \cdot \frac{w_{t-1}^2}{(n\wedge (t-1))\delta^2},
 	 \end{align*}
 	 i.e.,  $U_{t,1}$ and $U_{t,2}$ hold, which completes the  inductive arguments.   \qed

 	 	\subsection*{A.3. Proof of Theorem \ref{thm-lower bound}}

 	  {As preparation, we first present  the multivariate van Trees inequality, which will be used in Step~1 of the proof for Theorem \ref{thm-lower bound}.  For simplicity, we focus on the estimation problem for a real-valued function when stating the multivariate van Trees inequality,  which is sufficient for our use,   and we refer the interested readers to  \cite{gill2001applications} for  the more general  version on estimating a vector-valued function.  
 	 	\begin{lemma}[Multivariate van Trees Inequality, Theorem 1, \cite{gill2001applications}]\label{lemma-van trees ineq}
 	 		Consider estimating a real-valued function  $\psi(\theta)$ with    parameter
 	 		$\theta$ being an $s$-dimensional vector. Suppose we are given $n$ i.i.d. observations $X_1, X_2, \ldots, X_n$ drawn  from a common   distribution with probability density function  
 	 		$f(x, \theta)$.  Suppose $\theta$ is in the compact set $\Theta \subseteq \mathbb{R}^s$, the prior probability density function of $\theta$ is denoted by   $\lambda(\theta)$, 
 	 		and $C(\theta)$ is an $s$-dimensional row vector. If $f(x, \theta)$, $\lambda(\theta)$, $C(\theta)$ satisfy  certain regularity  conditions (see Assumptions in Section 4 of \cite{gill2001applications}), and in particular,  $\lambda(\theta)$ is positive in the interior of $\Theta$ and zero on its boundary, 
 	 			then for any estimator $\psi_n$ based on $X_1, X_2,  \ldots, X_n$, 
 	 			\begin{align*}
 	 			\mathbb{E}_{\lambda}\left[ \mathbb{E}_{\theta}\left[(\psi_n -\psi(\theta))^2\right]\right] \ge \frac{(\mathbb{E}_{ \lambda}[\Tr(C(\theta)(\frac{\partial \psi}{\partial \theta})^{\top})])^2}{ \widetilde{\mathcal{I}}(\lambda)+n  \cdot  \mathbb{E}_{\lambda}[\Tr(C(\theta) \mathcal{I}(\theta)(C(\theta))^{\top})]}, 
 	 			\end{align*}
 	 			where  $\Tr(A)$ denotes the trace for a square matrix $A$, and $\widetilde{\mathcal{I}}(\lambda)=\int_{\Theta} \Big( \sum_{k=1}^{s}\frac{\partial}{\partial \theta_k} \big(C_k(\theta) \lambda(\theta)\big)\Big) \frac{1}{\lambda(\theta)} d\theta$. 
 	 	
 	 	\end{lemma}  	 
 	 }

 	 It suffices to consider the case when $\varepsilon$ follows a normal distribution with standard deviation $R$.  Without loss of generality, we assume  $\xi=\frac{1}{2}$, and the analysis can   be easily extended to general $\xi \in (0,1)$.

 	 \underline{\textbf{Step 1.} }
 	 As the first step, we will prove the following result:  for  any pricing policy $\pi \in \Pi$, 
 	 \begin{align}\label{ineq1-proof-thm1}
 	 \sup_{ \theta \in \Theta_0(\delta)} R^{\pi}_{\theta}(T) =\Omega\Big( \big(\sqrt T \wedge  \big(\frac{T}{\delta^{-2}+(n \wedge T)\delta^2}\big)\big)\vee \log (1+T\delta^{2})\Big), 
 	 \end{align} 
 	 where   $\Theta_0(  \delta) = \big\{\theta \in \Theta^{\dag}: \, \psi(\theta)-\hat p \in [\frac{\delta}{2}, \delta]\big\}$. When   $\delta \ge \frac{lR}{16|\beta_{\min}|}  \sqrt{\frac{|\beta_{\max}|}{2K_0e}}T^{-\frac{1}{4}}(\log T)^{-\frac{1}{2}\lambda_0}$, the above \eqref{ineq1-proof-thm1} implies the desired lower bound in Theorem \ref{thm-lower bound}.  In what follows, we will   prove  three   lower bounds for $\sup_{ \theta \in \Theta_0(\delta)} R^{\pi}_{\theta}(T)$:  $\Omega(\log (1+T\delta^{2}))$, $\Omega(\sqrt{T} \wedge \frac{T}{\delta^{-2}+T\delta^2})$, and $\Omega(\sqrt{T} \wedge \frac{T}{\delta^{-2}+n\delta^2})$, which, when combined together, imply the lower bound in \eqref{ineq1-proof-thm1}.

 Before invoking the multivariate van Trees inequality in Lemma \ref{lemma-van trees ineq}, we first note that  since  $\Theta_0(\delta)= \{\theta \in \Theta^{\dag}: -\frac{\alpha}{2\hat p+\delta} \le \beta \le -\frac{\alpha}{2\hat p+2\delta}\}$, there exist some positive constants $x_0$, $y_0$, $\epsilon$  such that $\Theta_1(\delta):=[x_0-\frac{1}{2}\epsilon\delta, x_0+\frac{3}{2}\epsilon \delta] \times [-y_0-\frac{3}{2}\epsilon \delta, -y_0+\frac{1}{2}\epsilon\delta] \subseteq \Theta_0(\delta)$. Then we define a prior distribution for $\theta$ on $\Theta_1(\delta)$ as follows: 
 \begin{align}\label{def-prior density function}
 q(x,y)=\frac{1}{(\epsilon\delta)^2}  \cos^2\Big(\frac{\pi(x-\frac{2x_0+\epsilon \delta}{2})}{2\epsilon \delta}\Big) \cdot \cos^2\Big(\frac{\pi(y+\frac{2y_0+\epsilon \delta}{2})}{2\epsilon \delta}\Big), \, \forall (x,y)\in \Theta_1(\delta). 
 \end{align}
 In addition, we have the following inequality: 
 \begin{align}\label{lower bound-ineq2}
 \sup_{\theta \in \Theta_0(\delta)} R^{\pi}_{\theta}(T) 
 &\ge \sup_{\theta \in \Theta_1(\delta)} R^{\pi}_{\theta}(T) 
 = \sup_{\theta \in \Theta_1(\delta)} \sum_{t=1}^{T} (-\beta)\cdot \sum_{t=1}^{T}\mathbb{E}_\theta^{\pi}[(p_t-\psi(\theta))^2]  \nonumber \\
 &\ge |\beta_{\max}| \cdot \sum_{t=1}^{T}\mathbb{E}_{q}\big[\mathbb{E}_\theta^{\pi}[(p_t-\psi(\theta))^2] \big],
 \end{align}
 where the first inequality holds since $ \Theta_1(\delta) \subseteq \Theta_0(\delta)$, the   identity follows from the property of quadratic function and optimality of $\psi(\theta)$, and the second inequality holds since $q(\theta)$ is a probability density  distribution defined on $\Theta_1(\delta)$.  
 Note that the reason for which we consider a subset of $\Theta_0(\delta)$
 is that the Fisher information defined on the rectangle, i.e., $\Theta_1(\theta)$,  will be easier to calculate later.

   	 Then for each $t \ge 2$, by letting $n=1$, $X_1=(\hat \varepsilon_1, \ldots, \hat \varepsilon_n, \varepsilon_1, \ldots, \varepsilon_{t-1})$, $\psi_n=p_t$, $\lambda(\cdot)=q(\cdot)$ in Lemma~\ref{lemma-van trees ineq},  we have 
 	 \begin{align}\label{lower bound-ineq1}
 	 \mathbb{E}_{q}\big[\mathbb{E}_{\theta}^{\pi}[(p_t-\psi(\theta))^2]\big] \ge \frac{(\mathbb{E}_{q}[C(\theta)^{\mathsf{T}}\frac{\partial \psi}{\partial \theta}])^2}{{\mathcal{I}}(q)+\mathbb{E}_{ q}[\mathbb{E}_{\theta}^{\pi}[C(\theta)^{\mathsf{T}} \mathcal{I}_{t-1}^{\pi}(\theta)C(\theta)]]},
 	 \end{align}
 	 where $C(\theta)$ is any two-dimensional vector to be specified,  
 	 and
 	 \begin{align*}
 	 \mathcal{I}(q)=\int_{(\theta_1, \theta_2)\in\Theta_1(\delta)} \sum_{i=1}^{2} 
 	 \sum_{j=1}^{2}\frac{\partial}{\partial \theta_i}\big(C_i(\theta_1,\theta_2)\cdot  q(\theta_1,\theta_2)\big)\cdot \frac{\partial}{\partial \theta_j}\big(C_j(\theta_1,\theta_2)\cdot   q(\theta_1,\theta_2)\big) \cdot \frac{1}{ q(\theta_1, \theta_2)}d\theta_1 d\theta_2,
 	 \end{align*} 
 	 and $\mathcal{I}^{\pi}_{t-1}(\theta)$ is the Fisher information matrix defined as   \begin{align*} 
 	 \mathcal{I}_{t-1}^{\pi}(\theta)  
 	 = \frac{1}{R^2} \mathbb{E}_{\theta}^{\pi}\Bigg[
 	 \begin{matrix} 
 	 n+t-1 & n \hat p+\sum_{s=1}^{t-1}p_s \\
 	 n \hat p+\sum_{s=1}^{t-1}p_s & n \hat p^2+\sum_{s=1}^{t-1} p_s^2
 	 \end{matrix}
 	 \Bigg].
 	 \end{align*} 
We next start from   \eqref{lower bound-ineq1}, and  prove the three lower bounds by specifying different $C(\theta)$ and bounding the resulting  $\mathbb{E}_{q}[C(\theta)^{\mathsf{T}}\frac{\partial \psi}{\partial \theta}]$, $\mathcal{I}(q)$, and $\mathbb{E}_{ q}[\mathbb{E}_{\theta}^{\pi}[C(\theta)^{\mathsf{T}} \mathcal{I}_{t-1}^{\pi}(\theta)C(\theta)]]$ in the RHS of \eqref{lower bound-ineq1}.

\underline{To prove the first lower bound $\Omega(\log(1+T\delta^{2}))$}, let  $C(\theta)=(-\hat p, 1)$ in \eqref{lower bound-ineq1}, then we have 
 	 \begin{align*}
 	 \sum_{t=1}^{T} \mathbb{E}_{  q}\big[\mathbb{E}_{\theta}^{\pi}[(p_t-\psi(\theta))^2]\big] \ge \sum_{t=2}^{T} \frac{R^2c_1}{{R^2\mathcal{I}}( q)+\sum_{s=1}^{t-1}\mathbb{E}_{  q}[\mathbb{E}_{\theta}^{\pi}[(p_s-\hat p)^2]]} \ge \sum_{t=2}^{T} \frac{R^2c_1}{{R^2\mathcal{I}}(q)+(t-1)(u-l)^2}, 
 	 \end{align*}
 	 where $c_1=\big(\min_{\theta \in \Theta^{\dag}} \frac{\alpha+\beta \hat p}{2\beta^2}\big)^2$.  Since $C(\theta)=(-\hat p, 1)$ is independent of $\theta$, by changing variables in the integrals, we have 
 	 \begin{align*}
 	 \mathcal{I}(q)
 	 &=\frac{\pi}{2\epsilon^2\delta^2}\int_{-\frac{\pi}{2}}^{\frac{\pi}{2}} \int_{-\frac{\pi}{2}}^{\frac{\pi}{2}}  \sum_{i=1}^{2} 
 	 \sum_{j=1}^{2}\frac{\partial}{\partial \theta_i}\big( C_i(\theta_1,\theta_2)\cdot  \tilde q(\theta_1,\theta_2)\big)\cdot \frac{\partial}{\partial \theta_j}\big(C_j(\theta_1,\theta_2)\cdot   \tilde q(\theta_1,\theta_2)\big) \cdot \frac{1}{ \tilde q(\theta_1, \theta_2)}d\theta_1 d\theta_2,
 	 \end{align*}
 	 where  $\tilde q(\theta_1,\theta_2)=\cos^2 (\theta_1)\cdot\cos^2(\theta_2)$. Since  the  integral  in the RHS of the above equation is a constant independent of $\delta$, we have  $
 	 \mathcal{I}(q)=\Theta(\delta^{-2})
 	 $, it then follows from \eqref{lower bound-ineq2} that   
 	 \begin{align*} 
 	 \sup_{\theta  \in \Theta_0(\delta)}  
 	 \sum_{t=1}^{T}R_{\theta}^{\pi}(T) 
 \ge |\beta_{\max}| \cdot  \sup_{\theta  \in \Theta_1(\delta)}\sum_{t=1}^{T}\mathbb{E}_{\theta }^{\pi}[(p_t-\psi(\theta))^2]   
 	 = \Omega(\log (1+T\delta^{2})). 
 	 \end{align*} 
 	 
\underline{To   prove the second lower bound $\Omega(\sqrt T \wedge \frac{T}{\delta^{-2}+T\delta^2})$}, let $C(\theta)=(-\hat p,1)$ in \eqref{lower bound-ineq1} again, then we obtain 
 	 \begin{align}\label{lower bound-ineq9}
 	 \sum_{t=1}^{T} \mathbb{E}_{  q}\big[\mathbb{E}_{\theta}^{\pi}[(p_t-\psi(\theta))^2]\big] 
 	 &\ge \sum_{t=2}^{T} \frac{R^2c_1}{R^2{\mathcal{I}}( q)+\sum_{s=1}^{t-1}\mathbb{E}_{  q}[\mathbb{E}_{\theta}^{\pi}[(p_s-\hat p)^2]]} \nonumber  \\
 	 &\ge  \sum_{t=2}^{T} \frac{R^2c_1}{{R^2\mathcal{I}}( q)+2(t-1)\delta^2+\sum_{s=1}^{t-1}\mathbb{E}_{  q}[\mathbb{E}_{\theta}^{\pi}[(p_s-\psi(\theta))^2]]}  \nonumber \\
 	 &\ge  \frac{R^2c_1(T-1)}{{R^2\mathcal{I}}( q)+2T\delta^2+\sum_{t=1}^{T}\mathbb{E}_{  q}[\mathbb{E}_{\theta}^{\pi}[(p_t-\psi(\theta))^2]]}, 
 	 \end{align}
 	 where  the second inequality holds since  $(p_s-\hat p)^2 \le 2(p_s-\psi(\theta)))^2+2(\hat p-\psi(\theta))^2 \le 2(p_s-\psi(\theta))^2+2\delta^2$.  It is easily verified that the inequality $x^2+bx+c\ge 0$ for $b>0, c<0, x\ge 0$ implies 
 	 \begin{align}\label{lower bound-ineq6}
 	 x\ge \frac{1}{\sqrt 2+1}\min\big\{\sqrt{|c|}, \frac{2|c|}{b}\big\}. 
 	 \end{align}
 	 Applying \eqref{lower bound-ineq6} to the inequality \eqref{lower bound-ineq9}, we obtain from \eqref{lower bound-ineq2} that 
 	 \begin{align*}%
 	 \sup_{\theta  \in \Theta_0(\delta)}  
 	\sum_{t=1}^{T}R_{\theta}^{\pi}(T)  \ge |\beta_{\max}| \cdot  \sum_{t=1}^{T} \mathbb{E}_{q}\big[\mathbb{E}_{\theta}^{\pi}[(p_t-\psi(\theta))^2]\big]  \ge \Omega(\sqrt{T}\wedge \frac{T}{\mathcal{I}(q)+T\delta^2}) =\Omega(\sqrt{T}\wedge \frac{T}{\delta^{-2}+T \delta^2}),
 	 \end{align*}
 	 where in the identity, we utilize the fact that $
 	 \mathcal{I}(q) =\Theta(\delta^{-2})$. 
 	 
 	 \underline{To prove the third lower  bound  $\Omega(\sqrt T \wedge\frac{T}{\delta^{-2}+n\delta^2})$}, we choose another vector  $C(\theta)=(-\psi(\theta),1)$, and the inequality \eqref{lower bound-ineq1} becomes   
 	 \begin{align*}
 	 \sum_{t=1}^{T}\mathbb{E}_{  q}\big[\mathbb{E}_{\theta}^{\pi}[(p_t-\psi(\theta))^2] \big]  
 	 &\ge \sum_{t=2}^{T}\frac{R^2\alpha_{\min}^2/(4\beta_{\min}^2)^2}{R^2{\mathcal{I}}(q)+\sum_{s=1}^{t-1}\mathbb{E}_{q}[\mathbb{E}_{\theta}^{\pi}[(p_s-\psi(\theta))^2]]+n\delta^2} \\
 	 &\ge \frac{R^2\alpha_{\min}^2/(4\beta_{\min}^2)^2(T-1)}{R^2{\mathcal{I}}(q)+\sum_{t=1}^{T}\mathbb{E}_{q}[\mathbb{E}_{\theta}^{\pi}[(p_t-\psi(\theta))^2]]+n\delta^2},  
 	 \end{align*}
 	 which, combined with  \eqref{lower bound-ineq2} and  \eqref{lower bound-ineq6},  implies that 
 	 \begin{align} \label{ineq-lower bound-thm2-Fisher}
 	 \sup_{\theta \in \Theta_0(\delta)} R^{\pi}_{\theta}(T) \ge |\beta_{\max}|\sum_{t=1}^{T} \mathbb{E}_{q}\big[\mathbb{E}_{\theta}^{\pi}[(p_t-\psi(\theta))^2]\big]  \ge \Omega(\sqrt{T}\wedge \frac{T}{\mathcal{I}(q)+n\delta^2}).
 	 \end{align} 
 By definition, 
 	 \begin{align*}
 	 \mathcal{I}(q)
 	 &=\frac{\pi}{2\epsilon^2\delta^2}\int_{-\frac{\pi}{2}}^{\frac{\pi}{2}} \int_{-\frac{\pi}{2}}^{\frac{\pi}{2}}  \sum_{i=1}^{2} 
 	 \sum_{j=1}^{2}\frac{\partial}{\partial \theta_i}\big( \tilde C_i(\theta_1,\theta_2)\cdot  \tilde q(\theta_1,\theta_2)\big)\cdot \frac{\partial}{\partial \theta_j}\big(\tilde C_j(\theta_1,\theta_2)\cdot   \tilde q(\theta_1,\theta_2)\big) \cdot \frac{1}{ \tilde q(\theta_1, \theta_2)}d\theta_1 d\theta_2,
 	 \end{align*}
 	 where $\tilde C(\theta_1, \theta_2)=\big(\frac{\frac{2\epsilon \delta \theta_1}{\pi}+x_0+\frac{\epsilon \delta}{2}}{2(\frac{2\epsilon \delta \theta_2}{\pi}-y_0-\frac{\epsilon \delta}{2})}, 1\big)$. Since both $\tilde C(\cdot)$ and $C(\theta)$ are bounded by constants independent of $\delta$, it is easily verified that the integral in the RHS of the above identity can be  bounded by constant independent of $\delta$. Therefore,  $\mathcal{I}(q)=\Theta(\delta^{-2})$, and from  inequality \eqref{ineq-lower bound-thm2-Fisher}, we   have 
 	 \begin{align*}
 	 	\sup_{\theta \in \Theta_0(\delta)} R^{\pi}_{\theta}(T)  = \Omega(\sqrt T \wedge \frac{T}{\delta^{-2}+n\delta^2}). 
 	 \end{align*}

 	 \underline{\textbf{Step 2.}} In this step, we complete the proof by showing that when $\delta \le \frac{lR}{16|\beta_{\min}|}  \sqrt{\frac{|\beta_{\max}|}{2K_0e}}T^{-\frac{1}{4}}(\log T)^{-\frac{1}{2}\lambda_0}$,  for any admissible policy $\pi \in \Pi^{\circ}$, there exists $\theta \in \Theta^{\dag}$ satisfying  $|\psi(\theta)-\hat p| \in [\frac{1}{2}\delta, \frac{3}{2}\delta]$ such that
 	 \begin{align}\label{eq1-proof-thm1}
 	 R^{\pi}_{\theta}(T) = \Omega\Big(\frac{\sqrt T}{(\log T)^{\lambda_0}}\Big).
 	 \end{align} 
 	 
 	 Our proof of \eqref{eq1-proof-thm1} is based on the concept of  KL divergence, which is 
 	 a quantitative measure of distance between two distributions. The definition is given as follows. For any two distributions $P_1$ and $P_2$, the KL divergence is 
 	 \begin{align*}
 	 KL(P_1,P_2)=\mathbb{E}_{X \sim P_1}\Big[\log\frac{P_1(X)}{P_2(X)}\Big]. 
 	 \end{align*}   
 	 We now consider   two vectors of   demand parameters $\theta_1$ and $\theta_2$ satisfying the following conditions:
 	 \begin{align*}
 	 -\frac{\alpha_1}{2\beta_1}=\hat p+\delta, \quad  -\frac{\alpha_2}{2\beta_2}=\hat p+\delta+\Delta, \quad (\alpha_1-\alpha_2)+(\beta_1-\beta_2)\hat p=0.
 	 \end{align*} 
 	 where $\Delta>0$ is  to be determined. 
 	 For any  policy $\pi \in \Pi^{\circ}$, let  
 	 $P_1^{\pi}$ and $P_2^{\pi}$  be the following two  probability measures  induced by the common policy $\pi$ and  two parameters $\theta_1$ and $\theta_2$ respectively: 
 	 \begin{align*}
 	 P_i^{\pi}(\hat D_1, \ldots, \hat D_n,D_1, \ldots, D_T) =\prod_{t=1}^n \Big(\frac{1}{R}\phi\big(\frac{\hat D_t-(\alpha_i+\beta_i \hat p)}{R}\big)\Big) \cdot \prod_{t=1}^T \Big(\frac{1}{R}\phi\big(\frac{D_t-(\alpha_i+\beta_i p_t)}{R}\big)\Big), \, i=1,2,
 	 \end{align*}
 	 where $\phi(x)=\frac{1}{\sqrt{2\pi}}\exp(-\frac{x^2}{2})$ is the probability density function of the standard normal distribution. From the definition of KL divergence, we have 
 	 \begin{align}\label{proof-thm1-samll delta-eq4}
 	 KL(P_1^{\pi}, P_2^{\pi}) 
 	 &  = \frac{(\beta_1-\beta_2)^2}{2R^2} \Big(n\big(\frac{\alpha_1-\alpha_2}{\beta_1-\beta_2}+\hat p\big)^2+ \sum_{t=1}^{T} \mathbb{E}_{\theta_1}^{\pi}[(\frac{\alpha_1-\alpha_2}{\beta_1-\beta_2}+p_t)^2]  
 	 \Big)   \nonumber \\
 	 & =\frac{2\beta_2^2\Delta^2}{(\hat p+2\delta)^2R^2}
 	 \sum_{t=1}^{T}\mathbb{E}_{\theta_1}^{\pi}[(p_t-\hat p)^2]  \nonumber \\
 	 &\le \frac{4\beta_{\min}^2\Delta^2}{l^2R^2}
 	 \big(\sum_{t=1}^{T}\mathbb{E}_{\theta_1}^{\pi}[(p_t-\psi(\theta_1))^2]+T\delta^2\big).
 	 \end{align} 
 	 Since  $
 	 R^{\pi}_{\theta_1}(T)=(-\beta_1)\sum_{t=1}^{T}\mathbb{E}_{\theta_1}^{\pi}[(p_t- \psi(\theta_1))^2]$, it follows that 
 	 \begin{align}\label{proof-thm1-samll delta-eq3}
 	 R^{\pi}_{\theta_1}(T) 
 	 & \ge |\beta_{\max}|  \Big(\frac{ l^2R^2}{4\beta_{\min}^2\Delta^2} KL(P_1^{\pi}, P_2^{\pi}) -T\delta^2\Big).
 	 \end{align}
 	 
 	 We next  establish a lower bound on $KL(P_1^{\pi},P_2^{\pi})$ and choose a suitable $\Delta$ such  that the RHS of \eqref{proof-thm1-samll delta-eq3} can be further lower bounded by $ \Omega(\frac{\sqrt T}{(\log T)^{\lambda_0}})$.  
 	 Before proceeding, we define two disjoint  intervals  $I_1=[\hat p+\delta-\frac{1}{4}\Delta,\hat p+\delta+\frac{1}{4}\Delta]$, and $I_2=[\hat p+\delta+\frac{3}{4}\Delta,\hat p+\delta+\frac{5}{4}\Delta]$. For each $t \ge 1$, let $X_t$ be the following Bernoulli random variable: $X_t=1$ if $p_t \in I_1$ and $X_t=0$ otherwise. Then we have 
 	 \begin{align}\label{proof-thm1-samll delta-eq2}
 	 R^{\pi}_{\theta_1}(T)+R^{\pi}_{\theta_2}(T) 
 	 &\ge |\beta_{\max}| \sum_{t=1}^{T}\mathbb{E}_{\theta_1}^{\pi}[(p_t-\psi(\theta_1))^2] +|\beta_{\max}| \sum_{t=1}^{T}\mathbb{E}_{\theta_2}^{\pi}[(p_t-\psi(\theta_2))^2] \nonumber  \\& \ge \frac{1}{16}|\beta_{\max}|\Delta^2 \sum_{t=1}^{T}\big(P_1^{\pi}(p_t \notin I_1)+P_2^{\pi}(p_t \notin I_2)\big) \nonumber \\
 	 &\ge \frac{1}{16}|\beta_{\max}|\Delta^2 \sum_{t=1}^{T}\big(P_1^{\pi}(X_t=0)+P_2^{\pi}(X_t=1)\big) \nonumber \\ 
 	 &\ge \frac{1}{32}|\beta_{\max}| \cdot e^{-KL(P_1^{\pi}, P_2^{\pi})}\cdot T\Delta^2,
 	 \end{align}
 	 where the third inequality holds since $I_1$ and $I_2$ are disjoint, and 
 	 the last inequality follows from the
 	 Bretagnolle-Huber inequality  (Theorem 2.2 in \cite{Tsybakov2009}). 
 	 Since $\pi \in \Pi^{\circ}$,    
 	 \begin{align*}
 	 R^{\pi}_{\theta_1}(T)+R^{\pi}_{\theta_2}(T)  \le 2K_0\sqrt T (\log T)^{\lambda_0}, 
 	 \end{align*}
 	 which together with  inequality \eqref{proof-thm1-samll delta-eq2} implies 
 	 \begin{align*}
 	 KL(P_1^{\pi},P_2^{\pi}) \ge \log(\sqrt T \Delta^2)+\log\big(\frac{|\beta_{\max}|}{64K_0}\big)-\lambda_0\log \log T. 
 	 \end{align*}
 	 Thus,  combining the above inequality with   \eqref{proof-thm1-samll delta-eq3} and  letting $\Delta^2=  \frac{64K_0e(\log T)^{\lambda_0}}{|\beta_{\max}|\sqrt T}$, we have 
 	 \begin{align*}
 	 R^{\pi}_{\theta_1}(T) 
 	 & \ge  |\beta_{\max}| \Bigg(\frac{l^2R^2}{4\beta_{\min}^2\Delta^2}\Big(\log(\sqrt T \Delta^2)+\log\big(\frac{|\beta_{\max}|}{64K_0}\big)-\lambda_0\log \log T\Big)-T \delta^2 \Bigg) \\
 	 &= |\beta_{\max}| \Big( \frac{l^2R^2|\beta_{\max}|}{256\beta_{\min}^2 K_0e}\cdot  \frac{\sqrt T}{(\log T)^{\lambda_0}}-T\delta^2\Big) \\
 	 & \ge  \frac{l^2R^2\beta_{\max}^2}{512\beta_{\min}^2K_0e} \cdot  \frac{\sqrt T}{(\log T)^{\lambda_0}},
 	 \end{align*} where the second inequality follows from the choice of $\Delta$ and $\delta \le \frac{lR}{16|\beta_{\min}|}  \sqrt{\frac{|\beta_{\max}|}{2K_0e}}T^{-\frac{1}{4}}(\log T)^{-\frac{1}{2}\lambda_0}$. %
 Thus, $R^{\pi}_{\theta_1}(T)=\Omega(\frac{\sqrt T}{(\log T)^{\lambda_0}})$.

 	 Combining Step 1 and Step 2, we  conclude that for any admissible policy $\pi \in \Pi^{\circ}$, there exists  $\theta \in \Theta^{\dag}$ satisfying $|\psi(\theta)-\hat p| \in [\frac{1}{2}\delta,\frac{3}{2} \delta]$, such that 
 	 \begin{align*}
 	 R^{\pi}_{\theta}(T)=
 	 \left\{
 	 \begin{array}{ll}
 	 \Omega\big( (\sqrt T\wedge \frac{T}{ (n \wedge T) \delta^2})\vee \log T\big), & \text{if } \delta > \frac{lR}{16|\beta_{\min}|}  \sqrt{\frac{|\beta_{\max}|}{2K_0e}}T^{-\frac{1}{4}}(\log T)^{-\frac{1}{2}\lambda_0};\\
 	 \Omega\big((T\delta^2 )\vee \frac{\sqrt T}{(\log T)^{\lambda_0}} \big), & \text{if } \delta \le  \frac{lR}{16|\beta_{\min}|}  \sqrt{\frac{|\beta_{\max}|}{2K_0e}}T^{-\frac{1}{4}}(\log T)^{-\frac{1}{2}\lambda_0},
 	 \end{array}  
 	 \right.
 	 \end{align*}  
 	 which completes the proof of Theorem \ref{thm-lower bound}. 
 	 \qed

 	 \section*{Appendix B. Proofs of Statements in Section \ref{sec-extension-historical price}}
	
{\subsection*{B.1.  Proof of Theorem \ref{thm-upper bound-multiple prices-T}}
} 
To prove Theorem \ref{thm-upper bound-multiple prices-T},  we first show that    O3FU algorithm (after a natural modification to the multiple-historical-price setting) achieves the regret upper bound $\mathcal{O}(\sqrt{T}\log T)$ and  $\mathcal{O}\big(\frac{T (\log T)^2}{n \sigma^2+(n \wedge T)\delta^2}+1\big)$ in the following Step 1 and Step 2 respectively. Then  in Step 3, we use the results in Steps 1-2 to show that   M-O3FU algorithm achieves the desired upper bound.

 	 \underline{\textbf{Step 1.}} In this step, we prove the regret upper bound $\mathcal{O}(\sqrt T \log T)$ for   O3FU algorithm. Lemma \ref{UCB-concentration}  and   inequalities \eqref{proof-UCB-step1-ineq1} and \eqref{proof-UCB-step1-ineq2}  continue to hold by replacing each  $V_{t-1,n}$ with $\lambda I+ \sum_{i=1}^{n}[1\,\, \hat p_i]^{\top}[1\,\, \hat p_i]+\sum_{s=1}^{t-1}[1\,\,   p_s]^{\top}[1\,\, p_s]$. To apply Lemma \ref{proof-UCB-lemma1} to the RHS of the inequality \eqref{proof-UCB-step1-ineq2}, we just let $d=2$, $L=\sqrt{1+u^2}$, $\lambda = 1+u^2$, 
 	 \begin{align*}
 	 X_t=\Big[
 	 \begin{matrix}
 	 1\\p_t
 	 \end{matrix}
 	 \Big], \quad V=\lambda I+\sum_{i=1}^{n}\Bigg[
 	 \begin{matrix} 
 	 1 & \hat p_i \\
 	 \hat p_i & \hat p_i^2
 	 \end{matrix}
 	 \Bigg],  \quad 
 	 V_t=V+\sum_{s=1}^{t}\Bigg[
 	 \begin{matrix} 
 	 1 & p_s \\
 	 p_s& p_s^2
 	 \end{matrix}
 	 \Bigg].
 	 \end{align*}
 	 Then we get 
 	 \begin{align}\label{ineq-1-proof-thm4}
 	 \sum_{t=1}^{T}||x_t||_{V_{t-1,n}^{-1}}^2 
 	 &\le 2\Big(2\log\frac{(2\lambda+\sum_{i=1}^{n}(1+\hat p_i^2))+T(1+u^2)}{2}-\log \big(\lambda(\lambda+\sum_{i=1}^{n}(1+\hat p_i^2))\big)\Big) \nonumber \\
 	 &\le 2\log \Big(\frac{(1+u^2)(2+n+T)^2}{4(1+l^2)(1+n)}\Big).
 	 \end{align}
 	 The remaining proof remains the same as Theorem \ref{thm-UCB}, and is therefore omitted.
 	 
 	 \underline{\textbf{Step 2.}} In this step, we prove the  regret upper bound  $\mathcal{O}\big(\frac{T( \log T)^2}{n \sigma^2+(n \wedge T)\delta^2}+1\big)$ for   O3FU  algorithm. 
 	 Note that it suffices to consider the case when $n\sigma^2+(n\wedge T)\delta^2 =  \Omega(\sqrt T \log T)$, since otherwise, $\mathcal{O}((\sqrt T \log T)\wedge \frac{T(\log T)^2}{n\sigma^2+(n \wedge T)\delta^2}) = \mathcal{O}(\sqrt T \log T)$, which is already proven in Step 1. 
Under the assumption  $n\sigma^2+(n\wedge T)\delta^2 = \Omega(\sqrt T \log T)$, we 
 	consider the following two cases: 
 	 (1) $n \sigma^2 \lesssim (n\wedge T)\delta^2$;  
 	 (2) $n \sigma^2 \gtrsim (n\wedge T)\delta^2$.  

 	\underline{\textbf{Case 1.} $n \sigma^2 \lesssim (n\wedge T)\delta^2$.}  
 	In this case, the following three inequalities  hold:
 (i) $n\delta^2 \gtrsim \sqrt T \log T$; 	(ii) $\sigma \lesssim \delta$;  and (iii)  $\delta \gtrsim T^{-1/4}(\log T)^{\frac{1}{2}}$. The reason is as follows. Suppose (i) does not hold, we have $n\sigma^2+(n\wedge T)\delta^2  \lesssim \sqrt T \log T$, leading to contradiction with $n\sigma^2+(n\wedge T)\delta^2 = \Omega(\sqrt T \log T)$. Suppose (ii) does not hold,   then we have  $n\sigma^2  \gtrsim n\delta^2 \gtrsim (n\wedge T)\delta^2$, leading to contradiction with the assumption of Case 1. Finally, suppose (iii) does not hold, then we have  $(n\wedge T)\delta^2 \lesssim   \sqrt T\log T$,  leading to contradiction with  $n\sigma^2+(n\wedge T)\delta^2 = \Omega(\sqrt T \log T)$.    
 	Thus, when Case 1 happens, the conditions of   Lemma \ref{lemma-proof-multiple historical prices-step2}, i.e., $\sigma \lesssim \delta$  and  $ \delta \gtrsim \max\{{T^{\frac{1}{4}}w_Tn^{-\frac{1}{2}}}, T^{-\frac{1}{4}}\}$,  are satisfied. 
	By applying Lemma~\ref{lemma-proof-multiple historical prices-step2}, we have  
	\begin{align*}
	\sum_{t=2}^{T}\mathbb{E}[||\theta^*-\tilde{\theta}_t||^2] &=\sum_{t=2}^{T}\mathbb{E}\Big[||\theta^*-\tilde{\theta}_t||^2\cdot 1_{\{\forall 2\le s \le  t, \theta^* \in \mathcal C_s\}}\Big] +\sum_{t=2}^{T}\mathbb{E}\Big[||\theta^*-\tilde{\theta}_t||^2\cdot 1_{\{ \exists 2 \le s \le t, \theta^* \notin \mathcal C_s\}}\Big]  \\
	&\le \sum_{t=2}^{T}\mathbb{E}\Big[||\theta^*-\tilde{\theta}_t||^2\cdot 1_{\{ U_{t,4}\}}\Big] +\sum_{t=2}^{T}\big((\alpha_{\max}-\alpha_{\min})^2+(\beta_{\max}-\beta_{\min})^2\big)\frac{1}{T^2}  \\
	&\le  C_3\sum_{t=2}^{T}\frac{w_{t-1}^2}{(n \wedge (t-1))\delta^2+n\sigma^2} + ((\alpha_{\max}-\alpha_{\min})^2+ (\beta_{\max}-\beta_{\min})^2 )\frac{1}{T}, 
	\end{align*}
	where the first inequality follows from the proof of Lemma \ref{lemma-proof-multiple historical prices-step2} and the concentration ineqality in Lemma \ref{UCB-concentration} with $\epsilon=\frac{1}{T^2}\wedge \frac{1}{n\sigma^2} \le \frac{1}{T^2}$.
	 When $n \ge T$,  we have 
	\begin{align*}
	\sum_{t=2}^{T} \frac{w_{t-1}^2}{(n \wedge (t-1))\delta^2+n\sigma^2}  
	& \le w_T^2 \sum_{t=1}^{T-1}\frac{1}{t\delta^2+n\sigma^2}\\
	& = \mathcal{O}\big(\frac{\log T\cdot \log (T\delta^2+n\sigma^2)}{\delta^2}\big) \\
	&=\mathcal{O}\Big(\frac{T(\log T)^2}{(n \wedge T) \delta^2+n\sigma^2}\Big),
	\end{align*} 
	where the   second identity follows from  $n\sigma^2 \lesssim T\delta^2$. 
	When $n  < T$,  we have 
	\begin{align*}
	\sum_{t=2}^{T} \frac{w_{t-1}^2}{(n \wedge (t-1))\delta^2+n\sigma^2} &=\sum_{t=1}^{n}\frac{w_t^2}{t\delta^2+n\sigma^2}+\sum_{t=n+1}^{T-1}\frac{w_t^2}{n\delta^2+n\sigma^2} \\
	&= \mathcal{O}\Big(\frac{\log T\cdot \log (n\delta^2+n\sigma^2)}{\delta^2}\Big) +\mathcal{O}\Big(\frac{T\log T}{n\delta^2+n\sigma^2}\Big) \\
	&=\mathcal{O}\Big(\frac{T(\log T)^2}{(n\wedge T)\delta^2+n\sigma^2}\Big). 
	\end{align*}

		\underline{\textbf{Case 2.} $n \sigma^2 \gtrsim (n\wedge T)\delta^2$.}
	 In this case,  to prove the upper bound $\mathcal{O}\big(\frac{T(\log T)^2}{(n\wedge T)\delta^2+n\sigma^2}\big)$,  we first establish the following lemma, whose proof is deferred to  Appendix B.3. 
\begin{lemma}\label{appendix-lemma-thm4}
		Suppose   $\theta^* \in \mathcal C_t$ for each $t \in [T-1]$, then  for each $2 \le t\le T$, 
	\begin{align*}  
  ||\theta^* -\tilde \theta_{t}||^2 \le   2\big((4u+1)^2+1\big) \frac{w_{t-1}^2}{n\sigma^2}. 
	\end{align*} 
	\end{lemma}
	
	Based on the above Lemma \ref{appendix-lemma-thm4}, we have 
		\begin{align*}
	\sum_{t=2}^{T}\mathbb{E}[||\theta^*-\tilde{\theta}_t||^2] &=\sum_{t=2}^{T}\mathbb{E}\Big[||\theta^*-\tilde{\theta}_t||^2\cdot 1_{\{\forall s \in [t-1], \theta^* \in \mathcal C_s\}}\Big] +\sum_{t=2}^{T}\mathbb{E}\Big[||\theta^*-\tilde{\theta}_t||^2\cdot 1_{\{ \exists s \in [t-1], \theta^* \notin \mathcal C_s\}}\Big]  \\
	&\le   2((4u+1)^2+1)  \sum_{t=2}^{T}   \frac{w_{t-1}^2}{n\sigma^2} +\big((\alpha_{\max}-\alpha_{\min})^2+(\beta_{\max}-\beta_{\min})^2\big)\sum_{t=2}^{T} \frac{1}{T^2}\wedge \frac{1}{n\sigma^2} \\ 
 	 &= \mathcal{O}\Big(\frac{T\log T}{n\sigma^2}\Big) \\
 	 &=\mathcal{O}\Big(\frac{T \log T}{(n\wedge T)\delta^2+n \sigma^2}\Big), 
	\end{align*}
	where the inequality follows from Lemma  \ref{UCB-concentration} with $\epsilon =\frac{1}{T^2}\wedge \frac{1}{n\sigma^2}$ and Lemma  \ref{appendix-lemma-thm4},  and in the last identity, we utilize $n\sigma^2 \gtrsim (n \wedge T)\delta^2$.

		 \underline{\textbf{Step 3.}} In this step, we use the results in Step 1 and Step 2 to show that   M-O3FU algorithm achieves the regret upper bound $\mathcal{O}\big(T \delta^2+1\big)$ in the corner case, i.e., when $\delta^2 \lesssim\frac{1}{n\sigma^2}\lesssim \frac{1}{\sqrt T}$ holds, and $\mathcal{O}\big((\sqrt T \log T)\wedge \frac{T(\log T)^2}{(n \wedge T)\delta^2+n\sigma^2}+1\big)$ in the regular case, i.e., when $\delta^2 \lesssim\frac{1}{n\sigma^2}\lesssim \frac{1}{\sqrt T}$ does not hold.

		 Recall that $\hat \theta_0$ is the least-square estimator from   offline regression, and it follows from  Lemma \ref{UCB-concentration}  that    with probability $1-\epsilon$,  $||\theta^*-\hat \theta_0||^2_{V_{0,n}} \le w_0^2$ holds, where $w_0=R\sqrt{2\log\frac{n+1}{\epsilon}}+\sqrt{(1+u^2)(\alpha_{\max}^2+\beta_{\min}^2)}$.  Since $\lambda_{\min}(V_{0, n}) \ge \frac{2}{(1+2u-l)^2}n\sigma^2$ from Lemma 2 in \cite{KeskinZeevi2014}, 
		 it can be   verified that when $\theta^* \in \mathcal{C}_0$, there exists some constant $L_0>0$, such that the length of interval $\{\psi(\theta): \theta \in \mathcal{C}_0\}$ is   $\frac{L_0}{2\sqrt{n\sigma^2}}$. In other words,   $\mathbb{P}(\max_{\theta_1, \theta_2 \in \mathcal{C}_0}|\psi(\theta_1)-\psi(\theta_2)| \le \frac{L_0}{2\sqrt{n\sigma^2}}) \ge \mathbb{P}(\theta^* \in \mathcal{C}_0) \ge 1-\epsilon$.  
		   Let $\mathcal{P}_0=\{\psi(\theta):  \theta \in \mathcal{C}_0\}$, and   $A$ be the event $\big\{\min_{\theta \in \mathcal{C}_0} |\psi(\theta)-\bar p_{1:n}|  \le \frac{KL_0}{2\sqrt{n\sigma^2}}\big\}$ for some pre-determined constant  $K>1$.

		\underline{\textbf{Corner case: $\delta^2 \le \frac{K^2L_0^2}{4n\sigma^2}$ and $ n\sigma^2 \ge \sqrt T$.}}  In this case, if $\theta^* \in \mathcal{C}_0$, we have 
		\begin{align*}
		\min_{\theta \in \mathcal{C}_0} |\psi(\theta)-\bar p_{1:n}| \le  |\psi(\theta^*)-\bar p_{1:n}|  \le \frac{KL_0}{2\sqrt{n\sigma^2}}, 
		\end{align*} 
		and therefore,  $\mathbb{P}(A) \ge \mathbb{P}(\theta^* 
		\in \mathcal{C}_0) \ge  1-\epsilon$, and  when $A$ holds,   M-O3FU  algorithm will use the price $\bar p_{1:n}$ for any $1\le t \le T$ due to $n\sigma^2\ge   \sqrt T$. Thus, 
		\begin{align*} 
		R^{\pi}_{\theta^*}(T) 
		&= \mathbb{P}(A)\cdot    \sum_{t=1}^{T} \mathbb{E}\Big[r^*(\theta^*)-r(p_t; \theta^*)\Big|A\Big]+ \mathbb{P}(A^{\complement})\cdot    \sum_{t=1}^{T} \mathbb{E}\Big[r^*(\theta^*)-r(p_t; \theta^*)\Big|A^{\complement}\Big] \\
		& \lesssim  T\delta^2+\epsilon      \sqrt T \log T   \\
		&\lesssim  T\delta^2+1, 
		\end{align*}
		where the first inequality holds since when $A$ does not hold, M-O3FU algorithm directly applies O3FU algorithm, and incurs the regret $\mathcal{O}(\sqrt T \log T)$ from the result in Step 1, the second inequality holds since $\epsilon \sqrt  T \log T = (\frac{1}{T^2}\wedge \frac{1}{n\sigma^2})\sqrt T \log T \lesssim 1$.

		\underline{\textbf{Regular case 1:   $\delta^2 \le \frac{K^2L_0^2}{4n\sigma^2}$ and $n\sigma^2 < \sqrt T$.}} In this case,
	 since $n\sigma^2 < \sqrt T$, M-O3FU algorithm  runs O3FU algorithm from the beginning, and the regret is bounded by $\mathcal{O}\big( (\sqrt T\log T)\wedge (\frac{T(\log T)^2}{(n\wedge T)\delta^2+n\sigma^2}+1)\big)$ from the results in Steps 1 and 2.

	 	\underline{\textbf{Regular case 2:    
	 		$\frac{K^2L_0^2}{4n \sigma^2} \le  \delta^2 \le  \frac{K^2L_0^2}{n \sigma^2}$.}} In this case,  the condition  $\min_{\theta \in \mathcal{C}_0} |\psi(\theta)-\bar p_{1:n}| \le   \frac{KL_0}{2\sqrt{n\sigma^2}}$ can either hold or not.
	 If  the condition holds and $n\sigma^2\ge \sqrt T$, the regret is $\mathcal{O}(T\delta^2)$. Since in this case,  $T \delta^2 \lesssim \frac{T}{n\sigma^2}\lesssim \sqrt T$ and $n\sigma^2 \gtrsim  \sqrt T \gtrsim T \delta^2 \gtrsim (n\wedge T)\delta^2$, we have  $\mathcal{O}(T\delta^2) = \mathcal{O}((\sqrt T\log T) \wedge \frac{T(\log T)^2}{(n \wedge T)\delta^2+n\sigma^2}+1)$. 
	 If the condition does not hold, the regret is still bounded by  $\mathcal{O}((\sqrt T\log T) \wedge \frac{T(\log T)^2}{(n \wedge T)\delta^2+n\sigma^2}+1)$.

			\underline{\textbf{Regular case 3:    
				 $\delta^2 > \frac{K^2L_0^2}{n \sigma^2}$.}} In this case,  when $\theta \in \mathcal{C}_0$, we have 
			 \begin{align*}
			  \min_{\theta  \in \mathcal{C}_0} |\psi(\theta)-\bar p_{1:n}| \ge |\bar p_{1:n}-\psi(\theta^*)| - |\text{Proj}_{\mathcal{P}_0}(\bar p_{1:n}) -\psi(\theta^*)| \ge \frac{KL_0}{\sqrt{n\sigma^2}}-\frac{L_0}{2\sqrt{n\sigma^2}} > \frac{KL_0}{2\sqrt{n\sigma^2}}, 
			 \end{align*}
			 where  the first inequality follows from the  triangle inequality ($\text{Proj}_{\mathcal{P}_0}(\bar p_{1:n})$ denotes the projection of $\bar p_{1:n}$ to set $\mathcal{P}_0$), the second inequality holds since the length of $\mathcal{P}_0$ is $\frac{L_0}{2\sqrt{n\sigma^2}}$ and $\theta^* \in \mathcal{C}_0$, and the last inequality follows from $K>1$.  In this case, $\theta^* \in \mathcal{C}_0$ implies $A^{\complement}$.  Therefore, with probability $1-\epsilon$, $\mathbb{P}(A^{\complement}) \ge 1-\epsilon$. Thus,  if $n\sigma^2\ge  \sqrt T$,  the regret is upper bounded as follows: 
			 \begin{align*} 
			 R^{\pi}_{\theta^*}(T) 
			 &= \mathbb{P}(A)\cdot    \sum_{t=1}^{T} \mathbb{E}\Big[r^*(\theta^*)-r(p_t; \theta^*)\Big|A\Big]+ \mathbb{P}(A^{\complement})\cdot    \sum_{t=1}^{T} \mathbb{E}\Big[r^*(\theta^*)-r(p_t; \theta^*)\Big|A^{\complement}\Big] \\
			 & \lesssim \frac{1}{T^2} \cdot  T\delta^2 +( \sqrt T \log T) \wedge  \frac{T(\log T)^2}{(n \wedge T)\delta^2+n\sigma^2} \\
			 &\lesssim 1+( \sqrt T \log T) \wedge  \frac{T(\log T)^2}{(n \wedge T)\delta^2+n\sigma^2},
			 \end{align*}
			 where the first inequality holds since $\epsilon= \frac{1}{T^2}\wedge \frac{1}{n\sigma^2} \le \frac{1}{T^2}$.  
			 If $n\sigma^2 < \sqrt T$,  M-O3FU algorithm runs O3FU algorithm from the beginning, and the regret is bounded by $\mathcal{O}( (\sqrt T\log T)\wedge  \frac{T(\log T)^2}{(n \wedge T)\delta^2+n\sigma^2}+1)$.\qed

{ \subsection*{B.2. Proof of Lemma \ref{lemma-proof-multiple historical prices-step2}}} 
  When $t=1$, since $p_1=l\cdot\mathbb{I}\{\bar{p}_{1:n}>\frac{u+l}{2}\}+u\cdot\mathbb{I}\{\bar{p}_{1:n}\le\frac{u+l}{2}\}$, then $|p_1-\bar p_{1:n}| \ge \frac{u-l}{2}\ge \frac{1}{2}\delta$. Thus, when $t=1$,  $U_{t,3}$ holds.   
 	 
 	We next prove the following result: under the conditions of Lemma \ref{lemma-proof-multiple historical prices-step2}, suppose for each $1\le s \le t-1$ (for a fixed $2 \le t \le T$), the event $U_{s,3}$ holds, then $U_{t,3}$ and $U_{t,4}$ also hold. Let   $\Delta \alpha_t=\tilde{\alpha}_t-\alpha^*$, $\Delta \beta_t=\tilde{\beta}_t-\beta^*$, and $\gamma_t=\frac{\Delta \alpha_t}{\Delta \beta_t}$ (when $\Delta \beta_t \neq 0$). 	
Note that  the following generalized version of   the inequality  \eqref{ineq1-proof-lemma8} holds: 
 	\begin{align} \label{ineq1-proof-lemma2}
 	\lambda\big((\Delta \alpha_t)^2+(\Delta\beta_t)^2\big)+\sum_{i=1}^{n}\big( \Delta \alpha_t+\Delta\beta_t\hat p_i \big)^2+\sum_{s=1}^{t-1}\big(\Delta \alpha_t+\Delta\beta_t p_s\big)^2 \le 2w_{t-1}^2.
 	\end{align} 
 Similar to the proof of Lemma \ref{lemma-proof-UCB-step2}, we  also divide the proof into three cases. 
 
 \underline{\textbf{Case 1}: $\Delta \beta_t=0$.} 
 In this case, \eqref{ineq1-proof-lemma2} becomes  $
 (\Delta \alpha_t)^2(\lambda+n+t-1) \le 2w_{t-1}^2,$ and
 \begin{align}\label{ineq2-proof-lemma2}
 ||\theta^*-\tilde{\theta}_t||^2 =(\Delta \alpha_t)^2+(\Delta \beta_t)^2= (\Delta \alpha_t)^2 \le \frac{2w_{t-1}^2}{n+t-1}. 
 \end{align}
 Therefore, combining $\sigma \le u-l$, $\delta \le u-l$, and \eqref{ineq2-proof-lemma2},  we obtain  
 \begin{align*}
 ||\theta^*-\tilde{\theta}_t||^2  \le \frac{4(u-l)^2w_{t-1}^2}{(n\wedge (t-1))\delta^2+n \sigma^2}.
 \end{align*} 
In  addition, \eqref{ineq2-proof-lemma2} also implies 
 \begin{align*}
 |\bar p_{1:n}-p_t| \ge |\bar p_{1:n}-\psi(\theta^*)|- |p_t-\psi(\theta^*)| \ge  |\bar p_{1:n}-\psi(\theta^*)|-\frac{\sqrt{\alpha_{\max}^2+\beta_{\max}^2}}{2\beta_{\max}^2}\cdot \frac{\sqrt 2 w_{t-1}}{\sqrt{ n+t-1}}\ge \frac{1}{2}\delta,
 \end{align*}
 where the second inequality follows from  \eqref{ineq2-proof-lemma2}  and Lipschitz continuity of the function $\psi(\cdot)$,   and the last inequality holds since from the assumption  of $\delta \ge \frac{\sqrt{2(\alpha_{\max}^2+\beta_{\max}^2)}}{\beta_{\max}^2} \cdot \frac{T^{1/4}w_T}{n^{1/2}}$, we have 
 \begin{align}\label{ineq4-proof-lemma2}
 \frac{w_{t-1}}{\sqrt{n+t-1}} \le \frac{w_T}{\sqrt{n}}   \le  \frac{\beta_{\max}^2}{\sqrt{2(\alpha_{\max}^2+\beta_{\max}^2)}}\delta.
 \end{align}

 \underline{\textbf{Case 2}: $\Delta \beta_t\neq 0$, $|\gamma_t| \ge 4u+1$.} 
 In this case, we have 
 \begin{align}\label{ineq3-proof-lemma2}
 ||\theta^*-\tilde{\theta}_t||^2 
 \le \frac{2w_{t-1}^2(1+\gamma_t^2)}{\lambda(1+\gamma_t^2)+\sum_{i=1}^{n}(\gamma_t+\hat p_i)^2+\sum_{s=1}^{t-1}(\gamma_t+p_s)^2}  \le \frac{2w_{t-1}^2(1+\gamma_t^2)}{ n(\gamma_t+\bar p_{1:n})^2}   \le \frac{4w_{t-1}^2}{n},
 \end{align}
 where  the second inequality holds since $\sum_{i=1}^{n}(\gamma_t+\hat p_i)^2 \ge n(\gamma_t+\bar p_{1:n})^2$, and  
 the last inequality follows from  $ 1+\gamma_t^2\le 2(\gamma_t+\bar p_{1:n})^2$, which is easily verified by noting $
 (\gamma_t+2\bar p_{1:n})^2\ge( |\gamma_t|-2\bar{p}_{1:n})^2 \ge (2\bar{p}_{1:n}+1)^2 \ge 2\bar{p}_{1:n}^2+1$. 
 Then,  \eqref{ineq3-proof-lemma2} implies 
 \begin{align*}
 ||\theta^*-\tilde{\theta}_t||^2 \le \frac{8(u-l)^2w_{t-1}^2}{(n\wedge (t-1))\delta^2+n\sigma^2},
 \end{align*}
 and in addition, 
 \begin{align*}
 |\bar p_{1:n}-p_t| \ge |\bar p_{1:n}-\psi(\theta^*)|- |p_t-\psi(\theta^*)| \ge  |\bar p_{1:n}-\psi(\theta^*)|-\frac{\sqrt{\alpha_{\max}^2+\beta_{\max}^2}}{2\beta_{\max}^2} \frac{2w_{t-1}}{\sqrt{n}}\ge (1-\frac{\sqrt 2}{2})\delta,
 \end{align*}
 where the  last inequality follows from \eqref{ineq4-proof-lemma2}.

 \underline{\textbf{Case 3}: $\Delta \beta_t\neq 0$, $|\gamma_t| < 4u+1$.} 
Recall the definitions of $C_0$ and $C_1$: 
 \begin{align*}
 C_0=\frac{l|\beta_{\max}|}{u |\beta_{\min}|}, \quad C_1 = \frac{4(C_0+1)^2}{C_0^2}(1+(4u+1)^2).
 \end{align*} 
 	
 	\underline{\textbf{Subcase 3.1}: $1+\gamma_t^2 \le C_1 \frac{(\gamma_t+\bar p_{1:n})^2}{\delta^2}$.} In this subcase,  we have 
 	\begin{align*}%
 	||\theta^*-\tilde{\theta}_t||^2 
 	\le \frac{2 w_{t-1}^2(1+\gamma_t^2)}{ n(\gamma_t+\bar p_{1:n})^2} \le \frac{2C_1 w_{t-1}^2}{n\delta^2}.
 	\end{align*}
 	 From the assumption of $\sigma \le  \delta$, we have 
 	 \begin{align*}
 	 	||\theta^*-\tilde{\theta}_t||^2 \le \frac{4C_1 w_{t-1}^2}{n \delta^2+   n \sigma^2} \le \frac{4C_1  w_{t-1}^2}{(n\wedge (t-1))\delta^2+n\sigma^2}, 
 	 \end{align*}
 	 and  
in addition,   
\begin{align*}
|p_t-\bar p_{1:n}| 
&\ge |\psi(\theta^*)-\bar p_{1:n}|-|p_t-\psi(\theta^*)| \\
&\ge |\psi(\theta^*)-\bar p_{1:n}|-  \frac{\sqrt{\alpha_{\max}^2+\beta_{\max}^2}}{2\beta_{\max}^2}\frac{\sqrt{2C_1}w_{t-1}}{\sqrt{n}|\psi(\theta^*)-\bar p_{1:n}|} \\
& \ge  |\psi(\theta^*)-\bar p_{1:n}|- \frac{\sqrt C_1}{2T^{1/4}}\\ 
&\ge \frac{1}{2}\delta,
\end{align*}
{where the third inequality follows from \eqref{ineq4-proof-lemma2}}, and   in the last inequality, we utilize the assumption of $\delta \ge \sqrt{C_1}T^{-1/4}$.

 	\underline{\textbf{Subcase 3.2}: $1+\gamma_t^2 > C_1 \frac{(\gamma_t+\bar p_{1:n})^2}{\delta^2}$.}
 	In this subcase,  we have  
 	\begin{align*}
 	||\theta^*-\tilde{\theta}_t||^2  
 	 &\le  \frac{2w_{t-1}^2  ( \gamma_t^2+1)}{	\lambda(\gamma_t^2+1)+\sum_{i=1}^{n}( \gamma_t+ \hat p_i )^2+\sum_{s=1}^{t-1}(\gamma_t+ p_s)^2} \le  \frac{2w_{t-1}^2  ( (4u^2+1)^2+1)}{	\sum_{i=1}^{n}( \gamma_t+ \hat p_i )^2+\sum_{s=1}^{t-1}(\gamma_t+ p_s)^2}.
 	\end{align*}
 	To proceed, we establish the following inequality:
 	\begin{align}\label{ineq-proof-leamm2-multiple}
 	\sum_{i=1}^{n}( \gamma_t+ \hat p_i )^2+\sum_{s=1}^{t-1}(\gamma_t+ p_s)^2 \ge n\sigma^2+ (n \wedge (t-1))\min\Big\{(1-\frac{\sqrt 2}{2})^2, \frac{C_0^2}{4}\Big\}\cdot (\psi(\theta^*)-\bar p_{1:n})^2.
 	\end{align} 
 	Note that
 	$
 	\sum_{i=1}^{n}( \gamma_t+ \hat p_i )^2+\sum_{s=1}^{t-1}(\gamma_t+ p_s)^2
 	$ is convex in $\gamma_t$ and is minimized at $\gamma_t=-\frac{\sum_{i=1}^n\hat p_i+\sum_{s=1}^{t-1} p_s}{n+t-1}$. We have
 	\begin{align*}
 	\sum_{i=1}^{n}( \gamma_t+ \hat p_i )^2+\sum_{s=1}^{t-1}(\gamma_t+ p_s)^2&\ge \sum_{i=1}^{n}( \hat p_i -\frac{\sum_{i=1}^n\hat p_i+\sum_{s=1}^{t-1} p_s}{n+t-1})^2+\sum_{s=1}^{t-1}(p_s-\frac{\sum_{i=1}^n\hat p_i+\sum_{s=1}^{t-1} p_s}{n+t-1})^2\\
 	&=\text{Var}((\hat{p}_1,\dots,\hat{p}_n),(p_1,\dots,p_{t-1})),
 	\end{align*}
 	where $((\hat{p}_1,\dots,\hat{p}_n),(p_1,\dots,p_{t-1}))\in\mathbb{R}^{(n+t-1)\times1}$.
 	Define $$f(p_1,\dots,p_{t-1}):=\text{Var}((\hat{p}_1,\dots,\hat{p}_n),(p_1,\dots,p_{t-1})).$$ Then
 	\begin{align*}
 	f(p_1,\dots,p_{t-1})&=\Vert((\hat{p}_1,\dots,\hat{p}_n),(p_1,\dots,p_{t-1}))\Vert_2^2-\frac{\left[1_{(n+t-1)\times1}^\top ((\hat{p}_1,\dots,\hat{p}_n),(p_1,\dots,p_{t-1}))\right]^2}{(n+t-1)}\\
 	&=\Vert(\hat{p}_1,\dots,\hat{p}_n)\Vert_2^2+\Vert(p_1,\dots,p_{t-1})\Vert_2^2-\frac{\left[1_{n\times1}^\top(\hat{p}_1,\dots,\hat{p}_n)+1_{(t-1)\times1}^\top(p_1,\dots,p_{t-1})\right]^2}{n+t-1},
 	\end{align*}
 	thus
 	\begin{align*}
 	\frac{\partial f(p_1,\dots,p_{t-1})}{\partial(p_1,\dots,p_{t-1})}&=2(p_1,\dots,p_{t-1})-2\frac{\left[1_{n}^\top(\hat{p}_1,\dots,\hat{p}_n)+1_{t-1}^\top(p_1,\dots,p_{t-1})\right]}{n+t-1}1_{(t-1)\times1},
 	\end{align*}
 	\begin{align*}
 	\frac{\partial^2 f(p_1,\dots,p_{t-1})}{\partial(p_1,\dots,p_{t-1})^2}&=2\left(I_{(t-1)}-\frac{1_{(t-1)\times1}1_{(t-1)\times1}^\top}{n+t-1}\right)\succeq0.
 	\end{align*}
 	Therefore, we know that $f(p_1,\dots,p_{t-1})$ is convex in $(p_1,\dots,p_{t-1})$ and is minimized at $(p_1,\dots,p_{t-1})=\bar{p}_{1:n}1_{(t-1)\times1}$. By the Taylor series of $f(p_1,\dots,p_{t-1})$ at point $\bar{p}_{1:n}1_{(t-1)\times1}$, we have
 	\begin{align*}
 	&f(p_1,\dots,p_{t-1})-f\left(\bar{p}_{1:n}1_{(t-1)\times1}\right)\\
 	=&\left((p_1,\dots,p_{t-1})-\bar{p}_{1:n}1_{(t-1)\times1}\right)^\top\left[I_{(t-1)}-\frac{1_{(t-1)\times1}1_{(t-1)\times1}^\top}{n+t-1}\right]\left((p_1,\dots,p_{t-1})-\bar{p}_{1:n}1_{(t-1)\times1}\right)\\
 	=&\Vert (p_1,\dots,p_{t-1})-\bar{p}_{1:n}1_{(t-1)\times1}\Vert_2^2-\frac{\left(\sum_{s=1}^{t-1}(p_s-\bar{p}_{1:n})\right)^2}{n+t-1}\\
 	=&\sum_{s=1}^{t-1}(p_s-\bar{p}_{1:n})^2-\frac{\left(\sum_{s=1}^{t-1}(p_s-\bar{p}_{1:n})\right)^2}{n+t-1}\\
 	\ge&\frac{n}{n+t-1}\sum_{s=1}^{t-1}(p_s-\bar{p}_{1:n})^2\\
 	\ge&\frac{n(t-1)}{(n+t-1)}\cdot \min\Big\{(1-\frac{\sqrt 2}{2})^2, \frac{C_0^2}{4}\Big\}\cdot \delta^2,
 	\end{align*}
 	where the last inequality is by the induction assumption that  $U_{s,2}=\Big\{|p_s -\bar p_{1:n}| \ge  \min\big\{1-\frac{\sqrt 2}{2}, \frac{C_0}{2}\big\}\cdot \delta \Big\}$ holds for $s=1,\dots,t-1$. Using also the fact that $f\left(\bar{p}_{1:n}1_{(t-1)\times1}\right)=\text{Var}(\hat{p}_1,\dots,\hat{p}_n)=n\sigma^2$, we have
 	$$
 	\sum_{i=1}^{n}( \gamma_t+ \hat p_i )^2+\sum_{s=1}^{t-1}(\gamma_t+ p_s)^2\ge f(p_1,\dots,p_{t-1})\ge n\sigma^2+ (n \wedge (t-1))\min\Big\{(1-\frac{\sqrt 2}{2})^2, \frac{C_0^2}{4}\Big\}\cdot \delta^2.
 	$$
 Therefore, we have proven (\ref{ineq-proof-leamm2-multiple}) and can conclude that
 \begin{align*}
 ||\tilde \theta_t-\theta^*||^2 \le  2\max\big\{2(\sqrt 2+1)^2, \frac{4}{C_0^2}\big\}\cdot ((4u+1)^2+1)\cdot  \frac{w_{t-1}^2}{(n \wedge (t-1))\delta^2+n\sigma^2}. 
 \end{align*}

 	Now, it suffices to bound the term $|p_t-\bar{p}_{1:n}|$. 
 	We still have (\ref{ineq5-proof-lemma8}) (which we proved in the single-historical-price setting), i.e., the following inequality:
 	\begin{align} \label{ineq2-proof-leamm2-multiple}
 	|\gamma_t+p_t|\ge C_0|\gamma_t+\psi(\theta^*)|,
 	\end{align}
 	thus 
 	\begin{align*}
 	|p_t-\bar{p}_{1:n}| 
 	&\ge |p_t+ \gamma_t|-|\gamma_t+\bar{p}_{1:n}| \\
 	&\ge C_0|\gamma_t+\psi(\theta^*)|-|\gamma_t+\bar{p}_{1:n}| \\
 	&\ge C_0(|\psi(\theta^*)-\bar{p}_{1:n}|-|\gamma_t+\bar{p}_{1:n}|)-|\gamma_t+\bar{p}_{1:n}| \\
 	&=C_0|\psi(\theta^*)-\bar{p}_{1:n}|-(C_0+1)|\gamma_t+\bar{p}_{1:n}| \\
 	&\ge \Big(C_0-(C_0+1)\frac{\sqrt{1+(4u+1)^2}}{\sqrt{C_1}}\Big)|\psi(\theta^*)-\bar{p}_{1:n}|\\
 	&\ge \frac{C_0}{2} \delta,
 	\end{align*}
 	where the second inequality follows from \eqref{ineq2-proof-leamm2-multiple}, the fourth inequality follows from the assumption of Subcase 3.2, i.e., $1+\gamma_t^2 > C_1 \frac{(\gamma_t+\bar{p}_{1:n})^2}{\delta^2}$ and $|\gamma_t| \le 4u+1$, and the last inequality follows from the definition of $C_1$.  
 	
 	Therefore, combining the above three cases, we conclude that 
 	\begin{align*}
 	&|\bar p_{1:n}-\psi(\theta^*)| \ge \min\Big\{1-\frac{\sqrt 2}{2}, \frac{C_0}{2}\Big\}\cdot \delta,\\
 	& ||\theta^* -\tilde \theta_{t}||^2 \le \max\Big\{8(u-l)^2, 4C_1, 2\max\big\{2(\sqrt 2+1)^2, \frac{4}{C_0^2}\big\}\cdot ((4u+1)^2+1) \Big\} \cdot  \frac{w_{t-1}^2}{(n\wedge (t-1))\delta^2+n\sigma^2},
 	\end{align*}
 	 i.e.,  $U_{t,3}$ and $U_{t,4}$ hold, which completes the  inductive arguments.   \qed

 	\subsection*{B.3. Proof of Lemma \ref{appendix-lemma-thm4}}  
 Since  $\theta^* \in \mathcal{C}_t$ for each $t \in [T]$, and  $\tilde{\theta}_t \in \mathcal{C}_t$ for each $2 \le t \le T$, the inequality  \eqref{ineq1-proof-lemma2} still holds. 
For each $2 \le t \le T$, we bound $||\theta^*-\tilde \theta_t||^2$ by considering the following  three cases.

 \underline{\textbf{Case 1}: $\Delta \beta_t=0$.} 
 In this case, \eqref{ineq1-proof-lemma2} becomes  $
 (\Delta \alpha_t)^2(\lambda+n+t-1) \le 2w_{t-1}^2,$ and
 \begin{align*} 
 ||\theta^*-\tilde{\theta}_t||^2  = (\Delta \alpha_t)^2 \le \frac{2w_{t-1}^2}{n } \le \frac{2(u-l)^2w_{t-1}^2}{n\sigma^2},
 \end{align*}
  where the second inequality holds since $\sigma \le u-l$.

 \underline{\textbf{Case 2}: $\Delta \beta_t\neq 0$, $|\gamma_t| \ge 4u+1$.} 
 In this case, we have 
 \begin{align*} 
 ||\theta^*-\tilde{\theta}_t||^2 
 \le \frac{2w_{t-1}^2(1+\gamma_t^2)}{ \sum_{i=1}^{n}(\gamma_t+\hat p_i)^2 }  \le \frac{2w_{t-1}^2(1+\gamma_t^2)}{ n(\gamma_t+\bar p_{1:n})^2}   \le \frac{4w_{t-1}^2}{n} \le \frac{4(u-l)^2w_{t-1}^2}{n\sigma^2},
 \end{align*}
 where  the second inequality holds since $\sum_{i=1}^{n}(\gamma_t+\hat p_i)^2 \ge n(\gamma_t+\bar p_{1:n})^2$. and  
 the third inequality follows from  $ 1+\gamma_t^2\le 2(\gamma_t+\bar p_{1:n})^2$.

 \underline{\textbf{Case 3}: $\Delta \beta_t\neq 0$, $|\gamma_t| < 4u+1$.} In this case, 
  \begin{align*} 
 ||\theta^*-\tilde{\theta}_t||^2 \le \frac{2w_{t-1}^2(1+\gamma_t^2)}{ \sum_{i=1}^{n}(\gamma_t+\hat p_i)^2} \le \frac{2 ((4u+1)^2+1)w_{t-1}^2}{\sum_{i=1}^{n}(\hat p_i-\bar p_{1:n})^2} = \frac{2 ((4u+1)^2+1)w_{t-1}^2}{n\sigma^2},
 \end{align*}
 where the second inequality holds since $\sum_{i=1}^{n}(\hat p_i-\bar p_{1:n})^2 =\min_{x \in \mathbb{R}} (\hat p_i+x)^2 \le \sum_{i=1}^{n}(\hat p_i+\gamma_t)^2$.

 Therefore, combining the above three cases, we obtain $
  ||\theta^*-\tilde \theta_{t}||^2 \le    2((4u+1)^2+1)  \cdot \frac{w_{t-1}^2}{n\sigma^2}$,  which completes the proof. \qed

  \subsection*{B.4. Proof of Theorem \ref{thm-lower bound-multiple historical prices}}

  Similar to the proof of Theorem \ref{thm-lower bound}, we consider  normal random noise with standard deviation $R$, and for simplicity, we assume  $\xi=\frac{1}{2}$.   The proof is divided into two major steps.

  \underline{\textbf{Step  1.}} In the first step, we prove the following result: for any pricing policy $\pi$, 
  \begin{align}\label{ineq1-proof-thm3}
  \sup_{ \theta \in \Theta_0(\delta)} R^{\pi}_{\theta}(T, n, \sigma, \delta)=\Omega\Big(\sqrt T 
  \wedge \frac{T}{\delta^{-2}+n\sigma^2+(n \wedge T)\delta^2}\Big),
  \end{align}
  where   $\Theta_0(  \delta) = \big\{\theta \in \Theta^{\dag}: \, \psi(\theta)-\bar p_{1:n} \in [\frac{\delta}{2}, \delta]\big\}$. 
  When (i) $\delta > \frac{lR}{32|\beta_{\min}|}\sqrt{\frac{|\beta_{\max}|}{K_0e}}T^{-\frac{1}{4}}(\log T)^{-\frac{1}{2}\lambda_0}$; or (ii) $\delta \le \frac{lR}{32|\beta_{\min}|}\sqrt{\frac{|\beta_{\max}|}{K_0e}}T^{-\frac{1}{4}}(\log T)^{-\frac{1}{2}\lambda_0}$ and $n\sigma^2 >  \frac{l^2R^2|\beta_{\max}|}{512\beta_{\min}^2K_0e}\frac{\sqrt T}{(\log T)^{\lambda_0}}$,  \eqref{ineq1-proof-thm3} provides the desired lower bound in Theorem \ref{thm-lower bound-multiple historical prices}.  
  
  To prove \eqref{ineq1-proof-thm3},   it suffices to show that $\sup_{\theta  \in \Theta_0(\delta)}R^{\pi}_{\theta}(T,n, \sigma,\delta)$ is lower bounded by  $\Omega(\sqrt T \wedge \frac{T}{\delta^{-2}+n\sigma^2+n\delta^2})$ and $\Omega(\sqrt T \wedge \frac{T}{\delta^{-2}+n\sigma^2+T\delta^2})$.  The proofs of these two bounds  are similar to \eqref{ineq1-proof-thm1} in the proof of Theorem \ref{thm-lower bound}, and we only highlight the difference here.
  For the first lower bound $\Omega(\sqrt T \wedge \frac{T}{\delta^{-2}+n\sigma^2+n\delta^2})$, 
  by defining a similar   prior distribution  $ q$ as \eqref{def-prior density function} and letting $C(\theta)=(\psi(\theta),1)$, 	 we have 
  \begin{align*} %
  \sum_{t=1}^{T}\mathbb{E}_{q}[\mathbb{E}_{\theta}^{\pi}[(p_t-\psi(\theta))^2] ]& \ge  \sum_{t=2}^{T}\frac{R^2\alpha_{\min}^2/(4\beta_{\min}^2)}{R^2 {\mathcal{I}}( q)+ 
  	\sum_{i=1}^{n}\mathbb{E}_{q}[(\hat p_i-\psi(\theta))^2]+ \sum_{s=1}^{t-1}\mathbb{E}_{  q}[\mathbb{E}_{\theta}^{\pi}[(p_s-\psi(\theta))^2]]} \nonumber \\ 
  & \ge  \sum_{t=2}^{T}\frac{R^2\alpha_{\min}^2/(4\beta_{\min}^2)}{R^2 {\mathcal{I}}(q)+ 2n\sigma^2+2n\delta^2+
  	\sum_{s=1}^{t-1}\mathbb{E}_{  q}[\mathbb{E}_{\theta}^{\pi}[(p_s-\psi(\theta))^2]]}\nonumber \\
  &\ge \frac{(T-1)R^2\alpha_{\min}^2/(4\beta_{\min}^2)}{R^2{\mathcal{I}}( q)+ 2n\sigma^2+2n\delta^2+
  	\sum_{s=1}^{t-1}\mathbb{E}_{ q}[\mathbb{E}_{\theta}^{\pi}[(p_s-\psi(\theta))^2]]}, 
  \end{align*}
  where the second inequality follows from  $\sum_{i=1}^{n}(\hat p_i-p^*_{\theta})^2 \le 2\sum_{i=1}^{n}(\hat p_i-\bar p_{1:n})^2+2n(\bar p_{1:n}-p^*_{\theta})^2 \le 2n\sigma^2+2n\delta^2$.  
  Since $\mathcal{I}(q)=\Theta(\delta^{-2})$, the first lower bound $\Omega(\sqrt T \wedge \frac{T}{\delta^{-2}+n\sigma^2+n\delta^2})$ can be proved.   
  For   the second lower bound $\Omega(\sqrt T \wedge \frac{T}{\delta^{-2}+n\sigma^2+T\delta^2})$, letting $C(\theta)=(-\bar p_{1:n},1)$  and applying the multivariate van Trees inequality to the prior distribution $ q$ defined in \eqref{def-prior density function}, we have 
  \begin{align*}
  \mathbb{E}_{ q}\big[\mathbb{E}_{\theta}^{\pi}[(p_t-\psi(\theta))^2]\big] & \ge \frac{(\mathbb{E}_{q}[C(\theta)^{\mathsf{T}}\frac{\partial \psi}{\partial \theta}])^2}{{\mathcal{I}}( q)+\mathbb{E}_{ q}[C(\theta)^{\mathsf{T}} \mathcal{I}_{t-1}^{\pi}(\theta)C(\theta)]}\\ 
  & \ge \frac{R^2 (\alpha_{\min}+\beta_{\min} \bar p_{1:n})^2/(4\beta_{\min}^2)}{R^2{\mathcal{I}}( q)+\sum_{i=1}^{n}(\hat p_i-\bar p_{1:n})^2+\sum_{s=1}^{t-1}\mathbb{E}_{  q}[\mathbb{E}^{\pi}_\theta[(p_s-\bar p_{1:n})^2]]} \nonumber \\
  &\ge \frac{R^2 (\alpha_{\min}+\beta_{\min} \bar p_{1:n})^2/(4\beta_{\min}^2)}{R^2 {\mathcal{I}}(  q)+n\sigma^2+2(t-1)\delta^2+2\sum_{s=1}^{t-1}\mathbb{E}_{\tilde q}[\mathbb{E}^{\pi}_\theta[(p_s-p^*_{\theta})^2]]}, 
  \end{align*}
  where the second inequality follows from $(p_s-\bar p_{1:n})^2 \le 2(p_s-p^*_{\theta})^2+2(\bar p_{1:n}-p^*_{\theta})^2$.  
  Again noting that $\mathcal{I}(q)=\Theta(\delta^{-2})$, we conclude that the regret is lower bounded by $\Omega(\sqrt T\wedge\frac{T}{\delta^{-2}+n\sigma^2+T \delta^2})$.

  \underline{\textbf{Step 2.}} In this step, we complete the proof by
  showing that when  $\delta \le \frac{lR}{32|\beta_{\min}|}\sqrt{\frac{|\beta_{\max}|}{K_0e}}T^{-\frac{1}{4}}(\log T)^{-\frac{1}{2}\lambda_0}  \text{ and } n\sigma^2 \le  \frac{l^2R^2|\beta_{\max}|}{512\beta_{\min}^2K_0e}\frac{\sqrt T}{(\log T)^{\lambda_0}}$, for any admissible policy $\pi \in \Pi^{\circ}$,  there exists   $\theta \in \Theta^{\dag}$ satisfying $|\psi(\theta)-\bar p_{1:n}| \in [\frac{1}{2}\delta, \frac{3}{2}\delta]$ such that 
  \begin{align} \label{eq1-proof-thm3}
  R^{\pi}_{\theta}(T)=\Omega\left(\frac{\sqrt T}{(\log T)^{\lambda_0}}\right). 
  \end{align} 
  The proof of the above \eqref{eq1-proof-thm3}  is similar to  \eqref{eq1-proof-thm1} in the proof of Theorem \ref{thm-lower bound}.   For completeness, the details are illustrated as follows. 
  We first define two two vectors of demand parameters $\theta_1=(\alpha_1,\beta_1)$ and $\theta_2=(\alpha_2,\beta_2)$ as follows:
  \begin{align}\label{def-thm3-small delta}
  -\frac{\alpha_1}{2\beta_1} = \bar p_{1:n}+\delta, \quad 	-\frac{\alpha_2}{2\beta_2} = \bar p_{1:n} +\delta+\Delta, \quad  \alpha_1-\alpha_2+(\beta_1-\beta_2)\bar p_{1:n}=0, 
  \end{align}
  where  $\Delta>0$ is  to be determined. 
  We  consider $P_1^{\pi}$, $P_2^{\pi}$ as the two probability measures  induced by the common policy $\pi$ and  two demand parameters $\theta_1$ and $\theta_2$ respectively.   That is,  for each $i=1, 2$,  
  \begin{align*}
  P_i^{\pi}(\hat D_1, \ldots\, \hat D_n, D_1, \ldots, D_T) =\prod_{t=1}^n \Big(\frac{1}{R}\phi\big(\frac{\hat D_t-(\alpha_i+\beta_i\hat p_t)}{R}\big)\Big)\cdot \prod_{t=1}^T \Big(\frac{1}{R}\phi\big(\frac{ D_t-(\alpha_i+\beta_i p_t)}{R}\big)\Big).
  \end{align*} 
  It is easily verified that the KL divergence between $P_1^{\pi}$ and $P_2^{\pi}$ is 
  \begin{align*}
  KL(P_1^{\pi}, P_2^{\pi}) 
  &=\frac{1}{2R^2}\Big( \sum_{i=1}^{n}\big((\alpha_1-\alpha_2)+(\beta_1-\beta_2)\hat p_i\big)^2+\sum_{t=1}^{T} \mathbb{E}_{\theta_1}^{\pi}\big[\big((\alpha_1-\alpha_2)+(\beta_1-\beta_2) p_t\big)^2\big]\Big)
  \\ &=\frac{(\beta_1-\beta_2)^2}{2R^2}\Big( \sum_{i=1}^{n}\big(\hat p_i-\bar p_{1:n}\big)^2+\sum_{t=1}^{T} \mathbb{E}_{\theta_1}^{\pi}\big[\big( p_t-\bar p_{1:n}\big)^2\big]\Big) \\
  &=\frac{2\beta_2^2 \Delta^2 }{(\bar p_{1:n}+2\delta)^2R^2}\Big(n\sigma^2+\sum_{t=1}^{T} \mathbb{E}_{\theta_1}^{\pi}\big[\big( p_t-\bar p_{1:n}\big)^2\big]\Big)\\
  &\le \frac{2\beta_{\min}^2\Delta^2}{l^2R^2}\Big( n\sigma^2+2\sum_{t=1}^{T} \mathbb{E}_{\theta_1}^{\pi}\big[\big( p_t-\psi(\theta_1)\big)^2\big]+2T\delta^2\Big),
  \end{align*} 
  where the second identity follows from   \eqref{def-thm3-small delta} and the third identity holds since $(\beta_1-\beta_2)^2=\frac{4\beta_2^2\Delta^2}{(\bar p_{1:n}+2\delta)^2}$ due to \eqref{def-thm3-small delta}. Therefore, we have 
  \begin{align}\label{proof-thm3-samll delta-eq3}
 R_{\theta_1}^{\pi}(T) \ge |\beta_{\max}| \cdot \sum_{t=1}^{T}\mathbb{E}_{\theta_1}^{\pi}[(p_t-\psi(\theta_1))^2] 
  & \ge  |\beta_{\max}|\cdot\Big(\frac{l^2R^2}{4\beta_{\min}^2\Delta^2}KL(P_1^{\pi}, P_2^{\pi}) -\frac{n\sigma^2}{2}-T\delta^2 \Big).%
  \end{align} 
  On the other hand,  we have   
  \begin{align}\label{proof-thm3-samll delta-eq2} 
  \frac{1}{32}  e^{-KL(P_1^{\pi},P_2^{\pi})}\cdot T\Delta^2 
  &\le \sum_{t=1}^{T}\mathbb{E}_{\theta_1}^{\pi}[(p_t-\psi(\theta_1))^2] +  \sum_{t=1}^{T}\mathbb{E}_{\theta_2}^{\pi}[(p_t-\psi(\theta_1))^2]  \nonumber \\
  & \le \frac{1}{|\beta_{\max}|}2K_0\sqrt T (\log T)^{\lambda_0},
  \end{align}
  where the first inequality follows from  Theorem 2.2 in \cite{Tsybakov2009}, the second inequality follows from the assumption on the policy $\pi$. Therefore, \eqref{proof-thm3-samll delta-eq2} implies 
  \begin{align*}
  KL(P_1^{\pi}, P_2^{\pi}) \ge \log\Big(\frac{\sqrt T|\beta_{\max}| \Delta^2}{64K_0(\log T)^{\lambda_0}}\Big). 
  \end{align*}
  Thus, by letting $\Delta^2=\frac{64K_0e(\log T)^{\lambda_0}}{|\beta_{\max}|\sqrt T}$,  from \eqref{proof-thm3-samll delta-eq3}, the regret can be lower bounded by 
  \begin{align*}
 R_{\theta_1}^{\pi}(T)   %
  &\ge |\beta_{\max}| \cdot \Big(\frac{l^2R^2}{4\beta_{\min}^2\Delta^2}\log\Big(\frac{\sqrt T \Delta^2}{64K_0(\log T)^{\lambda_0}}\Big) - \frac{n\sigma^2}{2}-T\delta^2 \Big)\\  
&= |\beta_{\max}| \cdot \Big( \frac{l^2R^2|\beta_{\max}|}{256\beta_{\min}^2K_0e}\cdot \frac{\sqrt T}{(\log T)^{\lambda_0}}-\frac{n\sigma^2}{2}-T\delta^2\Big)\\ 
&\ge \frac{l^2R^2\beta_{\max}^2}{512 \beta_{\min}^2K_0e} \cdot  \frac{\sqrt T}{(\log T)^{\lambda_0}},
  \end{align*}
  where the   second inequality follows from the definition of $\Delta$, $\delta\le \frac{lR}{32|\beta_{\min}|}\sqrt{\frac{|\beta_{\max}|}{K_0e}}T^{-\frac{1}{4}}(\log T)^{-\frac{1}{2}\lambda_0}$ and $n \sigma^2 \le \frac{l^2R^2|\beta_{\max}|}{512\beta_{\min}^2K_0e}\frac{\sqrt T}{(\log T)^{\lambda_0}}$. Therefore,  $R_{\theta_1}^{\pi}(T) =\Omega(\frac{\sqrt T}{(\log T)^{\lambda_0}})$.

  Combining Step 1 and Step 2, we conclude that for any admissible policy $\pi \in \Pi^{\circ}$, there exists  $\theta \in \Theta^{\dag}$ satisfying $|\psi(\theta)-\bar p_{1:n}| \in [(1-\xi)\delta, (1+\xi)\delta]$, such that
  \begin{align*}
  R^{\pi}_{\theta}(T) =\left\{
  \begin{array}{ll}
  \Omega\Big(\sqrt{T} \wedge \frac{T}{\delta^{-2}+(n\wedge T)\delta^2+n\sigma^2}\Big),  \quad & \text{if } \delta > \frac{lR}{32|\beta_{\min}|}\sqrt{\frac{|\beta_{\max}|}{K_0e}}T^{-\frac{1}{4}}(\log T)^{-\frac{1}{2}\lambda_0}; \\
  \Omega \big( T\delta^2 \wedge \frac{T}{n\sigma^2}\big),  \quad & \text{if } \delta \le \frac{lR}{32|\beta_{\min}|}\sqrt{\frac{|\beta_{\max}|}{K_0e}}T^{-\frac{1}{4}}(\log T)^{-\frac{1}{2}\lambda_0}  \text{ and } n\sigma^2 >  \frac{l^2R^2|\beta_{\max}|}{512\beta_{\min}^2K_0e}\frac{\sqrt T}{(\log T)^{\lambda_0}}; \\
  \Omega(\frac{\sqrt T}{(\log T)^{\lambda_0}}),  \quad &\text{if } \delta \le \frac{lR}{32|\beta_{\min}|}\sqrt{\frac{|\beta_{\max}|}{K_0e}}T^{-\frac{1}{4}}(\log T)^{-\frac{1}{2}\lambda_0}  \text{ and } n\sigma^2 \le  \frac{l^2R^2|\beta_{\max}|}{512\beta_{\min}^2K_0e}\frac{\sqrt T}{(\log T)^{\lambda_0}},\end{array} 
  \right. 
  \end{align*} 	 which implies Theorem~\ref{thm-lower bound-multiple historical prices}. \qed

  \section*{Appendix C. Proof of Proposition \ref{prop-myopic alg} in Section  \ref{subsec-self exploration}} 
Suppose $\theta^* \in \mathcal{C}_0$ and  $\frac{\min_{\theta\in \mathcal C_0} |\psi(\theta)-\bar{p}_{1:n}|}{\max_{\theta_1,\theta_2\in \mathcal C_0} |\psi(\theta_1)-\psi(\theta_2)|}> K$, we have the following inequalities for each $t \ge 1$: 
\begin{align}\label{ineq1-myopic-optimal}
 |p_t^{\text{myopic}}- \bar p_{1:n}| 
 &\ge |\psi(\theta^*)-\bar p_{1:n}|-|\psi(\theta^*)-p_t^{\text{myopic}}| \ge |\psi(\theta^*)-\bar p_{1:n}|-\max_{\theta_1, \theta_2 \in \mathcal{C}_0}|\psi(\theta_1)-\psi(\theta_2)| \nonumber  \\
 &\ge |\psi(\theta^*)-\bar p_{1:n}|-\frac{1}{K} \min_{\theta  \in \mathcal{C}_0} |\psi(\theta)-\bar p_{1:n}| \ge(1-\frac{1}{K})\cdot \delta,
\end{align}
 where the first inequality follows from the triangle inequality, the second inequality holds since $\theta^* \in \mathcal{C}_0$, $p_t^{\text{myopic}} =\psi( \theta_{t-1}^{\text{LS}})$, and $ \theta_{t-1}^{\text{LS}} \in \mathcal{C}_0$ by its definition, and the last inequality holds since
 $\theta^* \in \mathcal{C}_0$. That is to say, events $\{U_{t, 3}: t\ge 1\}$ defined in Lemma~\ref{lemma-proof-multiple historical prices-step2} are automatically satisfied if ignoring the constant  factor   under   assumptions $\theta \in \mathcal{C}_0$ and  $\frac{\min_{\theta \in \mathcal C_0} |\psi(\theta)-\bar{p}_{1:n}|}{\max_{\theta_1,\theta_2\in \mathcal C_0} |\psi(\theta_1)-\psi(\theta_2)|}> K$. 
 
 We next bound the estimation error $||\theta^* -   \theta_t^{\text{LS}}||^2$ for each $t\ge 0$. Suppose  $\theta^* \in\mathcal{C}_{t}$, i.e.,  $|| \theta^* -   \theta_t^{\text{LS}}||^2_{V_{t, n}}\le  w_{t-1}^2$, and therefore, $|| \theta^* -   \theta_t^{\text{LS}}||^2 \le \frac{w_{t-1}^2}{ \lambda_{\min}(V_{t,n})}$. Then it suffices to bound the minimum eigenvalue of $V_{t,n}$ from below by $\Omega\big((n\wedge t)\delta^2+n\sigma^2\big)$.  
Note that  
 \begin{align*}
 \lambda_{\min}(V_{t, n}) = \min_{(x_1, x_2)\in \mathbb{R}^2: x_1^2+x_2^2=1} \left\{\sum_{i=1}^{n}(x_1+\hat p_ix_2)^2+\sum_{s=1}^{t}(x_1+p_sx_2)^2\right\}+\lambda. 
 \end{align*}
 Let $(x_1^*, x_2^*)$ be the optimal solution to the above optimization problem. Then we  consider the following cases: $|x_2^*| \ge \frac{1}{2(1+2u)}$ and $|x_2^*| < \frac{1}{2(1+2u)}$.

 \underline{Case 1: $|x_2^*| \ge \frac{1}{2(1+2u)}$.} In this case,  we have 
 \begin{align}\label{ineq1-R1}
 \lambda_{\min}(V_{t, n})  &=  \sum_{i=1}^{n}(x_1^*+\hat p_ix_2^*)^2+\sum_{s=1}^{t}(x_1^*+p_sx_2^*)^2 +\lambda \nonumber \\
 &= \sum_{i=1}^{n}\big(x_1^*+\bar p_{1:n}x_2^*)^2+(x_2^*)^2\sum_{i=1}^{n}(\hat p_i-\bar p_{1:n})^2+\sum_{s=1}^{t}(x_1^*+p_sx_2^*)^2+\lambda   \nonumber \\
   & \ge (x_2^*)^2\sum_{s=1}^{n\wedge t} (\bar p_{1:n}-p_s)^2+(x_2^*)^2 n\sigma^2 +\lambda  \nonumber \\
   &\ge \frac{1}{4(1+2u)^2} \Big( \big(1-\frac{1}{K}\big)^2\cdot (n \wedge t )\cdot \delta^2 +n\sigma^2\Big) +\lambda, 
 \end{align}
 where the first inequality follows from  $a^2+b^2 \ge \frac{1}{2}(a-b)^2$,   the second inequality follows from the assumption that $|x_2^*| \ge \frac{1}{2(1+2u)}$, and the last inequality follows from    \eqref{ineq1-myopic-optimal}.

 \underline{Case 2: $|x_2^*| < \frac{1}{2(1+2u)}$.} In this case, since $(x_1^*)^2+(x_2^*)^2=1$, we must have $(x_1^*)^2 \ge 1-\frac{1}{4(1+2u)^2}$, and therefore, 
 \begin{align}\label{ineq2-R1}
 \lambda_{\min}(V_{t, n})  &\ge    
 \sum_{i=1}^{n} \Big((x_1^*)^2  +2x_1^*x_2^*\hat p_i\Big)+\lambda
 \ge   n\big( (x_1^*)^2  -\frac{u}{1+2u}\big)+\lambda \nonumber \\
 &\ge n\big( 1-\frac{1}{4(1+2u)^2} -\frac{u}{1+2u} \big)+\lambda \ge \frac{1}{2}n +\lambda \nonumber \\
 &\ge  \frac{1}{4(u-l)^2}\Big( (n\wedge t)\delta^2+n\sigma^2\Big)+\lambda, 
 \end{align}
 where the second inequality follows from $ 2x_1^*x_2^*\hat p_i \ge -2 u |x_2^*|  \ge -\frac{u}{1+2u}$ due to $|x_1^*| \le 1$ and $|x_2^*| \le \frac{1}{2(1+2u)}$, the third inequality holds since $\frac{1}{4(1+2u)^2} +\frac{u}{1+2u} \le \frac{1}{2(1+2u)}+\frac{u}{1+2u}=\frac{1}{2}$.

 Combining inequalities \eqref{ineq1-R1} and \eqref{ineq2-R1}, when $\theta \in \mathcal{C}_t$ for each $t\ge 0$,  if  $\frac{\min_{\theta\in \mathcal C_0} |\psi(\theta)-\bar{p}_{1:n}|}{\max_{\theta_1,\theta_2\in \mathcal C_0} |\psi(\theta_1)-\psi(\theta_2)|}> K$, we have 
 \begin{align*}
 \lambda_{\min}(V_{t, n}) \ge \min\Big\{\frac{1}{4(1+2u)^2}(1-\frac{1}{K})^2, \frac{1}{4}(u-l)^2 \Big\}\cdot \Big( (n\wedge t)\delta^2+n\sigma^2\Big)+\lambda, 
 \end{align*}
 and thus, 
 \begin{align*}
  ||\theta^* -   \theta_t^{\text{LS}}||^2 \le  \big(\min\{\frac{1}{4(1+2u)^2}(1-\frac{1}{K})^2, \frac{1}{4}(u-l)^2\}\big)^{-1}\frac{   w_{t-1}^2}{(n\wedge t)
  	\delta^2+n\sigma^2}. 
 \end{align*} 
 Therefore, 
 the regret of the myopic policy is upper bounded as follows: 
 \begin{align*}
  & \sum_{t=1}^{T}  \psi(\theta^*)\big(\alpha^*+\beta^* \psi(\theta^*) \big)  -  p_t^{\text{myopic}}\big(\alpha^*+\beta^* p_{t}^{\text{myopic}}\big) \\
  & \, \le |\beta_{\min}| \cdot \sum_{t=1}^{T}  \big( \psi(\theta^*)-\psi(  \theta_{t-1}^{\text{LS}})\big)^2 \\
  &\, \le \frac{|\beta_{\min}|(\alpha_{\max}^2+\beta_{\max}^2)}{4\beta_{\max}^4} \sum_{t=1}^{T}   ||\theta^*-  \theta_{t-1}^{\text{LS}}||^2  \\
  &\le \frac{|\beta_{\min}|(\alpha_{\max}^2+\beta_{\max}^2)}{4\beta_{\max}^4} \sum_{t=1}^{T} \big(\min\{\frac{1}{4(1+2u)^2}(1-\frac{1}{K})^2, \frac{1}{4}(u-l)^2\}\big)^{-1}\frac{  w_{t-1}^2}{(n\wedge t)\delta^2+n\sigma^2} \\ 
  &=\mathcal{O}\Big(\frac{T\log T}{(n\wedge T)\delta^2+n\sigma^2}\Big). 
 \end{align*}
Note  that from Lemma \ref{UCB-concentration}, by letting $\epsilon=\frac{1}{T^2}\wedge \frac{1}{n\sigma^2}$,  we have  with probability $1-\frac{1}{T^2}\wedge \frac{1}{n\sigma^2}$,  $\theta \in \mathcal{C}_t$ for all $0 \le t \le T$. Thus, with probability $1-\frac{1}{T^2}\wedge \frac{1}{n\sigma^2}$,  if the condition $\frac{\min_{\theta \in \mathcal C_0} |\psi(\theta)-\bar{p}_{1:n}|}{\max_{\theta_1,\theta_2\in \mathcal C_0} |\psi(\theta_1)-\psi(\theta_2)|}> K$ holds, the myopic policy achieves the  regret $\widetilde{\mathcal{O}}(\frac{T}{(n\wedge T)\delta^2+n\sigma^2})$.\qed

  \section*{Appendix D. On the Definition of the Optimal Regret}
 \label{subsec-regret def} 
 
 In \S \ref{sec-UCB} and \S \ref{sec-extension-historical price}, we define the optimal regret as
 \begin{align*}
 R^{*}(T, n,  \delta, \sigma)=\inf_{\pi\in\Pi^{\circ}}\sup_{\mathcal{D} \in \mathcal{E}(R);  \atop \theta \in \Theta^{\dag}: |\psi(\theta)-\bar p_{1:n} | \in [(1-\xi)\delta, (1+\xi)\delta]} R^{\pi}_{\theta}(T),
 \end{align*}
 where the environment class is chosen as $\{\theta \in \Theta^{\dag}: |\psi(\theta)-\bar p_{1:n}| \in [(1-\xi)\delta, (1+\xi)\delta]\}$. In this appendix, we give some justifications on this definition of the instance-dependent environment class.

  \subsection*{D.1. Comparison to the ``Worst-Case'' Environment Class} 
 One possible way to define the environment class is to allow the   demand parameter $\theta \in \Theta^{\dag}$ to vary over the entire set $\Theta^{\dag}$. This corresponds to the  {\textit{optimal worst-case regret}} (also known as the \textit{minimax regret}):
 $$
 R^{\text{wc}}(T, n,  \sigma)=\inf_{\pi \in \Pi^{\circ}} \sup_{\mathcal{D} \in \mathcal{E}(R), \theta \in \Theta^{\dag}} R^{\pi}_{\theta}(T).
 $$
 As a byproduct of our results, we can easily characterize the rate of the optimal worst-case regret. 
 \begin{corollary}\label{corollary1}
 	Consider the \texttt{OPOD} problem. Then 	
 	\begin{align*}
 	R^{{\rm{wc}}}(T, n,  \sigma)=  \widetilde \Theta(\sqrt T \wedge \frac{T}{n \sigma^2}). 
 	\end{align*}
 \end{corollary}
 Corollary \ref{corollary1} shows that when  $n \sigma^2$ is within $\widetilde{\mathcal{O}}(\sqrt T)$, the optimal worst-case regret is always $\widetilde \Theta(\sqrt T)$, and when $n \sigma^2$ exceeds $\widetilde \Omega(\sqrt T)$, the optimal worst-case regret  decays according to $\widetilde \Theta( \frac{T}{n \sigma^2})$. This demonstrates that  the offline data may help to reduce the worst-case regret, but only when they are dispersive enough, i.e., $n \sigma^2 \gtrsim 
 \sqrt T$. For example, in the single-historical-price setting with $\sigma=0$, even if the seller has infinitely many offline data, i.e., $n=\infty$, the best achievable worst-case regret is still $\widetilde{\Theta}(\sqrt T)$, and does not improve over the classical setting where there is no offline data. This suggests that the optimal worst-case regret may fail to fully and precisely reflect the value of the offline data (especially when they are not so dispersive), and the goal of achieving the optimal worst-case regret may be too weak. Indeed, the worst case seldom happens in reality and the decision makers are more interested in the actually incurred regret. The offline data thus should play a more powerful role, not only to reduce the regret in the (rare) worst-case scenario, but also to reduce the regret in a per-instance way. The value of the offline data should also be characterized more precisely.
 
 Observing that the  definition of the optimal worst-case regret and the choice of the environment class $\Theta^{\dag}$ are too conservative, we  consider a less conservative environment class by restricting $|\psi(\theta)-\bar p_{1:n}|$ to have the same order as $\delta$ (note that our algorithm does not need to known $\delta$).  
 The resulting $\widetilde \Theta(\sqrt T \wedge \frac{T}{n \sigma^2+(n\wedge T)\delta^2})$ optimal instance-dependent regret significantly improves the $\widetilde \Theta(\sqrt T)$ optimal worst-case regret when $\delta$ is large enough, thus better characterizing the value of offline data. Our results imply that the location of the offline data is an important metric that intrinsically affects the statistical complexity of the \texttt{OPOD} problem.  To the best of knowledge, our results provide the first tight and general instance-dependent regret bounds for the dynamic pricing problem with an unknown linear demand model\footnote{We note that \cite{broder2012dynamic}, \cite{KeskinZeevi2014} and \cite{qiang2016dynamic} provide $\widetilde\Theta(\log T)$ regret bounds for this dynamic pricing problem under certain separability assumptions. However, they do not obtain a regret bound that directly depends on the instance parameters in a tight way.}, with the help of offline data.

 \subsection*{D.2.  Comparison to the ``Local'' Environment Class}  
 Another possible way to define the instance-dependent regret is to choose the  environment class as the set of all the demand parameters $\theta\in \Theta^{\dag}$ such that $|\psi(\theta)-\bar p_{1:n}|$ exactly equals the generalized distance $\delta$,  i.e., $\{\theta \in \Theta^{\dag}: |\psi(\theta)-\bar p_{1:n}|=\delta\}$.  This leads to the following definition of the \textit{local optimal regret}:
 $$R^{\text{loc}}(T,n,\delta, \sigma)=\inf_{\pi \in \Pi^{\circ}}  \sup_{\mathcal{D} \in \mathcal{E}(R);  \atop\theta \in \Theta^{\dag}: |\psi(\theta)-\bar{p}_{1:n}| = \delta} R^{\pi}_{\theta}(T).$$
 With this definition, we can establish the following result on  $R^{\text{loc}}(T,n, \sigma,\delta)$ when $\sigma=0$ and $\delta=\Theta(1)$, whose proof is deferred to Appendix D.3. 
 \begin{proposition}\label{prop-regret-delta}
 	Consider the \texttt{OPOD} problem with a single historical price $\hat p$. When $\delta=\Theta(1)$,   we have 
 	\begin{align*}
 	R^{\rm{loc}}(T,n,\delta):=  R^{\rm{loc}}(T,n, \delta, 0)=\left\{  
 	\begin{array}{ll}
 	\widetilde \Theta(\sqrt T),  \quad & \text{if } n \lesssim  \sqrt T; %
 	\\ 
 \widetilde{\Theta}(\log T ),  \quad & \text{if } n  \gtrsim \sqrt T. 	\end{array} 
 	\right. 
 	\end{align*}
 \end{proposition}

  Note that when $\sqrt{T}\lesssim n\lesssim T$, the local optimal regret $R^{\text{loc}}(T,n,\delta)=\widetilde{\Theta}(\log T)$ is significantly smaller than  the  optimal instance-dependent regret $R^{*}(T,n,\delta)=\widetilde{\Theta}(\frac{T}{n})$. But why does this happen?  The caveat is that the rate of $R^{\text{loc}}(T,n,\delta)$ is meaningless in the sense that it cannot be uniformly achieved by any single algorithm! That is to say, if we consider multiple different values of  $\delta$, e.g., $\delta=1, \delta=1.1,\delta=1.11,\ldots$, while  $R^{\text{loc}}(T,n,1)=\widetilde{\Theta}(\log T), R^{\text{loc}}(T,n,1.1)=\widetilde{\Theta}(\log T), R^{\text{loc}}(T,n,1.11)=\widetilde{\Theta}(\log T),\ldots$, they are  actually achieved by \textit{different} algorithms that are specially designed for $\delta=1, \delta=1.1,\delta=1.11,\ldots$ respectively, and there is no algorithm that can achieve $R^{\text{loc}}(T,n,\delta)=\widetilde{\Theta}(\log T)$ for all of $\delta=1, \delta=1.1,\delta=1.11,\ldots$ simultaneously. 
 	
 	To see this, we   give a concrete  algorithm $\tilde \pi\in\Pi^{\circ}$  that  achieves the regret  of $ \mathcal{O}(\log T)$  for some  specific  value of $\delta=\delta_0$ but incurs the regret of $\Omega(\sqrt T)$ for $\delta=\delta_0+T^{-\frac{1}{4}}$.  
 	The algorithm is named as ``Speculator$(\delta_0)$'' and is presented in Algorithm \ref{Alg-speculator}.
 	When $\delta=1$, and $n=\sqrt T$, the Speculator(1) algorithm incurs the regret of $\mathcal{O}(\log T)$ in the first stage and constant regret in the second stage. However, when $\delta=1+T^{-\frac{1}{4}}$,  %
 	the Speculator(1) algorithm must incur the regret of $\Omega({T}\times (T^{-\frac{1}{4}})^2)  =\Omega(\sqrt T)$ in the second stage, {since with high probability, the algorithm mistakenly charges $\hat{p}+\delta_0$ or $\hat{p}-\delta_0$ for the whole second stage.}

 	\begin{algorithm}[htbp] 	\label{Alg-speculator}
 		\caption{Speculator($\delta_0$): an algorithm that bets $\delta=\delta_0$}
 		\textbf{Input}: specific guess $\delta_0$, historical price $\hat p$, offline demand data $\hat D_1, \hat D_2, \ldots, \hat D_n$,  support of unknown parameters $\Theta^{\dag}$, support of feasible price $[l,u]$, length of the selling horizon $T$\; %
 		\While{$t \in [\lfloor \sqrt T \rfloor]$}{Treat the prices $\hat p+\delta_0$ and $\hat p-\delta_0$ as two arms, and run the UCB algorithm for the two-armed bandits;}  
 		Construct the confidence interval $\tilde C$ for the optimal price based on the least square regression on both the offline and online data\;
 		\eIf{$\hat p + \delta_0 \in \tilde C$ (\text{or } $\hat p - \delta_0 \in \tilde C$)}{Charge the price $\hat p+\delta_0$ (or $\hat p - \delta_0$) when $t= [\lfloor \sqrt T \rfloor]+1, \ldots, T$;}{Charge the myopic price from the least square estimation when $t= [\lfloor \sqrt T \rfloor]+1, \ldots, T$.}
 	\end{algorithm}

 	In fact,  with the above definition of the local optimal regret,  
 	any learning algorithm faces the above dilemma, i.e., its regret is not universally optimal when $\delta$ changes,  
 	and the reason  is as follows.  
 	Using KL-divergence arguments, we can show that when $n=\Theta(\sqrt{T})$, for any $\delta=\Theta(1)$ and any policy $\pi$, the sum of the local instance-dependent regrets under $\delta$ and $\delta+\Theta(T^{-\frac{1}{4}})$ is lower bounded by $\Omega(\sqrt T)$, i.e., 
 	\begin{align*}
 	\sup_{\mathcal{D} \in \mathcal{E}(R);  \atop\theta \in \Theta^{\dag}: |\psi(\theta)-\hat p| = \delta}R^{\pi}_{\theta}(T)+\sup_{\mathcal{D} \in \mathcal{E}(R);  \atop \theta \in \Theta^{\dag}: |\psi(\theta)-\hat p| = \delta+\Theta(T^{-\frac{1}{4}})}R^{\pi}_{\theta}(T) =\Omega(\sqrt T), 
 	\end{align*}   
 	which implies that  for any policy $\pi$, under  at least  one problem instance, i.e.,  $\delta$  or $\delta+\Theta(T^{-\frac{1}{4}})$, the regret  is greater than $\Omega(\sqrt T)$. 
 	The huge gap between $\Omega(\sqrt{T})$ and $\Theta(\log T)$ implies that when $n=\Theta(\sqrt T)$, the optimal rate of $R^{\text{loc}}(T,n,\delta)$  defined in Proposition \ref{prop-regret-delta} cannot be achieved by a single learning algorithm for different values of $\delta$. Thus, $R^{\text{loc}}(T,n,\delta)$ fails to be a valid complexity measure for the \texttt{OPOD} problem.
 	In fact, the statistical complexity of an online pricing problem heavily relies on the fact that there are infinitely many continuous and ``indistinguishable'' prices. If we  directly define the environment class as $\{\theta \in \Theta^{\dag}: |\psi(\theta)-\hat{p}|=\delta\}$, then the resulting $R^{\text{loc}}(T,n,\delta)$ becomes too ``sensitive and specific'' to two discrete prices $\hat{p}+\delta$ and $\hat{p}-\delta$, leaving chances for an algorithm that ``bets $\delta=\delta_0$'' to perform ``abnormally well'' when $\delta$ happens to be $\delta_0$.   By contrast, under the definition of the   optimal regret $R^*(T,n,\delta)$ in \S \ref{sec-UCB} and \S \ref{sec-extension-historical price}, we can design a  learning algorithm that  uniformly  achieves the optimal regret rate for any possible value  of $\delta$.

\subsection*{D.3. Proof of Proposition \ref{prop-regret-delta} in Appendix D.2}
  The proof will be divided into proving the regret lower bound and regret upper bound respectively. 
 	
 	\underline{\textbf{Lower bound: Case 1.}} We first prove that when $n \le \frac{R^2l^2|\beta_{\max}|}{256\beta_{\min}^2K_0e\delta^2}\sqrt T$, for any admissible policy $\pi \in \Pi^{\circ}$, and any $\theta \in \Theta^{\dag}$ with $\frac{\alpha}{-2\beta}=\hat p+\delta$, the regret is lower bounded by $\Omega(\sqrt T)$. To see this, we construct two problem instances $\theta_1=(\alpha_1, \beta_1)$ and $\theta_2=(\alpha_2, \beta_2)$ satisfying the following conditions: 
 	\begin{align*}
 	-\frac{\alpha_1}{2\beta_1} = \hat p+\delta, \quad -\frac{\alpha_2}{2\beta_2} = \hat p+\delta+\Delta, \quad (\alpha_1-\alpha_2)+(\beta_1-\beta_2)(\hat p+\delta)=0.
 	\end{align*}
 	where the value of $\Delta$ is to be specified. 
 	That is, the optimal price under the two problem instances is $\hat p+\delta$ and $\hat p+\delta+\Delta$ respectively, and the two demand functions intersect at the price $\hat p+\delta$. Using similar arguments in inequality \eqref{proof-thm1-samll delta-eq4}, we have 
 	\begin{align*}
 	KL(P_1^{\pi}, P_2^{\pi}) \le \frac{2\beta_{\min}^2\Delta^2}{R^2l^2}\Big(
 	n\delta^2+\sum_{t=1}^{T}\mathbb{E}_{\theta_1}^{\pi}[(p_t-\psi(\theta_1))^2]\Big). \end{align*}
 	In addition, by defining the two disjoint intervals $I_1=[\hat p+\delta-\frac{1}{4}\Delta, \hat p+\delta+\frac{1}{4}\Delta]$ and $I_2=[\hat p+\delta+\frac{3}{4}\Delta, \hat p+\delta+\frac{5}{4}\Delta]$, and using similar arguments to inequality \eqref{proof-thm1-samll delta-eq2}, we have the following lower bound on the sum of regret under $\theta_1$ and $\theta_2$: 
 	\begin{align*}
 	R_{\theta_1}^{\pi}(T)+R_{\theta_2}^{\pi}(T) \ge \frac{1}{32}|\beta_{\max}| \cdot e^{-KL(P_1^{\pi}, P_2^{\pi})} \cdot T\Delta^2. 
 	\end{align*}
 	Since $\pi \in \Pi^{\circ}$, we further have 
 	\begin{align*}
 	KL(P_1^{\pi}, P_2^{\pi})\ge \log (\sqrt T \Delta^2)+\log\Big(\frac{|\beta_{\max}|}{64K_0}\Big).
 	\end{align*}
 	Thus,   by letting $\Delta^2 = \frac{64K_0e}{\sqrt{T}|\beta_{\max}|}$, we have 
 	\begin{align*}
 	R_{\theta_1}^{\pi}(T) 
 	&\ge  |\beta_{\max}| \sum_{t=1}^{T}\mathbb{E}_{\theta_1}^{\pi}[(p_t-\psi(\theta_1))^2] \\
 	& \ge   |\beta_{\max}| \Big(\frac{R^2l^2 }{2\beta_{\min}^2\Delta^2}\cdot    \big(\log(\sqrt T \Delta^2)+\log \frac{|\beta_{\max}|}{64K_0}\big)-  n\delta^2\Big)\\
 	& =  |\beta_{\max}| \Big(\frac{R^2l^2 |\beta_{\max}|}{128\beta_{\min}^2 K_0e}\sqrt T -  n\delta^2\Big)\\&\ge \frac{R^2l^2 \beta_{\max}^2 }{256\beta_{\min}^2 K_0e}\sqrt T,
 	\end{align*}
 where the equation follows from the choice of $\Delta$, and the last inequality holds since $n \le \frac{R^2l^2|\beta_{\max}|}{256\beta_{\min}^2K_0e\delta^2}\sqrt T$.

 	\underline{\textbf{Lower bound: Case 2.}} We then prove when $n > \frac{R^2l^2|\beta_{\max}|}{256\beta_{\min}^2K_0e\delta^2}\sqrt T$, for any policy $\pi$ (not necessarily in the admissible policy class), $\max_{\theta\in \Theta^{\dag}: \psi(\theta)-\hat p=\delta} R^{\pi}_{\theta}(T)=\Omega(\log T)$. Since $\psi(\theta)=-\frac{\alpha}{2\beta}$, the constraint for $\theta$ becomes $\{(-2\beta(\hat p+\delta), \beta): \beta \in [\beta_{\min} \vee \frac{\alpha_{\max}}{-2(\hat p+\delta)},\beta_{\max} \wedge  \frac{\alpha_{\min}}{-2(\hat p+\delta)}]\}$. In this case,  the problem is reduced to a single-dimensional problem, and it suffices to prove that there exists some $\beta \in [\beta_{\min} \vee \frac{\alpha_{\max}}{-2(\hat p+\delta)},\beta_{\max} \wedge  \frac{\alpha_{\min}}{-2(\hat p+\delta)}]$, such that $
 	R^{\pi}_{\theta}(T)=\Omega(\log T)$, 
 	where $\theta=(-2\beta(\hat p+\delta), \beta)$.

 	To this end, we invoke again the van Trees inequality   in Lemma~\eqref{lemma-van trees ineq}, by letting $C(\theta)=(-\hat p\,\,1)$,  and $q(\cdot): \mathbb{R}\to \mathbb{R}^+$ be  an absolutely continuous density on  $\beta\in[\beta_{\min} \vee \frac{\alpha_{\max}}{-2(\hat p+\delta)},\beta_{\max} \wedge  \frac{\alpha_{\min}}{-2(\hat p+\delta)}]$ with positive value on $(\beta_{\min} \vee \frac{\alpha_{\max}}{-2(\hat p+\delta)},\beta_{\max} \wedge  \frac{\alpha_{\min}}{-2(\hat p+\delta)})$ and zero on the boundary $\{\beta_{\min} \vee \frac{\alpha_{\max}}{-2(\hat p+\delta)},\beta_{\max} \wedge  \frac{\alpha_{\min}}{-2(\hat p+\delta)}\}$. In this case, similar to the first lower bound in Step 1 of the proof of Theorem \ref{thm-lower bound}, we obtain the following inequality for $\theta=(-2\beta(\hat p+\delta), \beta)$: 
 	\begin{align*}
 	\sum_{t=1}^{T} \mathbb{E}_{  q}\big[\mathbb{E}_{\theta}^{\pi}[(p_t-\psi(\theta))^2]\big] \ge \sum_{t=2}^{T} \frac{R^2c_1'}{{R^2\mathcal{I}}( q)+\sum_{s=1}^{t-1}\mathbb{E}_{q}[\mathbb{E}_{\theta}^{\pi}[(p_s-\hat p)^2]]} \ge \sum_{t=2}^{T} \frac{R^2c_1'}{{R^2\mathcal{I}}(q)+(t-1)(u-l)^2}, 
 	\end{align*}
 	where $c_1'=(\min_{\theta\in \Theta^{\dag}: \psi(\theta)=\hat p+\delta}\frac{\alpha+\beta \hat p}{2\beta^2})^2$, and $\mathcal{I}(q)$ is defined in Lemma~\eqref{lemma-van trees ineq}. Since both $c_1'$ and $\mathcal{I}(q)$ are constants (recall that $\delta$ is assumed to be a constant), then we have $\max_{\beta \in [\beta_{\min} \vee \frac{\alpha_{\max}}{-2(\hat p+\delta)},\beta_{\max} \wedge  \frac{\alpha_{\min}}{-2(\hat p+\delta)}]}R_{\theta}^{\pi}(T) =\Omega(\log T)$, which completes the proof.

 	$\underline{\textbf{Upper bound.}}$ When $n<\sqrt T$, from Theorem \ref{thm-UCB},  O3FU algorithm is admissible and achieves the regret upper bound ${\mathcal{O}}(\sqrt T)$, which matches the lower bound proven in the above Case 1.  In the following, we first prove that when $n \ge \sqrt T$,  Speculator($\delta_0$)  achieves the regret upper bound ${\mathcal{O}}(\sqrt T)$ for any $\theta \in 
 	\Theta^{\dag}$, and therefore is admissible. Then we will prove that when $\delta=\delta_0$,  Speculator($\delta_0$)  achieves the regret upper bound  $\mathcal{O}(\log T)$.

 	When $\theta^*$ is arbitrary,  $\delta=|\psi(\theta^*)-\hat p|$ is also arbitrary and not necessarily equals $\delta_0$,  the regret in the first $\sqrt T$ periods is $\mathcal{O}(\sqrt T)$ due to at most a constant loss in each period. In addition, it can be easily verified that the sum of squared dispersion for  $n$ offline prices and  $\lfloor\sqrt T\rfloor$ online prices  is lower bounded by $\Omega(\sqrt T)$. Specifically, from \eqref{ineq-variance p lower bound}, for $\hat p_1=\ldots=\hat p_n=\hat p$ and $p_t \in \{\hat p+\delta_0, \hat p-\delta_0\}$ for each $t \in [\lfloor \sqrt T \rfloor]$, we have  
 		\begin{align*}
 	J(\hat p_1, \ldots, \hat p_n, p_1, \ldots, p_{\lfloor \sqrt T \rfloor})  
 	&\ge J(\hat p_1, \ldots, \hat p_n)+ \frac{n}{n+\sqrt T}\sum_{s=1}^{\lfloor \sqrt T \rfloor}(p_s-\bar p_{1:n})^2 =\frac{n\lfloor \sqrt T \rfloor}{n+\lfloor \sqrt T \rfloor}\delta_0^2 \gtrsim  \sqrt T \delta_0^2.
 	\end{align*}
 Therefore, $\lambda_{\min}(V_{\lfloor \sqrt T \rfloor, n})=\Omega(\sqrt T)$, and the squared radius of the confidence interval $\tilde C$ is at most $\Theta(\frac{1}{\lambda_{\min}(V_{\lfloor\sqrt{T} \rfloor, n})})=\Theta(\sqrt T)$. Since  the true optimal price lies in $\tilde C$ with high probability, it follows that for any price within $\tilde C$, its squared deviation from the optimal price in each period $\lfloor \sqrt T \rfloor+1, \ldots, T$  is no more than $\frac{1}{\sqrt T}$, and therefore, the cumulative  revenue loss in periods $\lfloor \sqrt T \rfloor+1,  \ldots, T$ is no more than $\mathcal{O}(\sqrt T)$.

 When  $\delta=\delta_0$, from Theorem 5 in \cite{abbasi2011improved}, the regret of Speculator($\delta_0$) in the first $\lfloor \sqrt T\rfloor$ periods is upper bounded by $\widetilde{\mathcal{O}}(\log T)$. In the remaining periods from $\lfloor \sqrt T\rfloor+1$ to $T$, since the optimal price is either $\hat p+\delta_0$ or  $\hat p-\delta_0$, which belongs to the confidence interval $\tilde C$ with high probability, by construction, Speculator($\delta_0$)  chooses the optimal price from period  $\lfloor \sqrt T\rfloor+1$ to $T$ with high probability. Note that the squared length of $\tilde C$ is $\Theta(\frac{1}{\sqrt T})$, so  $\hat p+\delta_0$ and  $\hat p-\delta_0$ cannot belong to  $\tilde C$ at the same time.  In this case, it can be verified that the regret from period  $\lfloor \sqrt T\rfloor+1$ to $T$   is upper bounded by $\widetilde{\mathcal{O}}(\log T)$.\qed

  \section*{Appendix E. Extension to Generalized Linear Model}
  In this appendix, we discuss the extension of our regret upper bounds to the generalized linear model. For simplicity, we focus on the single-historical-price setting, and leave the discussion on the multiple-historical-price setting  to the interested readers.  
  Consider the following  demand model: 
  \begin{align}\label{def-GLM}
       D_t = g(\alpha^*+\beta^* p_t) +\varepsilon_t, 
  \end{align}
  where $g(\cdot)$ is an increasing function whose   form is known to the  seller (we refer to $g(\cdot)$ as the \textit{link function}),  
  $(\alpha^*, \beta^*)$ is the unknown demand parameter  in the compact set $\Theta^{\dag}$, and $\{\epsilon_t\}_{t\ge 1}$ is a sequence of i.i.d. sub-Gaussian random variables.  We also assume that  the conditional probability of $D_t$ given $p_t$ is   from the exponential family, which is a standard assumption  in the literature, see, e.g., \cite{filippi2010parametric}.   Since  the expected demand function is the composition of the link function $g(\cdot): \mathbb{R} \to \mathbb{R}$ and the linear function $p\mapsto  \alpha^*+\beta^* p $, the above equation \eqref{def-GLM} is referred to as the \textit{generalized linear model} (GLM).  
Similar as before, we let  $\theta:=(\alpha, \beta)$, $r(p; \theta):=p\cdot g(\alpha+\beta p)$ and  $\psi(\theta):=\arg \max_{p\in [\underline{p}, \overline{p}]} r(p;
      	\theta)$. The definition of the regret $R_{\theta^*}^{\pi}(T)$ for any given policy $\pi$  remains the same.

We make the following assumptions on the optimal price $\psi(\theta)$,  the expected revenue $r(p; \theta)$, and  the link function  $g(\cdot)$. 
\begin{assumption}\label{ass-GLM}
There exist   constants $L_0>0$, $0<\lambda_1<\lambda_2$  and  $0<L_1<L_2$, such that
      \begin{itemize}
      	\item[(a)]  
      	$|\psi(\theta_1)-\psi(\theta_2)| \le L_0\cdot ||\theta_1-\theta_2||$ for any   $\theta_1, \theta_2  \in \Theta^{\dag}$;  
      	\item[(b)] $	\lambda_1 \cdot \left(\psi(\theta)-p\right)^2 \le r\left(\psi(\theta); \theta\right) -  r\left(p;\theta\right) \le \lambda_2 \cdot \left(\psi(\theta)-p\right)^2$ for any $ p \in [\underline{p}, \overline p]$ and   $\theta \in \Theta^{\dag}$; 
      	\item[(c)]   $g(x)$ is twice differentiable in $\mathcal{X}:=  \{\alpha+\beta p:  (\alpha, \beta) \in \Theta^{\dag}, p \in [\underline{p}, \overline{p}]\}$,  with  $L_1 \le g'(x) \le L_2$ for any $x\in \mathcal{X}$, and bounded second-order derivative in $\mathcal{X}$. 
      \end{itemize}
\end{assumption}

Condition (a) requires that the optimal price $\psi(\theta)$ 
is Lipschitz continuous in $\Theta^{\dag}$ with Lipschitz constant $L_0$, which is satisfied if $\psi(\cdot)$ is differentiable and the norm of its gradient is upper bounded. Condition (b)  is satisfied if for any $\theta \in \Theta^{\dag}$, the optimal price $\psi(\theta)$ is an interior point of $[\underline  p, \overline p]$, and the second-order derivative of $r(p; \theta)$, with respect to $p$, exists, and is lower bounded by $\lambda_1$ and upper bounded by  $\lambda_2$.    
Condition (c) is similar to Assumptions~1 and 2 in \cite{li2017provably} on the generalized linear contextual bandit, and our condition is slightly stronger to 
make sure that our instance-dependent upper bound holds. 
Note that under condition (c), condition (b) can also be satisfied if for any $\theta\in \Theta^{\dag}$, $\psi(\theta)$ is an interior point of $[\underline{p},\overline{p}]$, and the expected revenue, as a function of the \textit{mean demand}, is concave whose  second-order derivative is lower bounded by $\lambda_1$  and upper bounded by  $\lambda_2$. Note that the concavity of the expected revenue with respect to the mean demand (instead of the price) is more commonly assumed in the literature of  revenue management, see, e.g., \cite{wang2014close}.
   All of  conditions (a)-(c) can be satisfied by the commonly used linear model (i.e., $g(x)=x$),  logit model (i.e.,  $g(x)=\frac{e^x}{1+e^x}$), and exponential model (i.e., $g(x)=e^x$).

 The algorithm   for the generalized linear model \eqref{def-GLM} can be  
 modified from O3FU as follows. Let $\hat \theta_t$ be the following maximum likelihood
estimator (instead of the least-squares estimator in O3FU):
\begin{align*}
     \hat \theta_t:=\argmax_{\theta=(\alpha, \beta) \in \Theta^{\dag}} n \left(\hat D_i \cdot (\alpha+\beta \hat p)-m(\alpha +\beta \hat p)\right)+\sum_{s=1}^t \big( D_s\cdot (\alpha+\beta p_s)-m(\alpha +\beta p_s)\big),
\end{align*}
where $m(\cdot)$ is the function such that $m(\alpha+\beta p)=g'(\alpha+\beta p)$ for any $(\alpha, \beta) \in \Theta^{\dag}$ and   $p \in [l,u]$. Then we let $(p_t, \tilde \theta_t):=
     \argmax_{p \in [l, u], \theta=(\alpha,\beta)\in \mathcal{C}_{t-1}} p\cdot g(\alpha+\beta p)$. Besides, for the confidence ellipsoid $\mathcal{C}_{t-1}$, the confidence radius $w_t$   needs to be modified accordingly by applying the high-probability confidence bound in Lemma~3 of \cite{li2017provably}. 
 We refer to this modified algorithm as O3FU-GLM.

The following proposition establishes a similar regret upper bound for O3FU-GLM  to O3FU in Theorem~\ref{thm-upper bound-multiple prices-T}. 
\begin{proposition}\label{prop-upper bound-GLM}
     	Let $\pi$  be   O3FU-GLM algorithm for the $\texttt{OPOD}$ problem. Then there exists a finite  constant $K_5>0$  such that for any  $T \ge 1$, $n\ge 0$   and $\hat p \in [l,u]$, and for any possible value of $\theta^* \in \Theta^{\dag}$, we have 
 	\begin{align*} 
 	R^{\pi}_{\theta^*}(T)  \le   K_5  \Big( \sqrt T \wedge \frac{T\log T}{(n \wedge T)\delta^2}\Big)\cdot  \log T.   
 	\end{align*}
\end{proposition}

\textit{Proof of Proposition~\ref{prop-upper bound-GLM}.}
Under Assumption~\ref{ass-GLM},  Proposition~\ref{prop-upper bound-GLM} can be proven under a similar framework to  Theorem~\ref{thm-UCB}. 
We next only highlight the main differences and omit the detailed verification. 

First,    to prove the instance-independent upper bound $\widetilde{\mathcal{O}}(\sqrt T\log T)$, we first note the following upper bound on the regret of algorithm O3FU-GLM:  when $\theta^* \in \mathcal{C}_{t-1}$, 
\begin{align*}
    \psi(\theta^*)\cdot g(\alpha^*+\beta^* \psi(\theta^*))-p_t  \cdot g(\alpha^*+\beta^* p_t ) \le  p_t\cdot g(\tilde \alpha_t+\tilde \beta_t p_t)-p_t \cdot g(\alpha^*+\beta^* p_t) \le u\cdot L_2|(\tilde{\theta}_t -\theta^*)^{\top} x_t|,
\end{align*}
where $x_t=[1 \, \, p_t]^{\top}$, the first inequality follows from $\theta^* \in \mathcal{C}_{t-1}$, $\psi(\theta^*) \in [l,u]$ and the definition of $(p_t, \tilde \theta_t)$, and the second inequality follows from  condition~(c) of  Assumption~\ref{ass-GLM} and the mean value theorem. With the above inequality, the regret upper bound  $\widetilde{\mathcal{O}}(\sqrt T\log T)$ can be proven similar to Step 1 of Theorem~\ref{thm-UCB}. In particular, to bound the probability for  the event $\{\theta^* \in \mathcal{C}_t\}_{t\ge 1}$, 
Lemma~3 in \cite{li2017provably}  established for the generalized linear contextual bandit will be useful, and plays a similar role to Theorem~2 in \cite{abbasi2011improved} established for  the linear contextual bandit, which is applied in our previous proof.

Second, to prove the instance-dependent upper bound $\widetilde{\mathcal{O}}(\frac{T(\log T)^2}{(n\wedge T)\delta^2})$, we first note that 
 the regret of  
O3FU-GLM is  upper  bounded by the cumulative estimation error for the true parameter $\theta^*$  as follows: 
\begin{align*}
     \sum_{t=2}^T \mathbb{E}_{\theta^*}^{\pi}\Big[ r\big(\psi(\theta^*); \theta^*\big) -  r\big(p_t;\theta^*\big) \Big] \le \lambda_2   \sum_{t=1}^T \mathbb{E}_{\theta^*}^{\pi}\Big[\big(\psi(\theta^*)-p_t\big)^2 \Big] \le \lambda_2L_0 \sum_{t=1}^T \mathbb{E}_{\theta^*}^{\pi}\big[||\theta^* -\tilde \theta_t||^2 \big],
\end{align*}
where the two inequalities hold  due to condition (b)  and condition (a) in Assumption~\ref{ass-GLM}  respectively.  
With the above inequality,  it suffices to establish  Lemma~\ref{lemma-proof-UCB-step2} for O3FU-GLM. To this end, we also start from  the same inequality to \eqref{ineq1-proof-lemma8} in Step 2 of the proof of Lemma~\ref{lemma-proof-UCB-step2}, and discuss the same  three cases.  For Case  1, Case 2 and  Case 3.1, the proof is similar under  condition (a) in Assumption~\ref{ass-GLM}  that  $\psi(\cdot)$  is Lipschitz continuous. For Case 3.2,  the crucial step is to show inequality~\eqref{ineq5-proof-lemma8}, whose proof can be modified by invoking conditions (b) and (c) in Assumption~\ref{ass-GLM}. Specifically, we refine $A_1$, $A_2$, $A_3$ and $A_4$ as 
\begin{align*}
      &A_1=p_t \cdot g(\tilde{\alpha}_t+\tilde \beta_t p_t), \quad A_2=p_t\cdot g(\alpha^*+ \beta^* p_t),\\
      &A_3=\psi(\theta^*)\cdot g(\tilde{\alpha}_t+\tilde \beta_t \psi(\theta^*)),  \quad  A_4=\psi(\theta^*)\cdot g(\alpha^*+\beta ^*\psi(\theta^*)),
\end{align*}
and inequalities in  \eqref{ineq3-proof-lemma8}  and \eqref{ineq4-proof-lemma8} continue to hold, i.e., 
\begin{align*}
     A_1\ge A_3, \quad A_1\ge A_4\ge A_2. 
\end{align*} 
For the    case when $A_3\ge A_2$, inequality \eqref{ineq6-proof-lemma8}  will be modified to  \begin{align*}  
 	 |\Delta \alpha_t+\Delta \beta_t p_t| 
 	 &\ge  \frac{1}{L_2}\cdot \frac{ A_1-A_2 }{p_t} \\
 	 &\ge \frac{1}{L_2} \cdot \frac{|A_4-A_3| }{p_t}  \\
 	 & = \frac{1}{L_2} \cdot \frac{\psi(\theta^*)}{p_t} \cdot   \left|g(\alpha^*+\beta^* \psi(\theta^*))-g(\tilde \alpha_t+\tilde \beta_t \psi(\theta^*))\right|  \\
 	 &
 	 \ge  \frac{L_1}{L_2}\cdot \frac{\psi(\theta^*)}{p_t} |\Delta \alpha_t+\Delta \beta_t \psi(\theta^*)| \\
 	 &\ge  \frac{L_1}{L_2}\cdot \frac{l}{u} |\Delta \alpha_t+\Delta \beta_t \psi(\theta^*)|, 
 	 \end{align*} 
 	 where the first inequality follows from $|g(x)-g(y)| \le L_2|x-y|$ guaranteed by condition (c) of Assumption~\ref{ass-GLM} and the mean value theorem, the second inequality holds since $A_3, A_4\in [A_2, A_1]$, and the third inequality holds due to  $|g(x)-g(y)| \ge L_1|x-y|$ guaranteed by condition (c) of Assumption~\ref{ass-GLM} and the mean value theorem.   
For the   case  when $A_3 < A_2$,  inequality \eqref{ineq7-proof-lemma8} will be modified to 
 \begin{align*} 
 	 |\Delta \alpha_t+\Delta \beta_t p_t| 
 	 &\ge \frac{1}{L_2}\cdot \frac{ A_1-A_2 }{p_t}\\
 	& \ge  \frac{1}{L_2}\cdot \frac{ A_4-A_2 }{p_t} \\
 	&=\frac{1}{L_2} \cdot \frac{r(\psi(\theta^*); \theta^*)-r(p_t;\theta^*)}{p_t}\\ 
 	&\ge \frac{\lambda_1}{L_2}\cdot \frac{(\psi(\theta^*)-p_t)^2}{p_t}\\ 
 	&\ge \frac{\lambda_1}{L_2 \lambda_2 p_t}\cdot    |A_1-A_3|\\ 
 	 	&\ge \frac{\lambda_1}{L_2 \lambda_2 p_t}\cdot    |A_4-A_3|\\ 
 	&\ge  \frac{\lambda_1 L_1}{L_2 \lambda_2}\cdot  \frac{\psi(\theta^*)}{p_t}  \cdot 
 	|\Delta \alpha_t+\Delta \beta_t \psi(\theta^*)|\\
 	&\ge \frac{\lambda_1 L_1}{L_2 \lambda_2}\cdot  \frac{l}{u}  \cdot 
 	|\Delta \alpha_t+\Delta \beta_t \psi(\theta^*)|, 
 	 \end{align*}
 	 where  the third and fourth inequalities follow  from condition (b) in Assumption~\ref{ass-GLM}, the fifth inequality follows from the assumption that $A_3 <A_2$,  and the sixth inequality follows from condition (c) in Assumption
 	 ~\ref{ass-GLM} and the mean value theorem. The remaining analysis for Case 3.2 is similar, whose details are therefore omitted. \qed

 \section*{Appendix F. Extension to  Adaptive Offline Data}
 \label{subsec-correlated offline data}
 In this appendix,  we extend our main results to the setting that 
 in the offline stage, the seller's pricing decisions are made adaptively  based on the previous price and sales data  according to some possibly unknown policy $\hat \pi$. 
 Therefore, for each $i=2, \ldots, n$, 
 $\hat p_i$ may depend on the previous data $\hat p_1, \hat D_1, \ldots, \hat p_{i-1}, \hat D_{i-1}$.

 When  the offline data  are generated adaptively according to some possibly unknown  policy $\hat \pi$, the historical price $\hat  p_i$ is a function of $\hat p_1, \hat D_1, \ldots, \hat p_{i-1}, \hat D_{i-1}$,  for each $i=2, \ldots, n$, which contains uncertainty arising from the random noise, and therefore is a random variable.  Nevertheless, in many practical scenarios, the seller's primary
 concern is to understand the effect of this \textit{particular}  pricing sequence $\{\hat p_1, \ldots, \hat p_n\}$ on the online learning process. 
 Thus, we will measure the performance of a learning algorithm via the conditional  expected revenue  given the realization of $\hat p_1, \ldots, \hat p_n$, and study the impact of this exact sequence  on the online learning process. 
 
 Specifically, for any pricing  policy $\pi$, let  $  R^{\pi}_{\theta^*}(T, \hat p_1, \ldots, \hat p_n)$  be  defined as the conditional regret as follows: 
 \begin{align*}
 R^{\pi}_{\theta^*}(T,\hat p_1, \ldots, \hat p_n) = \mathbb{E}^{ \pi}_{\theta^*}\big[Tr^*(\theta^*)-\sum_{t=1}^T p_t(\alpha^*+\beta^* p_t) \big|\hat p_1, \ldots, \hat p_n
 \big]. 
 \end{align*} 
 For any pricing policy $\pi$, it is said to be admissible if there exists some constant $K_0>0$ such that $  R^{\pi}_{\theta^*}(T,\hat p_1, \ldots, \hat p_n)  \le K_0 \sqrt T \log T$, for any $T \ge 1$, $n \ge 0$, $\theta^* \in \Theta^{\dag}$, and $\hat p_1, \ldots, \hat p_n \in [l,u]$. Let  
 $\hat  \Pi^{\circ}$ be  the set of all admissible policies. For notation  convenience, we also define $\hat \delta=|\bar p_{1:n}-\psi(\theta^*)|$,  
 and $\hat \sigma=\sqrt{\frac{1}{n}\sum_{i=1}^{n}(\hat p_i-\bar p_{1:n})^2}$, both of which depend on the realizations of $\hat p_1, \ldots, \hat p_n$. We provide matching upper and lower bounds on regret in Proposition \ref{prop-extension}, which indicates that   M-O3FU algorithm remains optimal even for adaptive offline data. 
 
 \begin{proposition}\label{prop-extension}
 	Consider the \texttt{OPOD} problem with the offline data generated from some  possibly unknown policy $\hat \pi$. 
 	\begin{itemize}
 		\item[(a)] Let $\pi$ be   M-O3FU algorithm.  For   any sample path of historical prices $\hat  p_1, \ldots \hat p_n$,  $T \ge 1$, $n \ge 1$, and any possible value of $\theta^* \in \Theta^{\dag}$, 
 		\begin{align*}
 		R^{\pi}_{\theta^*}(T, \hat p_1, \ldots, \hat p_n) =  
 		\left\{  
 		\begin{array}{ll}
 		\widetilde{\mathcal{O}} \big(T \hat \delta^2 +1\big),  \quad & \text{if } \hat\delta^2  \lesssim \frac{1}{n\hat \sigma^2} \lesssim \frac{1}{\sqrt T};\\
\widetilde{\mathcal{O}} \big( \sqrt T \wedge \frac{T }{(n\wedge T) \hat \delta^2 +n\hat \sigma^2}\big),  \quad & \text{otherwise}.  
 		\end{array} 
 		\right.  
 		\end{align*} 
 		\item[(b)] For any pricing policy $\pi$, 
 	  $T \ge 2$, $n \ge 1$, $\hat \delta \in [0,u-l]$,  and realization of $\hat p_1, \ldots, \hat  p_n \in [l,u]$,  
 		\begin{align*}
 		\sup_{ \mathcal{D} \in \mathcal{E}(R);  \atop \theta \in \Theta^{\dag}: |\bar p_{1:n}-\psi(\theta)| \in [(1-\xi)\hat\delta,(1+\xi)\hat\delta]}    R^{\pi}_{\theta}(T, \hat p_1, \ldots, \hat p_n) =\widetilde{\Omega}(\sqrt T\wedge \frac{T}{{\hat\delta}^{-2}+(n \wedge T)  {\hat \delta}^{2}+n\hat \sigma^2}).
 		\end{align*}  If for any value of $\theta \in \Theta^{\dag}$,  $\mathbb{E}^{\hat \pi}_{\theta}[\hat \delta(\theta)] \lesssim T^{-\frac{1}{4}}(\log T)^{-\frac{1}{2}}$  and $\mathbb{E}^{\hat \pi}_{\theta}[n \hat \sigma^2] \lesssim \frac{\sqrt T}{\log T}$ (where the expectation is taken over $\hat p_1, \hat p_2, \ldots, \hat p_n$), then for any admissible policy $\pi \in \hat \Pi^{\circ}$, $T \ge 2$, $n \ge 1$,  $\theta^* \in \Theta^{\dag}$, $
 		\mathbb{E}^{\hat \pi}[ R_{\theta}^{\pi}(T, \hat p_1, \ldots, \hat p_n)] = \widetilde \Omega(\sqrt T)$. 
 	\end{itemize}  
 	
 \end{proposition}
 
\textit{Proof of Proposition \ref{prop-extension}.}
  The proof is similar to   Theorem \ref{thm-upper bound-multiple prices-T} and Theorem \ref{thm-lower bound-multiple historical prices}, and we  only highlight the   differences and omit detailed analysis. 
 
 (a) Similar to the proof of Theorem \ref{thm-upper bound-multiple prices-T}, we need to show the two upper bounds $\mathcal{O}(\sqrt T \log T)$ and $\mathcal{O}(\frac{T(\log T)^2}{n\hat \sigma^2+(n \wedge T)\hat \delta^2})$. 
 
 To see the first upper bound, if conditioning on the realization of  $\hat  p_1, \ldots, \hat  p_t$,   the upper bound  on $\sum_{t=1}^{T} ||x_t||^2_{V_{t-1,n}^{-1}}$, i.e., \eqref{ineq-1-proof-thm4}, and the concentration inequality in Lemma  \ref{UCB-concentration}  still hold, then we can apply similar arguments to Step 1 in the proof of Theorem \ref{thm-upper bound-multiple prices-T} to obtain the first upper bound $\mathcal{O}(\sqrt T \log T)$. To this end, we notice that the upper bound  \eqref{ineq-1-proof-thm4} is derived from Lemma  \ref{proof-UCB-lemma1}, and in the statement of Lemma  \ref{proof-UCB-lemma1},  the sequence $\{X_t:t \ge 1\}$ and the matrix $V$  can be arbitrary. Therefore, for any given realization of $\hat p_1, \ldots, \hat  p_n$, by letting $V=\lambda I +\sum_{i=1}^{n}x_ix_i^{\top}$, we have similar upper bound on $\sum_{t=1}^{T} ||x_t||^2_{V_{t-1,n}^{-1}}$. Moreover, the key ingredient to prove  Lemma  \ref{UCB-concentration}  in \cite{abbasi2011improved} is their Theorem 1. 
 {For any given realization of $\hat p_1, \ldots, \hat  p_n$, 
 	Theorem 1 in \cite{abbasi2011improved} continues to hold, and therefore, the  bound for the conditional probability given $\hat p_1, \ldots \hat p_n$ in Lemma   \ref{UCB-concentration}   also holds.}
 
 To see the second upper bound, it suffices to establish the concentration inequality in Lemma  \ref{UCB-concentration},    
 the sample-path inequality in  Lemma   \ref{lemma-proof-multiple historical prices-step2} and Lemma \ref{appendix-lemma-thm4}. As discussed above, given realization of $\hat p_1, \ldots, \hat p_n$, Lemma  \ref{UCB-concentration} continues to hold.  In both Lemma   \ref{lemma-proof-multiple historical prices-step2} and Lemma \ref{appendix-lemma-thm4}, we conduct the sample-path analysis and treat each quantity as an arbitrary and deterministic number. Therefore,  conditioning on the realization of  $\hat  p_1, \ldots, \hat  p_t$,  Lemma  \ref{lemma-proof-multiple historical prices-step2} and Lemma \ref{appendix-lemma-thm4} also continue to hold.  
 
 (b) We divide the proof for the lower bound into two steps.

 \underline{\textbf{Lower bound: Step  1.}} In this step,  similar  to  \eqref{ineq1-proof-thm3}, we  prove  
 for any   policy $\pi$, 
 \begin{align}\label{ineq1-proof-prop}
 \sup_{ \theta \in \Theta_0(\hat \delta, \hat p_1, \ldots, \hat p_n)} R^{\pi}_{\theta}(T, \hat p_1, \ldots, \hat p_n)=\Omega\Big(\sqrt T 
 \wedge \frac{T}{ {\hat \delta}^{-2}+n \hat \sigma^2+(n \wedge T){\hat \delta}^2}\Big),
 \end{align}
 where  $\Theta_0(\hat \delta, \hat p_1, \ldots, \hat p_n)= \big\{\theta \in \Theta^{\dag}: \, \psi(\theta)-\bar p_{1:n}\in [\frac{1}{2} \hat \delta, \hat \delta] \big\}$.  Here we highlight the dependence of    set $\Theta_0$  on $\hat p_1, \ldots, \hat p_n$.

 To see \eqref{ineq1-proof-prop}, we first note that since $l \le \bar p_{1:n} \le u$, $\Theta_0'(\hat \delta):=\{\theta\in \Theta^{\dag}: \psi(\theta)-u \ge \frac{1}{2}\hat\delta, \, \psi(\theta)-l\le \hat\delta\}$ must be a subset of $\Theta_0(\hat\delta, \hat p_1, \ldots, \hat p_n)$. From the definition of $\Theta_0'(\hat\delta)$, there exist  some positive constants $x_0$, $y_0$, $\epsilon$ such that  
 $\Theta_1(\hat\delta) :=[x_0-\frac{1}{2}\epsilon \hat \delta, x_0+\frac{3}{2}\epsilon \hat \delta ]\times [y_0-\frac{1}{2}\epsilon  \hat\delta, y_0+\frac{3}{2}\epsilon  \hat\delta] \subseteq \Theta_0'(\hat\delta) $.  
 Then we define a  prior distribution $q(\cdot)$  for the unknown parameter $\theta$ on the set $\Theta_1(\hat\delta)$, whose expression is the same as   \eqref{ineq1-proof-thm3}.
 The remaining proof is similar to  that of  \eqref{ineq1-proof-thm3} as long as when applying the van Trees inequality, we consider the expectation $\mathbb{E}_q[\mathbb{E}_{\theta}^{\pi}[(p_t-\psi(\theta))^2|\hat p_1, \ldots, \hat p_n]]$ by conditioning on the realization of $\hat p_1, \ldots, \hat p_n$. 
 In particular, although the Fisher information function $\mathcal{I}(q)$ depends on the historical prices $\hat  p_1, \ldots, \hat p_n$, since the support of $q(\cdot)$ is independent of $\hat p_1, \ldots, \hat p_n$ and the function $C(\theta)$ and its derivative are bounded,  we can verify that  
 $\mathcal{I}(q)=\Theta({\hat\delta}^{-2})$ with the hidden constant   independent of the realization of $\hat  p_1, \ldots, \hat p_n$.

 \underline{\textbf{Lower bound: Step 2.}} In this step, we complete the proof by
 showing that when  $ \mathbb{E}_{\theta}^{\hat \pi}[\hat  \delta(\theta)] \lesssim T^{-\frac{1}{4}}(\log T)^{-\frac{1}{2}}$ and $\mathbb{E}_{\theta}^{\hat \pi}[n \hat \sigma^2] \lesssim  \frac{\sqrt T}{\log T}$, then for any admissible policy $\pi \in \hat \Pi^{\circ}$, 
 \begin{align}\label{eq1-proof-extension}
 \mathbb{E}_{\hat p_1, \ldots, \hat p_n}[ R^{\pi}_{\theta}(T,\hat p_1, \ldots, \hat p_{n})]=\Omega(\frac{\sqrt T}{\log T}). 
 \end{align}  
 To show \eqref{eq1-proof-extension},   for  any given realization of $\hat p_1, \ldots, \hat p_n$, we  define  two parameters $\theta_1$ and $\theta_2$ satisfying 
 \begin{align*}%
 -\frac{\alpha_1}{2\beta_1} = \bar p_{1:n}+\hat \delta, \quad 	-\frac{\alpha_2}{2\beta_2} = \bar p_{1:n} +\hat \delta+\Delta, \quad  \alpha_1-\alpha_2=-(\beta_1-\beta_2)\bar p_{1:n}, 
 \end{align*}
 where  $\Delta>0$ is  to be determined.  
 Then we define two random variables $X=(\hat D_1, \ldots, \hat D_n, D_1, \ldots, D_n, p_1, \ldots, p_n)$ and $Y=(\hat p_1, \ldots, \hat p_n)$. For any policy $\pi$, let $\prob^{\pi}_i(X, Y)$ be the joint distribution of $(X, Y)$, $\prob^{\pi}_{i}(X|Y)$ be the conditional probability measure of $X$ given $Y$, and $\prob^{\pi}_i(X)$ be the marginal distribution of $X$, each of which is associated with the policy $\pi$ and demand parameter $\theta_i$, $i=1, 2$.  Then we have 
 \begin{align}\label{eq3-proof-extension}
 &\mathbb{E}_{Y \sim \prob^{\pi}_1}\big[KL\big(\prob^{\pi}_1(X|Y),\prob^{\pi}_2(X|Y)\big)\big]  \nonumber \\
 &\quad \le KL(\prob^{\pi}_1(X,Y), \prob^{\pi}_2(X,Y)) 
 \nonumber \nonumber \\
 &\quad  =\frac{1}{2R^2}\Big( \sum_{i=1}^{n}\mathbb{E}_{\theta_1}^{ \hat \pi}[\big((\alpha_1-\alpha_2)+(\beta_1-\beta_2) \hat p_i\big)^2]+\sum_{t=1}^{T} \mathbb{E}_{\theta_1}^{ \hat \pi,  \pi}\big[\big((\alpha_1-\alpha_2)+(\beta_1-\beta_2) p_t\big)^2\big]\Big) \nonumber \\ 
 &\quad \le \frac{(\beta_1-\beta_2)^2}{2R^2}\Big( n\mathbb{E}_{\theta_1}^{ \hat \pi}[\hat \sigma^2]+2\sum_{t=1}^{T} \mathbb{E}_{\theta_1}^{\hat \pi, \pi}\big[\big( p_t-\psi(\theta_1)\big)^2\big]+2T\mathbb{E}_{\theta_1}^{\hat \pi}[\hat \delta^2(\theta_1)]\Big),
 \end{align}   
 where the first inequality holds since  from the chain rule of KL divergence, $KL\big(\prob^{\pi}_1(X,Y),\prob^{\pi}_2(X,Y)\big) = KL\big(\prob^{\pi}_1(Y),\prob^{\pi}_2(Y)\big) +\mathbb{E}_{Y \sim \prob^{\pi}_1}\big[KL\big(\prob^{\pi}_1(X|Y),\prob^{\pi}_2(X|Y)\big)\big]$, and   $KL\big(\prob^{\pi}_1(Y),\prob^{\pi}_2(Y)\big)\ge 0$. 
 
 On the other hand,  by applying  Theorem 2.2 in \cite{Tsybakov2009} and using the fact that $\pi$ is admissible,  we have   
 \begin{align}\label{eq2-proof-extension}
 \frac{1}{32}  e^{-KL(\prob^{\pi}_1(X|Y), \prob^{\pi}_2(X|Y)) }\cdot T\Delta^2  & \le \sum_{t=1}^{T}\mathbb{E}_{\theta_1}^{\pi}[(p_t-\psi(\theta_1))^2|\hat p_1, \ldots, \hat p_n] +  \sum_{t=1}^{T}\mathbb{E}_{\theta_2}^{\pi}[(p_t-\psi(\theta_1))^2|\hat p_1, \ldots, \hat p_n]   \nonumber \\
 & \le 2K_0\sqrt T \log T.
 \end{align}
 Taking the expectation over $Y$ on both sides of \eqref{eq2-proof-extension},  we conclude from Jensen's inequality that 
 \begin{align*}
 \mathbb{E}_{Y \sim \prob^{\pi}_1}[KL(\prob^{\pi}_1(X|Y), \prob^{\pi}_2(X|Y)) ]\ge \log\big(\frac{\sqrt T\Delta^2}{64K_0\log T}\big),
 \end{align*}
 With   inequalities  \eqref{eq3-proof-extension}, the remaining analysis is similar to  Step 2 in the proof of Theorem~\ref{thm-lower bound-multiple historical prices} and therefore is omitted.\qed

	\section*{Appendix G. Multi-Armed Bandits with Offline  Data}\label{app-MAB}
  	 In this appendix, we discuss  the MAB with offline data and show the optimal regret rate exhibits phase transitions by deploying results from \cite{shivaswamy2012multi} and \cite{gur2019adaptive}. We also compare the MAB problem with the \texttt{OPOD} problem studied in this paper.

 Consider  a  $K$-armed bandit, where the seller chooses arms from   set $\{1,2,\ldots,K\}$ for each period $t\in[T]$. The distribution of reward for each arm $i$ is sub-Gaussian, denoted by $\mathcal{D}_i$, with the mean value $\mu_i$, $i \in [K]$. Let  $i^*$ be  the arm with the highest mean reward, i.e., $\mu_{i^*}=\max\{\mu_i: i \in [K]\}$, 
 $\Delta_i$ be the  sub-optimality gap for each  arm $i \neq i^*$, i.e., 
 $\Delta_i=\mu_{i^*}-\mu_i$, and $\Delta$ be a lower bound such that $0<\Delta \le \min\{\Delta_{i}: i \in [K], i\neq i^*\}$. We denote  $\mathcal{S}=(\Delta,\mathcal{D}_1, \ldots, \mathcal{D}_K)$ as the class that includes any possible latent rewards distributions with  the lower bound $\Delta$. 
 
 We assume that before the start of online learning, 
 there are some pre-existing offline data, which consists of  $H_i$  observations of random rewards for each arm $i \in [K]$. The decision maker can use  the offline data as well as   online data to make online decisions. For any given  latent distributions of rewards $\mathcal{D}:=(\mathcal{D}_1, \mathcal{D}_2, \ldots, \mathcal{D}_K)$,  
 we define the regret of any learning policy $\pi$ as the worst-case difference between the expected rewards generated  by the optimal clairvoyant policy and the policy $\pi$: 
$
 R^{\pi}(T)=\sup_{\mathcal{S}} \big\{T\mu_{i^*}-\mathbb{E}_{\mathcal{D}}^{\pi}\big[\sum_{t=1}^{T} \mu_{\pi_t}\big]\big\},
$
 where the operator $\mathbb{E}_{\mathcal{D}}^{\pi}[\cdot]$ denotes the expectation induced by the policy $\pi$ and the latent distribution $\mathcal{D}$. 
 The optimal regret is defined as $R^*(T)=\inf_{\pi} R^{\pi}(T)$, which naturally depends on the number of offline observations $H_1, H_2, \ldots, H_K$, and therefore, is also denoted as $R^*(T,H_1,\ldots,H_K)$. 
 
 We first present the upper and lower bounds on the optimal regret for the $K$-armed bandit with offline data  in the following proposition,  where the regret  upper bound is provided in  Theorem 2 of \cite{shivaswamy2012multi}, and the regret  lower bound is implied  from  Theorem 1 of \cite{gur2019adaptive}. 
 \begin{proposition}\label{prop-bandit with history}
 	There exist positive constants $C_1, C_2, C_3, C_4$ such that the optimal regret $R^*(T,H_1,\ldots,H_K)$ satisfies  
 	\begin{align}
 	&R^*(T,H_1,\ldots,H_K)\le \sum_{i=1}^{K}\Delta_{i}\Big(\big(\frac{8 \log (T+H_i)}{\Delta_i^2}-H_i\big)^++C_1\Big), \label{ineq-upper bound-karmed bandit}\\
 	& R^*(T,H_1,\ldots,H_K)\ge C_2 \sum_{i=1}^{K}\Delta\Big(\frac{\log T}{\Delta^2}-C_3H_i+\frac{1}{\Delta^2} \log \frac{C_4\Delta^2}{K}\Big)^+.  \label{ineq-lower bound-karmed bandit}
 	\end{align}
 \end{proposition}

\begin{figure}[htbp!]
 	\centering
 	\includegraphics[width=0.4\textwidth]{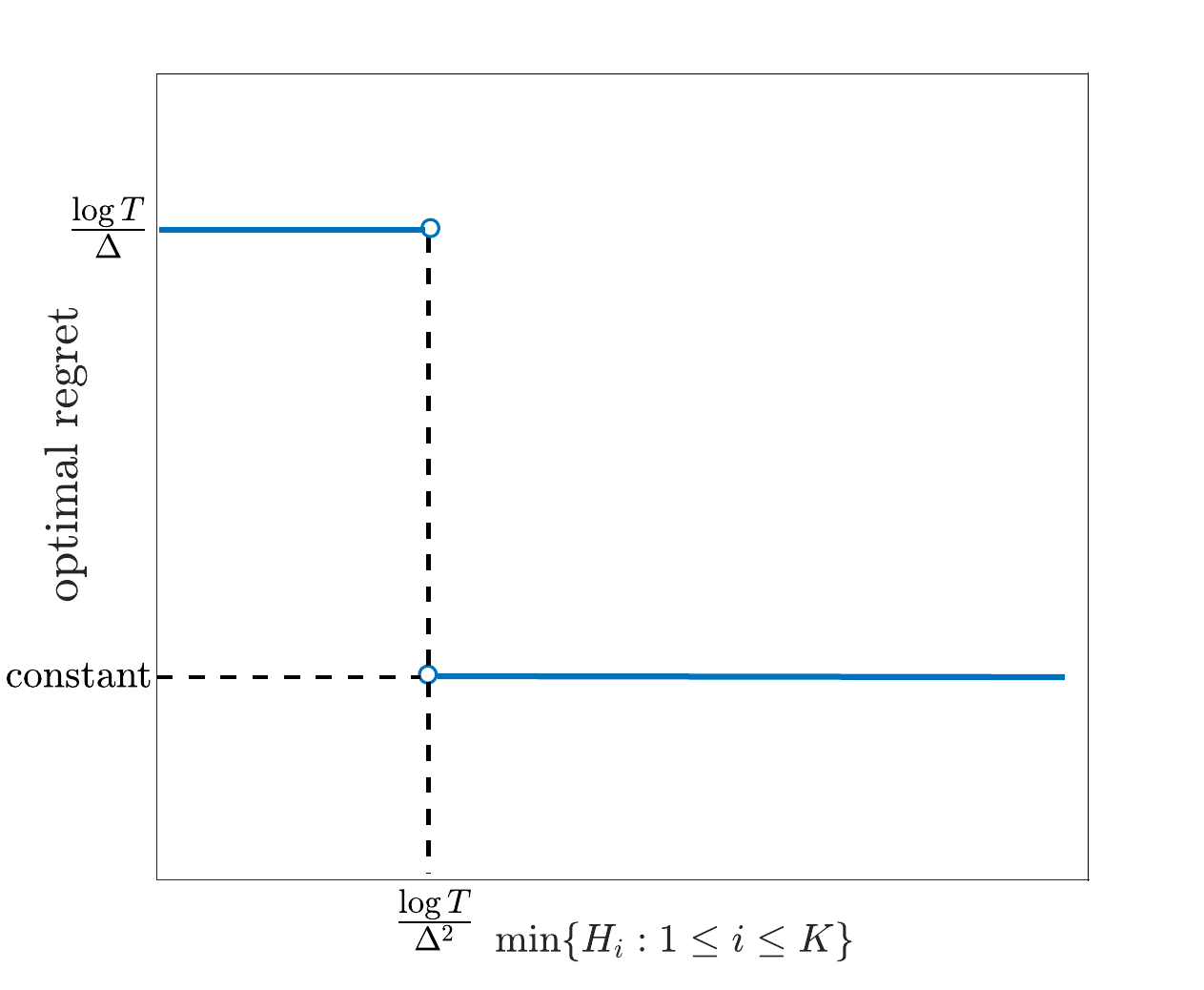}
 	\caption{Phase transition in $K$-armed bandits with offline data when $\Delta=\Omega(T^{-\frac{1}{2}})$}
 	\label{Karm bandit-phase transition}
 \end{figure}

 Note that the  regret lower bound in  \eqref{ineq-lower bound-karmed bandit}  is nontrivial only when $\Delta\gtrsim T^{-\frac{1}{2}}$.  
 Combining  \eqref{ineq-upper bound-karmed bandit} and \eqref{ineq-lower bound-karmed bandit},  we discover the following phase transitions for $K$-armed bandits under the assumption of $\Delta=\Omega(T^{-\frac{1}{2}})$:  
 the optimal regret rate in $K$-armed bandits decreases from $\Theta( \frac{\log T}{\Delta})$ to constant when the number of offline observations for each arm  exceeds $\Theta(\frac{\log T}{\Delta^2})$.   See  Figure \ref{Karm bandit-phase transition} for illustration.  We also point out that the optimal regret and phase transitions are not characterized  in  \cite{shivaswamy2012multi} or \cite{gur2019adaptive}, since there is no regret  lower bound developed in \cite{shivaswamy2012multi}, and in order to characterize more general information flow,  the upper bound developed in \cite{gur2019adaptive} does not match  their lower bound when $\Delta$ is not necessarily a constant. 
 
There are  three  key differences between the $K$-armed bandit with   offline data and the \texttt{OPOD} problem considered in this paper. First, in the $K$-armed bandit, the phase transition of the optimal regret  only occur  when the amount of offline data for \textit{each} arm exceeds the threshold $\Theta(\frac{\log T}{\Delta^2})$; in other words, the offline data need to be {\textit{balanced}} among different arms. By contrast, in the \texttt{OPOD} problem, even when there is only one historical price, i.e., $\sigma=0$,   
 the optimal regret  rate can drop from $\widetilde \Theta(\sqrt T)$ to $\widetilde{\Theta}(\frac{\log T}{\delta^2})$ as the amount of offline data $n$ increases.  The reason is that in $K$-armed  bandit, different arms are independent, and the knowledge about the reward of one arm does not help to understand that of another. However, in dynamic pricing, the demands under different prices  are connected with each other by the parametric assumption of the linear demand function.    Therefore, knowing even one point at the demand curve can  lead to a significant decrease in the optimal regret rate (see our Corollary~\ref{coro-opt regret-single} when  $n =\infty$, or the incumbent price setting in \citealt{KeskinZeevi2014}). 
 Second, in the $K$-armed bandit,  the optimal regret rate only shows two phases. In the first phase when the amount of the offline data is small, i.e., $\min\{H_i: i \in [K]\}=O(\frac{\log T}{\Delta^2})$, the optimal regret is always $\Theta(\frac{\log T}{\Delta})$ and the offline data do not help to  reduce the optimal regret.  In the second phase when the amount of the offline data is large, i.e.,  $\min\{H_i: i \in [K]\}=\Omega(\frac{\log T}{\Delta^2})$, the optimal regret becomes a constant.  
 In  the \texttt{OPOD} problem, however, the optimal regret gradually  changes  as the size of the offline data increases,  experiencing different phase transitions depending on the magnitude of $\sigma$ and $\delta$. Third, under the so-called ``well-separated'' condition in bandits, where the sub-optimality gap $\Delta$ is a constant independent of $T$, the $K$-armed bandit exhibits \textit{weak} phase transition in the sense that the drop of the optimal regret rate is within $\log T$. By comparison, under our well-separated condition , i.e.,  $\delta$ is a constant independent of  $T$, the \texttt{OPOD} problem shows \textit{strong} phase transitions in the sense that  
 the drops of the optimal regret rate in multiple phases are measured in $T^{\kappa}$ for some  $\kappa >0$, which are much more significant even if  we ignore the logarithmic factor.

 	\section*{Appendix H. Tables in Sections \ref{sec-UCB} and \ref{sec-extension-historical price}}
 
\begin{table}[h!] \caption{Regret upper bound   in Theorem \ref{thm-UCB} for the  single-historical-price setting}
	\label{table-general case-upper bound-single}\vspace{.1in}
	\begin{center}
			\begin{threeparttable}     
		\setlength{\tabcolsep}{2.5mm}{
			\begin{tabular}{l||c|c|c}
				\toprule 	\hline 
				\multicolumn{4}{l}{ \textbf{Case 1:}  $\delta   \gtrsim T^{-\frac{1}{4}}(\log T)^{\frac{1}{2}}$}                                                                                 \\ \hline %
				offline sample size        & $0 \le n \lesssim \frac{\sqrt T\log T}{\delta^{2}}$ & $\frac{\sqrt T \log T}{\delta^{2}} \lesssim n \lesssim T$  & $n \gtrsim T$                        \\ \hline
				upper bound  & $\mathcal{O}(\sqrt T \log T)$               & $\mathcal{O}(\frac{T(\log T)^2}{n\delta^2})$                        & $\mathcal{O}(\frac{(\log T)^2}{\delta^2})$                       \\ \hline \hline 
				\multicolumn{4}{l}{ \textbf{Case 2:}  $\delta \lesssim  T^{-\frac{1}{4}}(\log T)^{\frac{1}{2}}$}                                                                               \\ \hline %
				offline sample size                                          & \multicolumn{3}{c}{$n\ge 0$}  \\ \hline
				upper bound & \multicolumn{3}{c}{$\mathcal{O}(\sqrt T \log T)$}                                                                                                                    \\
				\hline \bottomrule
		\end{tabular}
	 	 }
		\end{threeparttable}     
	\end{center}
\end{table}

\begin{table}[h!] \caption{Regret lower bound   in Theorem \ref{thm-lower bound} for the  single-historical-price setting}
	\label{table-general case-lower bound-single}\vspace{.1in}
	\begin{center}
		\begin{threeparttable}
		\setlength{\tabcolsep}{2.5mm}{
			\begin{tabular}{l||c|c|c}
				\toprule 	\hline 
				\multicolumn{4}{l}{ \textbf{Case 1:} $\delta \gtrsim T^{-\frac{1}{4}}$}                                                                                 \\ \hline %
				offline sample size        & $0<n \lesssim \frac{\sqrt T}{\delta^{2}}$ & $\frac{\sqrt T}{\delta^{2}} \lesssim n \lesssim T$  & $n \gtrsim T$                        \\ \hline
				lower bound  & $\Omega(\sqrt T)$               &  $\Omega(\frac{T}{n\delta^2}\vee \log T)$                  & $\Omega(\frac{1}{\delta^2}\vee \log T)$                       \\ \hline \hline 
				\multicolumn{4}{l}{ \textbf{Case 2:} $T^{-\frac{1}{4}}(\log T)^{-\frac{1}{2}}\lesssim \delta \lesssim T^{-\frac{1}{4}}$}                                                                                \\ \hline %
				offline sample size                                          & \multicolumn{3}{c}{$n > 0$}
				\\ \hline
				lower bound & \multicolumn{3}{c}{$\Omega(T\delta^2)$}                                                                                                               \\  
				\hline 
				\multicolumn{4}{l}{ \textbf{Case 3:} $\delta \lesssim  T^{-\frac{1}{4}}(\log T)^{-\frac{1}{2}}$}                                                                                \\ \hline %
				offline sample size                                          & \multicolumn{3}{c}{$n > 0$}
				\\ \hline
				lower bound & \multicolumn{3}{c}{$\Omega(\frac{\sqrt T}{\log T})$}                                                                                                               \\
				\hline \bottomrule
		\end{tabular} 
}
\end{threeparttable}
	\end{center}
\end{table}

\begin{table}[htbp!] \caption{Regret upper bound in Theorem  \ref{thm-upper bound-multiple prices-T} for the multiple-historical-price setting }
	\label{table-multiple historical prices-upper bound}
	\begin{center}
		\begin{threeparttable}
		\setlength{\tabcolsep}{4mm}{
			\begin{tabular}{l||c|c|c|c}
				\toprule 	\hline 
				
				\multicolumn{5}{l}{\textbf{Case 1:} $\delta \gtrsim T^{-\frac{1}{4}}(\log T)^{\frac{1}{2}}$ and  $\sigma \lesssim  \delta  $}                                                                                \\ \hline %
				offline sample size                                 & $0\le n \lesssim \frac{\sqrt{T}\log T}{\delta^2}$   & $  \frac{\sqrt{T}\log T}{\delta^2} \lesssim n \lesssim T$ & $T \lesssim  n  \lesssim \frac{T \delta^2}{\sigma^2}$ & $n \gtrsim \frac{T \delta^2}{\sigma^2}$ \\ \hline 
				upper bound & $\mathcal{O}(\sqrt T \log T)$ & $\mathcal{O}(\frac{T(\log T)^2}{n\delta^2})$   &$\mathcal{O}(\frac{(\log T)^2}{\delta^2})$    & $\mathcal{O}(\frac{T(\log T)^2}{n\sigma^2}+1)$         \\ \hline
				\hline
				\multicolumn{5}{l}{\textbf{Case 2:} $\delta \gtrsim T^{-\frac{1}{4}}(\log T)^{\frac{1}{2}}$ and $\sigma \gtrsim \delta$ 
				}                                                                               
				\\ \hline  
				offline sample size & \multicolumn{2}{c|}{$0 \le n \lesssim \frac{\sqrt T\log T}{\sigma^2}$ } &  \multicolumn{2}{c}{$n \gtrsim \frac{\sqrt T\log T}{\sigma^2}$}  \\  
				\hline
				upper bound & \multicolumn{2}{c|}{$\mathcal{O}(\sqrt T\log T)$ } &  \multicolumn{2}{c}{$\mathcal{O}(\frac{T(\log T)^2}{n\sigma^2}+1)$}  \\  
				\hline
				\hline
				\multicolumn{5}{l}{\textbf{Case 3:} $ \delta \lesssim T^{-\frac{1}{4}}(\log T)^{\frac{1}{2}}$
				}   \\
				\hline  
				offline sample size & \multicolumn{1}{c|}{$0 \le n \lesssim \frac{\sqrt T\log T}{\sigma^2}$ } &  \multicolumn{1}{c|}{$\frac{\sqrt T\log T}{\sigma^2} \lesssim n \lesssim \frac{(\log T)^2}{\sigma^2\delta^2}$}  & \multicolumn{2}{c}{ $ n\gtrsim  \frac{(\log T)^2}{\sigma^2\delta^2}$} \\  
				\hline
				upper bound & \multicolumn{1}{c|}{$\mathcal{O}(\sqrt T \log T)$ } &  \multicolumn{1}{c|}{$\mathcal{O}(T\delta^2+1)$}  & \multicolumn{2}{c}{$\mathcal{O}(\frac{T(\log T)^2}{n\sigma^2}+1)$}
				\\ \hline \bottomrule 
		\end{tabular}}
		 
\end{threeparttable}  
	\end{center}
\end{table}

 \begin{table}[htbp!] \caption{Regret lower bound in Theorem \ref{thm-lower bound-multiple historical prices} for the multiple-historical-price setting }
 	\label{table-multiple historical prices-lower bound}\vspace{.1in}
 	\begin{center}
 	\begin{threeparttable} 
 		\setlength{\tabcolsep}{4.85mm}{
 			\begin{tabular}{l||c|c|c|c}
 				\toprule 	\hline 
 				
 				\multicolumn{5}{l}{\textbf{Case 1:} $\delta \gtrsim T^{-\frac{1}{4}}(\log T)^{\frac{1}{2}}$ and  $\sigma \lesssim  \delta $}                                                                                \\ \hline %
 				offline sample size                                 & $0\le n \lesssim \frac{\sqrt{T}\log T}{\delta^2}$   & $  \frac{\sqrt{T} \log T}{\delta^2} \lesssim n \lesssim T$ & $T \lesssim  n  \lesssim \frac{T \delta^2}{\sigma^2}$ & $n \gtrsim \frac{T \delta^2}{\sigma^2}$ \\ \hline 
 				lower bound & $\Omega(\frac{\sqrt T}{\log T})$ & $\Omega(\frac{T}{n\delta^2})$   &$\Omega(\frac{1}{\delta^2})$    & $\Omega(\frac{T}{n\sigma^2})$         \\ \hline
 				\hline
 				\multicolumn{5}{l}{  \textbf{Case 2:} $\delta \gtrsim T^{-\frac{1}{4}}(\log T)^{\frac{1}{2}}$ and $\sigma \gtrsim \delta$ 
 				}                                                                               
 				\\ \hline  
 				offline sample size & \multicolumn{2}{c|}{$0 \le n \lesssim \frac{\sqrt T \log T}{\sigma^2}$ } &  \multicolumn{2}{c}{$n \gtrsim \frac{\sqrt T\log T}{\sigma^2}$}  \\  
 				\hline
 				lower bound & \multicolumn{2}{c|}{$\Omega(\frac{\sqrt T}{\log T})$ } &  \multicolumn{2}{c}{$\Omega(\frac{T}{n\sigma^2})$}  \\  
 				\hline
 				\hline
 				\multicolumn{5}{l}{  \textbf{Case 3:} $T^{-\frac{1}{4}}(\log T)^{-\frac{1}{2}} \lesssim \delta \lesssim T^{-\frac{1}{4}}(\log T)^{\frac{1}{2}}$
 				}   \\
 				\hline  
 				offline sample size & \multicolumn{2}{c|}{$0 \le n \lesssim \frac{1}{\sigma^2\delta^2}$ } &  \multicolumn{2}{c}{$n \gtrsim \frac{1}{\sigma^2\delta^2}$}  \\  
 				\hline
 				lower bound & \multicolumn{2}{c|}{$\Omega(\frac{\sqrt T}{\log T})$ } &  \multicolumn{2}{c}{$\Omega(\frac{T}{n\sigma^2})$}  \\  
 				\hline
 				\multicolumn{5}{l}{  \textbf{Case 4:} $ \delta \lesssim T^{-\frac{1}{4}}(\log T)^{-\frac{1}{2}}$
 				}   \\
 				\hline  
 				offline sample size & \multicolumn{1}{c|}{ $0 \le n \lesssim \frac{\sqrt T\log T}{\sigma^2}$ } &  \multicolumn{1}{c|}{ $\frac{\sqrt T\log T}{\sigma^2 } \lesssim n \lesssim \frac{1}{\sigma^2\delta^2}$}  & \multicolumn{2}{c}{$ n\gtrsim  \frac{1}{\sigma^2\delta^2}$} \\  
 				\hline
 				lower bound & \multicolumn{1}{c|}{$\Omega(\frac{\sqrt T}{\log T})$ } &  \multicolumn{1}{c|}{$\Omega(T\delta^2)$}  & \multicolumn{2}{c}{$\Omega(\frac{T}{n\sigma^2})$}
 				\\ \hline \bottomrule 
 		\end{tabular}
}
\end{threeparttable}  
 	\end{center}
 \end{table}

 \end{document}